\documentclass[preprint,12pt,number]{elsarticle}
\usepackage{float}
\usepackage[utf8]{inputenc}
\usepackage[T5]{fontenc}
\usepackage{xcolor}
\usepackage{caption}
\usepackage[justification=centering]{caption}
\usepackage{tabularx}
\usepackage{array}
\usepackage{booktabs} 
\usepackage{lscape}
\usepackage{orcidlink}

\usepackage{longtable}   
\usepackage{adjustbox}
\usepackage{multirow}
\usepackage{subcaption}
\usepackage{pdflscape}
\usepackage{amssymb}
\usepackage{amsmath}
\usepackage{graphicx}
\usepackage{hyperref}

\begin{document}

\begin{frontmatter}

\title{Cross-Architectural Mixture-of-Experts with Adaptive Soft Routing for Plant Leaf Disease Classification}

\author{Phi-Hung Hoang}
\ead{hunghpde180523@fpt.edu.vn}

\author{Thi-Thu-Hong Phan\corref{cor}\orcidlink{0000-0001-6880-3721}}
\ead{hongptt11@fe.edu.vn}

\cortext[cor]{Corresponding author}

\hypersetup{pdfauthor={Thi-Thu-Hong Phan}}

\affiliation{organization={AIT laboratory, Faculty of Artificial Intelligence, \\FPT University},
            addressline={Da Nang},
            postcode={550000}, 
            country={Viet Nam}}

\begin{abstract}

Plant leaf disease classification is crucial for crop protection and precision agriculture but remains challenging under complex backgrounds, illumination variations, and severe class imbalance. Moreover, single-architecture models often fail to effectively capture both local and global representations. To address these challenges, this study proposes an adaptive soft Mixture-of-Experts (MoE) framework with cross-architectural routing that integrates EfficientNet-B0, DenseNet-121, and Swin-Tiny to exploit complementary multi-scale, local, and global features. A soft gating mechanism dynamically assigns input-dependent expert weights, while a two-stage refinement training strategy improves optimization stability and generalization. Experiments on a highly imbalanced potato leaf disease dataset achieve 91.68\% recall and 92.62\% F1-score, surpassing the strongest individual expert by 5.91\% and 5.03\%, respectively. Additional evaluations on durian and sesame leaf disease datasets yield F1-scores of 94.03\% and 97.04\%, demonstrating robust cross-dataset generalization and the potential of the proposed framework for reliable real-world crop health monitoring.
\end{abstract}

\begin{keyword}

Adaptive routing; Mixture-of-experts (MoE); Plant leaf disease classification; Class imbalance; Grad-CAM

\end{keyword}

\end{frontmatter}

\section{Introduction}

Plant leaf disease classification plays an important role in crop protection, food security, and sustainable agriculture. In precision agriculture, automated and real-time crop health monitoring has become increasingly important, as traditional diagnosis based on visual inspection by agronomists is time-consuming, labor-intensive, subjective, and impractical for large-scale deployment~\citep{Yao2023LeafDiseaseML, Pacal2024PlantDiseaseDL}. Therefore, developing robust computer vision systems for accurate plant leaf disease identification remains a critical research challenge.

Early plant leaf disease recognition methods relied on handcrafted features and conventional machine learning techniques, which struggled to generalize under complex real-world conditions due to their dependence on manual feature engineering~\citep{Sridhar2025, Hoang2026BorderlineSMOTE, Hoang2026FeatureSelection}. These limitations drove the adoption of deep learning approaches, particularly Convolutional Neural Networks (CNNs), which substantially improved classification through automatic hierarchical feature learning~\citep{Shabrina2024, Rivaldo2025, Mhala2025}. More recently, Vision Transformers (ViTs) have further enhanced contextual modeling through self-attention mechanisms~\citep{Meghana2025, Murugavalli2025, Tabassum2026}.

Despite these advances, robust plant leaf disease classification under real-world agricultural conditions remains challenging due to complex backgrounds, illumination variation, viewpoint changes, and severe class imbalance. A central limitation lies in architectural adaptability: CNNs excel at local texture modeling but struggle with global contextual understanding, whereas Transformer-based models capture long-range dependencies but may lack a strong local inductive bias for subtle lesion representation. Hybrid and ensemble approaches attempt to combine these complementary strengths; however, most rely on static fusion strategies that assign fixed importance to expert models regardless of input complexity~\citep{Apleni2025, Sinamenye2025, Ahmad2026}. Consequently, they cannot adaptively emphasize the most informative representations for diverse disease patterns. Severe class imbalance further compounds this challenge in practical agricultural environments.

To address these challenges, this study proposes an Adaptive Soft Mixture-of-Experts (MoE) framework that enables adaptive collaboration among heterogeneous expert architectures for robust plant leaf disease classification. Unlike conventional static fusion approaches, the proposed framework dynamically adjusts expert contributions based on learned feature representations, allowing the model to emphasize the most informative expert responses for different disease patterns. Furthermore, a refinement-based training strategy is introduced to improve optimization stability and enhance generalization under challenging imbalanced conditions.

The main contributions of this study are as follows:

\begin{itemize}

\item We propose a cross-architectural adaptive soft MoE framework that dynamically routes complementary CNN- and Transformer-based expert contributions based on learned feature representations, thereby overcoming the limitations of static fusion strategies.

\item We introduce a two-stage refinement training strategy that progressively improves optimization stability and generalization under class-imbalanced conditions.

\item We conduct a comprehensive empirical evaluation on a highly imbalanced real-world potato leaf disease dataset and further assess cross-dataset generalization on durian and sesame leaf disease datasets.

\item We perform interpretability-driven analysis using gating entropy and Grad-CAM to provide insight into adaptive expert collaboration and feature attention behavior.

\end{itemize}

The rest of the paper is organized as follows. Section \ref{sec:related_works} reviews related works in the literature. Section \ref{sec:methodology} describes the proposed adaptive soft MoE framework. Section \ref{sec:experiments} presents the datasets and experimental setup. Section \ref{sec:results_discussion} reports the experimental results, interpretability analysis, and discussion. Finally, Section \ref{sec:conclusion} concludes the paper and outlines future research directions.

\section{Related works}\label{sec:related_works}

Traditional plant disease classification methods primarily relied on handcrafted features combined with conventional machine learning (ML) classifiers. To capture disease characteristics, researchers typically extracted color, texture, and shape descriptors to train ML classifiers. For instance, Al-Shamasneh~\cite{AlShamasneh2025} proposed a method based on Generalized Jones Polynomials (GJPs) with a Support Vector Machine (SVM), while Sridhar and Angamuthu~\cite{Sridhar2025} combined bilateral filtering, GraphCut segmentation, and hybrid texture features (GLCM, LBP) for improved classification. To address high dimensionality, optimization techniques have also been employed, such as the Whale Optimization Algorithm (WOA) in Li and Javidan~\cite{Li2025} for feature selection. While effective in controlled settings, these pipelines often lack robustness under real-world conditions. To improve robustness in more complex environments, Hoang and Phan (2026a)~\cite{Hoang2026BorderlineSMOTE} integrated global color statistics with local descriptors (SIFT, KAZE) using a Bag-of-Visual-Words (BoVW) framework with LightGBM. In a related study, Hoang and Phan (2026b)~\cite{Hoang2026FeatureSelection} further showed that embedded feature selection can reduce dimensionality while maintaining classification stability. Despite these efforts, handcrafted feature-based approaches remain limited in scalability and robustness under complex environmental variations.

To overcome these limitations, Convolutional Neural Networks (CNNs) became the prevailing approach in agricultural image analysis. Pre-trained CNN models have been widely adopted with transfer learning and optimization strategies to improve disease identification. For instance, Shabrina et al.~\cite{Shabrina2024} evaluated several pre-trained architectures, including EfficientNet, MobileNet, VGG, ResNet, and DenseNet, under uncontrolled conditions. To mitigate the effects of noise and class imbalance, Mhala et al.~\cite{Mhala2025} proposed an optimized DenseNet-based approach incorporating data augmentation, early stopping, and L2 regularization. Subsequent studies explored more efficient CNN architectures to balance classification performance and computational cost. Hoang et al.~\cite{Hoang2025} introduced a hybrid knowledge distillation framework for edge devices, while Zhang et al.~\cite{Zhang2025} proposed a lightweight LDL-MobileNetV3S architecture that integrates multi-scale fusion, dilated convolutions, and localized attention to improve lesion detection with reduced complexity. Despite these advances, CNNs remain limited by their localized receptive fields, which restrict global context modeling. Vision Transformers (ViTs) and hierarchical variants have therefore emerged, leveraging self-attention to capture long-range dependencies. Studies by Meghana et al.~\cite{Meghana2025} and Murugavalli and Gopi~\cite{Murugavalli2025} demonstrated improved recognition of subtle and dispersed disease patterns using ViT architectures, while Tabassum and Nunavath~\cite{Tabassum2026} demonstrated the effectiveness of Swin Transformer models in fine-grained agricultural tasks. However, relying on a single architectural paradigm still leads to a trade-off between local detail modeling and global contextual understanding. 

Moreover, architectural improvements alone are insufficient, as real-world agricultural environments also present severe class imbalance challenges. Recent studies have therefore explored both data-level and algorithm-level mitigation strategies. For example, Aishawarya et al.~\cite{Aishwarya2025} and Rofiqi et al.~\cite{Rofiqi2026} applied SMOTE with hierarchical transformers and lightweight CNNs to balance training distributions. At the algorithmic level, Mandhani et al.~\cite{Mandhani2026} combined Focal Loss with class weighting to emphasize hard-to-classify samples without generating synthetic data. 
Nevertheless, these approaches remain limited: oversampling may introduce noisy synthetic samples, while static loss weighting lacks adaptability to input-specific characteristics.

Ensemble learning and hybrid architectures have been widely explored to improve the robustness and generalization of individual deep learning models. By combining multiple networks, these approaches exploit complementary feature representations to enhance performance under complex conditions. For instance, Ahmad and Alamsyah~\cite{Ahmad2026} proposed a feature-level ensemble of DenseNet201 and MobileNetV2, demonstrating that combining representational richness with computational efficiency can improve disease classification. Similarly, Apleni et al.~\cite{Apleni2025} fused VGG16, ResNet50, and InceptionV3 through feature concatenation followed by a Multi-Layer Perceptron (MLP), achieving more consistent performance than single architectures under environmental variability. Hybrid CNN–Transformer models have also been explored to capture both local textures and global contextual information. Synamenye et al.~\cite{Sinamenye2025} combined EfficientNet-V2 with Vision Transformer (ViT) in a parallel design, achieving stronger performance under diverse conditions. Despite these advances, most ensemble methods rely on static fusion strategies such as averaging or fixed feature concatenation, assigning uniform weights regardless of input complexity. 
Consequently, they lack the adaptability needed to dynamically emphasize the most relevant features for different samples.

The Mixture of Experts (MoE) paradigm offers a promising alternative to static ensembles by introducing a dynamic, input-dependent routing mechanism. Rather than uniformly combining all models, MoE adaptively allocates expert contributions according to the characteristics of each input. In agricultural applications, several recent studies have explored its potential for disease and crop classification. Raya-González at al.~\cite{RayaGonzalez2025} proposed a hybrid CNN-SVM MoE for crop disease recognition, assigning visually similar classes to specialized experts to reduce class overlap. More recently, Salman et al.~\cite{Salman2025} integrated MoE with a Vision Transformer backbone for plant disease classification under field conditions, using entropy and orthogonal regularization to balance expert utilization. To address long-tail distributions and intra-class variability in complex agricultural environments, Lu et al.~\cite{Lu2025} introduced DMoE-ViT, which routes samples based on difficulty to specialized Vision Transformer (ViT) experts to reduce prediction uncertainty. Beyond image classification, Xu et al.~\cite{Xu2026} extended MoE to spectral data analysis and showed that SoftMoE, which aggregates all experts through weighted combination, outperforms hard-routing approaches by better capturing continuous variations.

Despite these promising advances, existing MoE studies in agricultural applications remain relatively limited, particularly in exploring adaptive soft routing across heterogeneous architectures for robust plant disease classification under real-world conditions. This motivates the present study, which investigates an adaptive soft MoE framework designed to dynamically integrate complementary expert representations for improved classification performance.

\section{Methodology}\label{sec:methodology}

\subsection{Overview of the proposed approach}

\begin{landscape}

\begin{figure}[H]
    \centering

    \begin{subfigure}{\linewidth}
        \centering
        \includegraphics[width=0.95\linewidth]{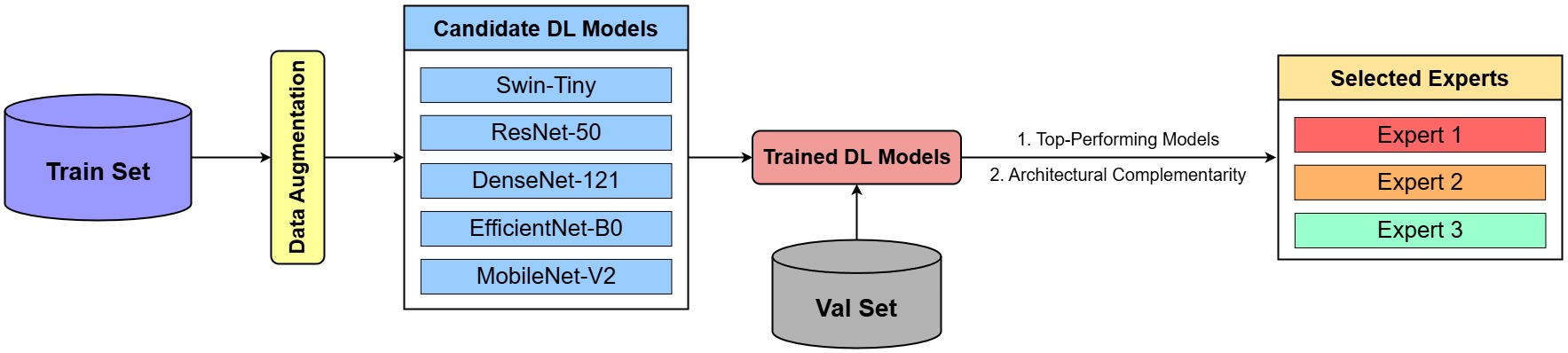}
        \caption{Phase 1: Selection of experts from candidate DL models.}
        \label{fig:subfig1}
    \end{subfigure}

    \vspace{0.2cm}

    \begin{subfigure}{\linewidth}
        \centering
        \includegraphics[width=0.95\linewidth]{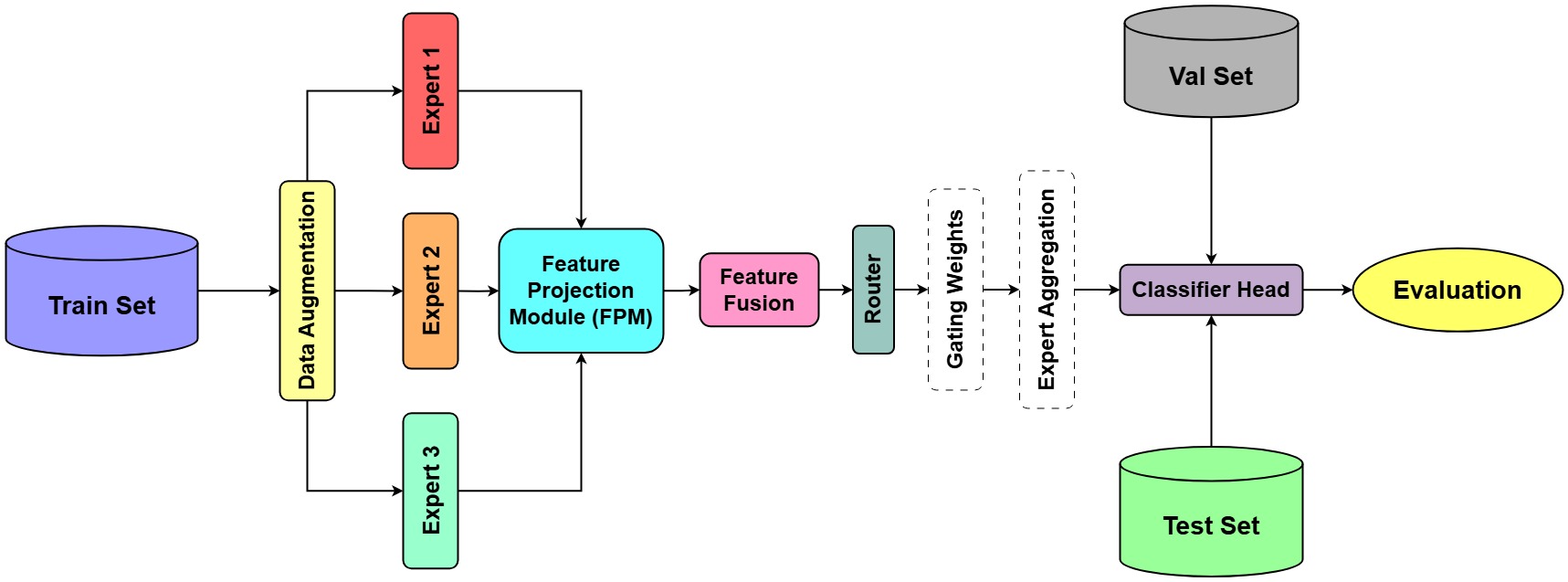}
        \caption{Phase 2: Proposed adaptive soft MoE classification framework.}
        \label{fig:subfig2}
    \end{subfigure}

    \caption{Overall architecture of the proposed Adaptive Soft MoE framework.}
    \label{fig:pipeline}
\end{figure}

\end{landscape}

Figure~\ref{fig:pipeline} illustrates the overall workflow of the proposed adaptive soft Mixture of Experts (MoE) framework, which is designed to address visual variations in real-world agricultural environments. The framework consists of two main phases: expert selection and construction of the adaptive soft MoE classification framework.

In the expert selection phase, a pool of heterogeneous deep learning architectures is considered as candidate experts to provide diverse feature extraction capabilities, including local texture modeling, multi-scale representation learning, and global contextual understanding. Candidate suitability is assessed based on architectural complementarity and competitive validation performance, ensuring that the selected experts provide diverse yet effective representations for the downstream MoE framework.

In the second phase, the selected backbones are integrated into the proposed adaptive soft MoE framework. Given an input image, feature extraction is performed in parallel by the three expert networks. Since heterogeneous architectures produce feature maps with different channel dimensions, a Feature Projection Module (FPM) is introduced to align them into a unified feature space, enabling stable cross-expert fusion.

To dynamically determine expert contributions, an input-adaptive routing mechanism is employed. Specifically, the aligned feature representations are first combined through a feature fusion step to construct a unified descriptor, which is subsequently processed by a routing network to generate continuous gating weights for each expert. The final fused representation is then obtained through weighted aggregation of the expert features, allowing adaptive integration of complementary visual information.

Finally, the aggregated representation is passed through a classifier head to generate the output probability distribution over the target classes. To optimize the proposed framework, a two-stage refinement training strategy 

\clearpage

\clearpage

\noindent is adopted. In the first stage, the entire framework is jointly fine-tuned to learn discriminative feature representations and input-dependent routing behavior. In the second stage, training resumes from the best-performing checkpoint using a reduced learning rate to further refine the learned representations, improve optimization stability, and enhance generalization.

\subsection{Candidate expert architectures}

To provide diverse feature extraction capabilities, five pre-trained architectures are considered as candidate expert models in Phase 1, including MobileNet-V2, EfficientNet-B0, DenseNet-121, ResNet-50, and Swin-Tiny. These models represent different architectural characteristics, such as lightweight convolution, dense feature reuse, multi-scale representation, and global self-attention. The architectural details of each candidate model are briefly described in the following subsections.

\subsubsection{MobileNet-V2}

\begin{figure}[H]
    \centering
    \includegraphics[width=\textwidth]{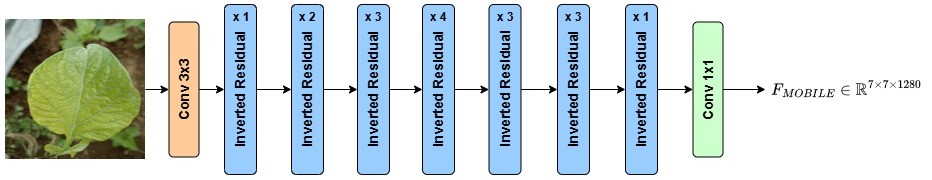}
    \caption{Architecture of the MobileNet-V2.}
    \label{fig:mobilenet_expert}
\end{figure}

MobileNet-V2 is a lightweight convolutional architecture known for its computational efficiency and compact design~\citep{Sandler2019}. As shown in Figure~\ref{fig:mobilenet_expert}, the network adopts the Inverted Residual with Linear Bottleneck design, where feature maps are expanded, processed using depthwise separable convolutions, and projected through linear $1 \times 1$ convolutions. This design reduces computational cost while preserving representative feature information. For an input image of size $224 \times 224$, the extracted feature map is denoted as $F_{MOBILE} \in \mathbb{R}^{7 \times 7 \times 1280}$.

\subsubsection{EfficientNet-B0}

\begin{figure}[H]
    \centering
    \includegraphics[width=\textwidth]{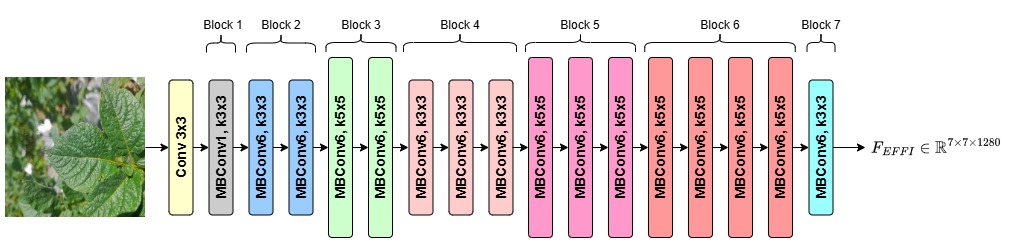}
    \caption{Architecture of the EfficientNet-B0.}
    \label{fig:efficientnet_expert}
\end{figure}

EfficientNet-B0 is designed for efficient feature extraction through compound scaling of network depth, width, and input resolution~\citep{Tan2020}. As shown in Figure~\ref{fig:efficientnet_expert}, the architecture is based on MBConv blocks with inverted residual structures, combining depthwise convolution, channel expansion, projection layers, and Squeeze-and-Excitation (SE) modules to enhance channel-wise feature representation. For an input image of size $224 \times 224$, the extracted feature map is denoted as $F_{EFFI} \in \mathbb{R}^{7 \times 7 \times 1280}$.

\subsubsection{DenseNet-121}

\begin{figure}[H]
    \centering
    \includegraphics[width=\textwidth]{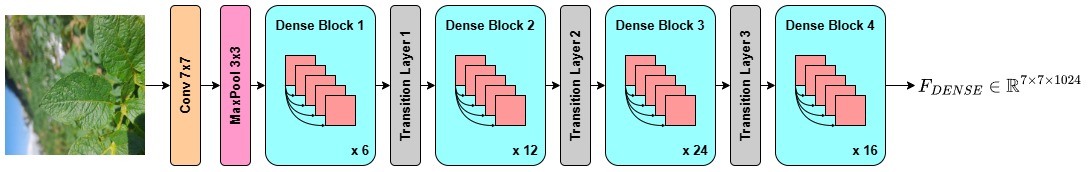}
    \caption{Architecture of the DenseNet-121.}
    \label{fig:densenet_expert}
\end{figure}

DenseNet-121 employs dense connectivity, where each layer receives feature maps from all preceding layers to encourage feature reuse and improved information flow~\citep{Huang2017}. As shown in Figure~\ref{fig:densenet_expert}, the architecture consists of dense blocks interconnected by transition layers with $1 \times 1$ convolution and average pooling for dimensionality reduction. The network follows a block configuration of $[6, 12, 24, 16]$ with a growth rate of 32. For an input image of size $224 \times 224$, the extracted feature map is denoted as $F_{DENSE} \in \mathbb{R}^{7 \times 7 \times 1024}$.

\subsubsection{ResNet-50}

\begin{figure}[H]
    \centering
    \includegraphics[width=\textwidth]{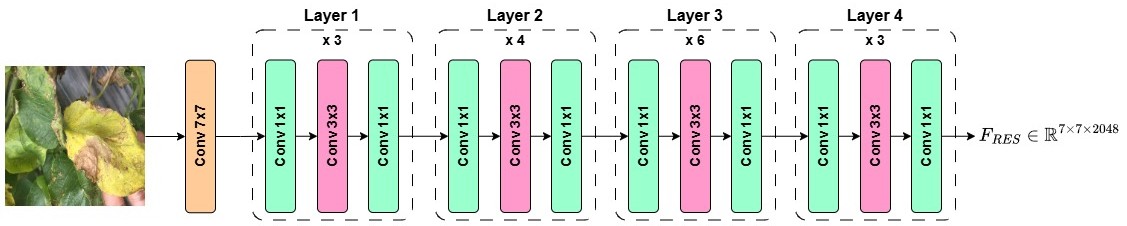}
    \caption{Architecture of the ResNet-50.}
    \label{fig:resnet_expert}
\end{figure}

ResNet-50 leverages residual learning through bottleneck residual blocks with shortcut connections to facilitate deep feature extraction~\citep{He2015}. Each block combines $1 \times 1$ convolutions for channel compression and restoration with a central $3 \times 3$ convolution for spatial feature extraction, as illustrated in Figure~\ref{fig:resnet_expert}. This design enables the network to capture hierarchical visual representations efficiently. For an input image of size $224 \times 224$, the extracted feature map is denoted as $F_{RES} \in \mathbb{R}^{7 \times 7 \times 2048}$.

\subsubsection{Swin-Tiny}

Swin-Tiny is a hierarchical Vision Transformer that captures visual representations using localized self-attention with linear computational complexity~\citep{Liu2021}. The architecture employs window-based self-attention with a shifted window mechanism to enable cross-window interaction. A hierarchical pyramid structure progressively reduces spatial resolution while increasing channel capacity through patch merging. For an input image of size $224 \times 224$, the extracted feature map is denoted as $F_{SWIN} \in \mathbb{R}^{7 \times 7 \times 768}$.

\begin{figure}[H]
    \centering
    \includegraphics[width=\textwidth]{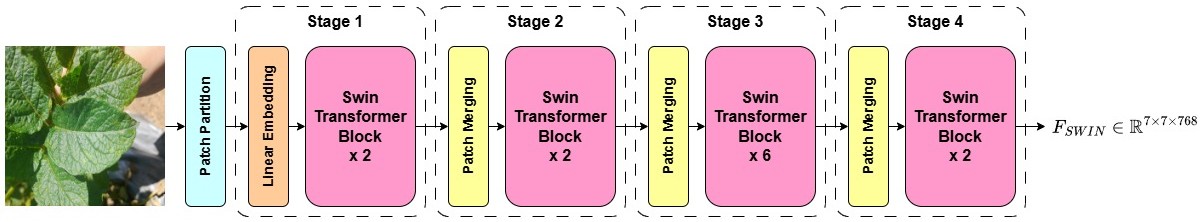}
    \caption{Architecture of the Swin-Tiny.}
    \label{fig:swin_expert}
\end{figure}

\subsection{Adaptive soft mixture of experts (MoE) framework}

\begin{figure}[H]
    \centering
    \includegraphics[width=\textwidth]{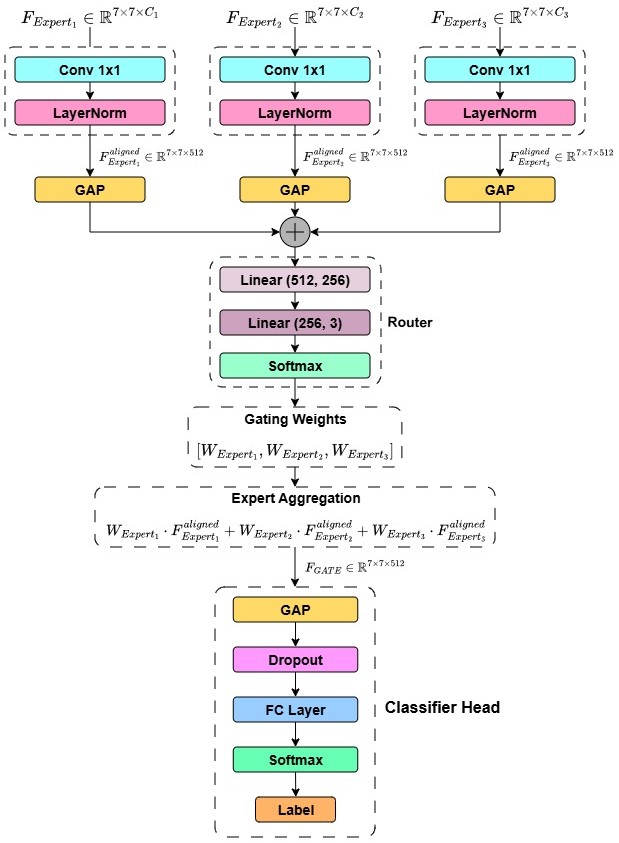}
    \caption{Detailed architecture of the proposed MoE framework.}
    \label{fig:moe}
\end{figure}

Building upon the heterogeneous feature representations extracted by the selected expert networks, the proposed framework adopts an adaptive soft MoE architecture to integrate complementary visual information. As illustrated in Figure~\ref{fig:moe}, the proposed framework employs an adaptive routing mechanism that dynamically reweights expert contributions based on the learned representation of each input.

Let the extracted feature maps from the three selected experts be denoted as $F_{Expert_1} \in \mathbb{R}^{7 \times 7 \times C_1}$, $F_{Expert_2} \in \mathbb{R}^{7 \times 7 \times C_2}$, and $F_{Expert_3} \in \mathbb{R}^{7 \times 7 \times C_3}$. Since heterogeneous backbones produce feature maps with different channel dimensions, a Feature Projection Module (FPM) is introduced to align them into a shared feature space. Specifically, each feature map is projected through a $1 \times 1$ convolution to a unified channel dimension of 512, providing lightweight dimensional alignment while preserving spatial structure. 
Layer Normalization (LayerNorm) is subsequently applied to normalize expert-specific feature statistics prior to fusion, providing more stable feature alignment across heterogeneous backbones and producing aligned feature representations denoted as $F^{aligned}_{Expert_1}$, $F^{aligned}_{Expert_2}$, and $F^{aligned}_{Expert_3} \in \mathbb{R}^{7 \times 7 \times 512}$.

After feature alignment, Global Average Pooling (GAP) is applied to each expert representation to obtain global feature vectors. These vectors are then combined through element-wise summation ($\oplus$) to construct a compact shared descriptor for routing while avoiding the dimensional expansion and additional parameter overhead associated with concatenation-based fusion. The resulting descriptor is processed by a routing network implemented as a two-layer Multi-Layer Perceptron (MLP) with a Linear ($512 \rightarrow 256 \rightarrow 3$) structure, followed by a Softmax activation to generate continuous gating weights $W_{Expert_1}$, $W_{Expert_2}$, and $W_{Expert_3}$. This enables the framework to adaptively modulate the contribution of each expert for individual inputs.
The final fused feature map is computed as:

\vspace{-0.5cm}

\[
F_{GATE} = W_{Expert_1} \cdot F^{aligned}_{Expert_1} + W_{Expert_2} \cdot F^{aligned}_{Expert_2} + W_{Expert_3} \cdot F^{aligned}_{Expert_3}
\]

\noindent where $W_{Expert_i} \in [0,1]$ and $\sum_{i=1}^{3} W_{Expert_i} = 1$, with $W_{Expert_i}$ representing the gating weight of the $i$-th expert. This soft aggregation strategy allows the model to adaptively combine expert outputs according to the input characteristics.

The fused feature map $F_{GATE} \in \mathbb{R}^{7 \times 7 \times 512}$ is subsequently processed by the classifier head. Specifically, GAP is applied to obtain a 512-dimensional feature vector, followed by Dropout for regularization. The resulting representation is then passed through a single fully connected layer to generate the final class prediction.

\subsection{Two-stage refinement training strategy}\label{sec:training_strategy}

The proposed MoE framework is optimized using a two-stage refinement training strategy. In the first stage, all backbone networks are initialized with ImageNet-pretrained weights and jointly fine-tuned together with the routing module and classification head. This stage enables the framework to adapt the pre-trained feature representations to the target domain while learning input-dependent routing behavior.

In the second stage, training resumes from the best-performing checkpoint obtained in the first stage, using a reduced learning rate for further refinement. Starting from an already converged solution allows more controlled parameter updates, enabling finer adjustment of both the learned feature representations and the gating mechanism. This refinement stage helps improve optimization stability and may help the optimization converge to a better solution.

\subsection{Gradient-weighted Class Activation Mapping (Grad-CAM)}

Grad-CAM~\citep{Selvaraju2019} is a gradient-based visualization technique used to interpret deep learning models by leveraging the gradients of a target class flowing back to the final convolutional layer to generate class-discriminative localization maps. These heatmaps highlight the spatial regions that contribute most strongly to the model's prediction, thereby providing visual insight into the model's decision-making behavior.

In this study, Grad-CAM is employed to qualitatively analyze the attention behavior of the proposed MoE framework and the specialization characteristics of individual experts. Specifically, Grad-CAM generates attention heatmaps that visualize the spatial regions emphasized by each backbone during prediction. This facilitates comparison of attention patterns across different architectures and provides qualitative insight into how complementary feature representations are integrated within the proposed framework.

\section{Experiments}\label{sec:experiments}

\subsection{Data description}

To evaluate the effectiveness and generalization capability of the proposed MoE framework, experiments were conducted on three leaf disease datasets. The potato dataset serves as the primary benchmark, while the durian and sesame datasets are used to assess performance consistency across different crop conditions and data distributions. Detailed dataset descriptions are provided in the following subsections.

\subsubsection{Potato leaf disease dataset}

\begin{figure}[H]
    \centering
    
    \hspace{0.14\linewidth}
    \begin{subfigure}{0.22\linewidth}
        \centering
        \includegraphics[width=\linewidth]{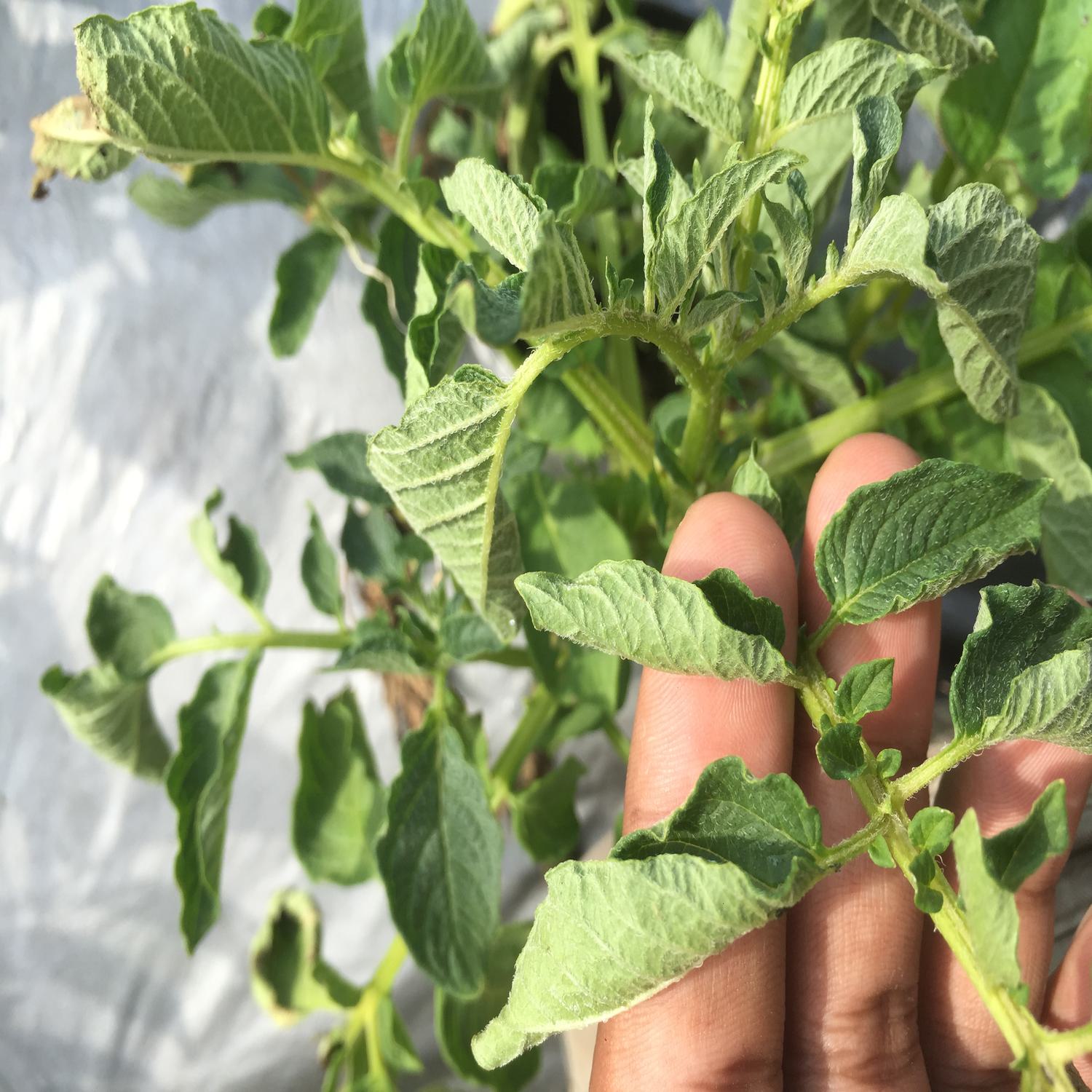}
        \caption{Bacteria}
    \end{subfigure}
    \hfill
    \begin{subfigure}{0.22\linewidth}
        \centering
        \includegraphics[width=\linewidth]{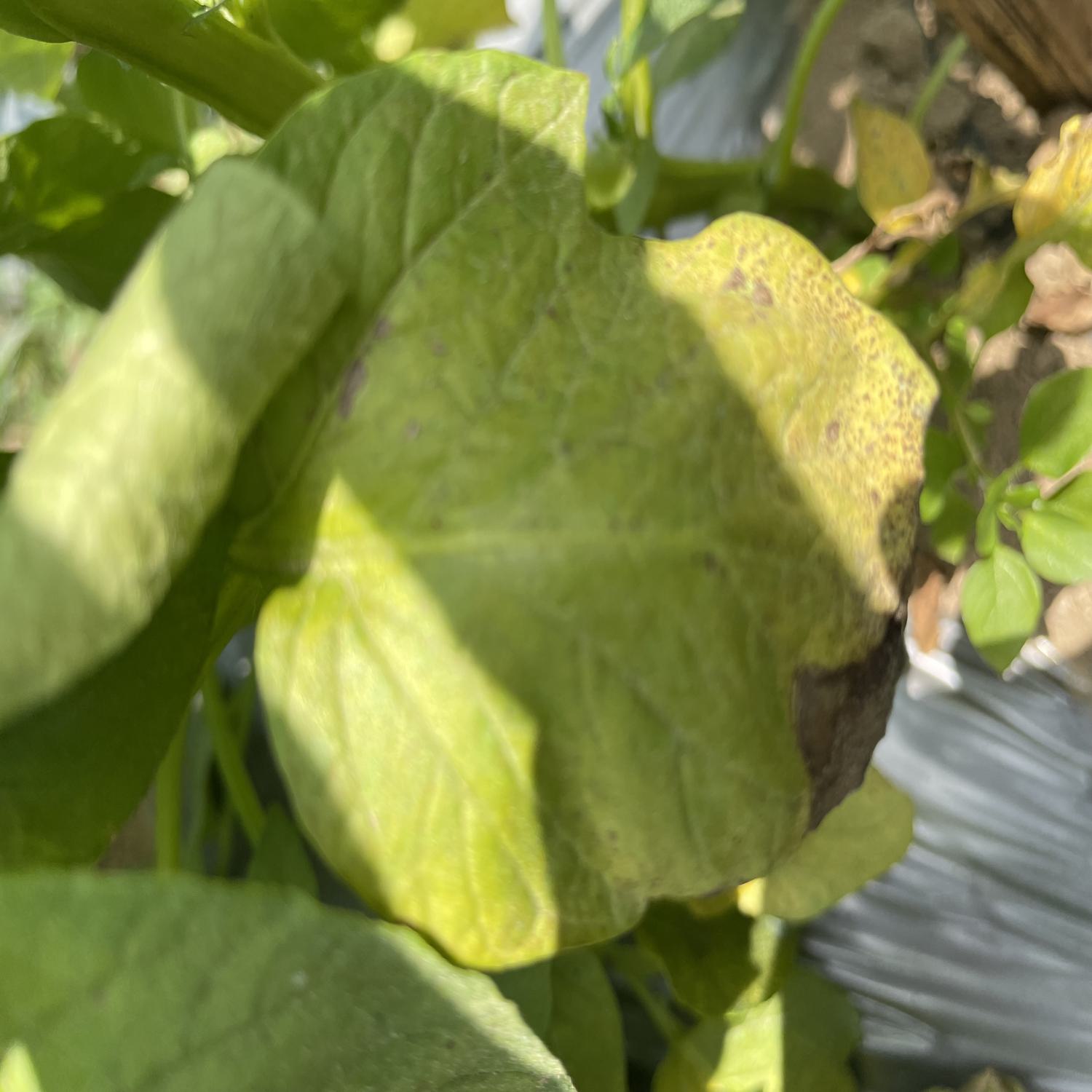}
        \caption{Fungi}
    \end{subfigure}
    \hfill
    \begin{subfigure}{0.22\linewidth}
        \centering
        \includegraphics[width=\linewidth]{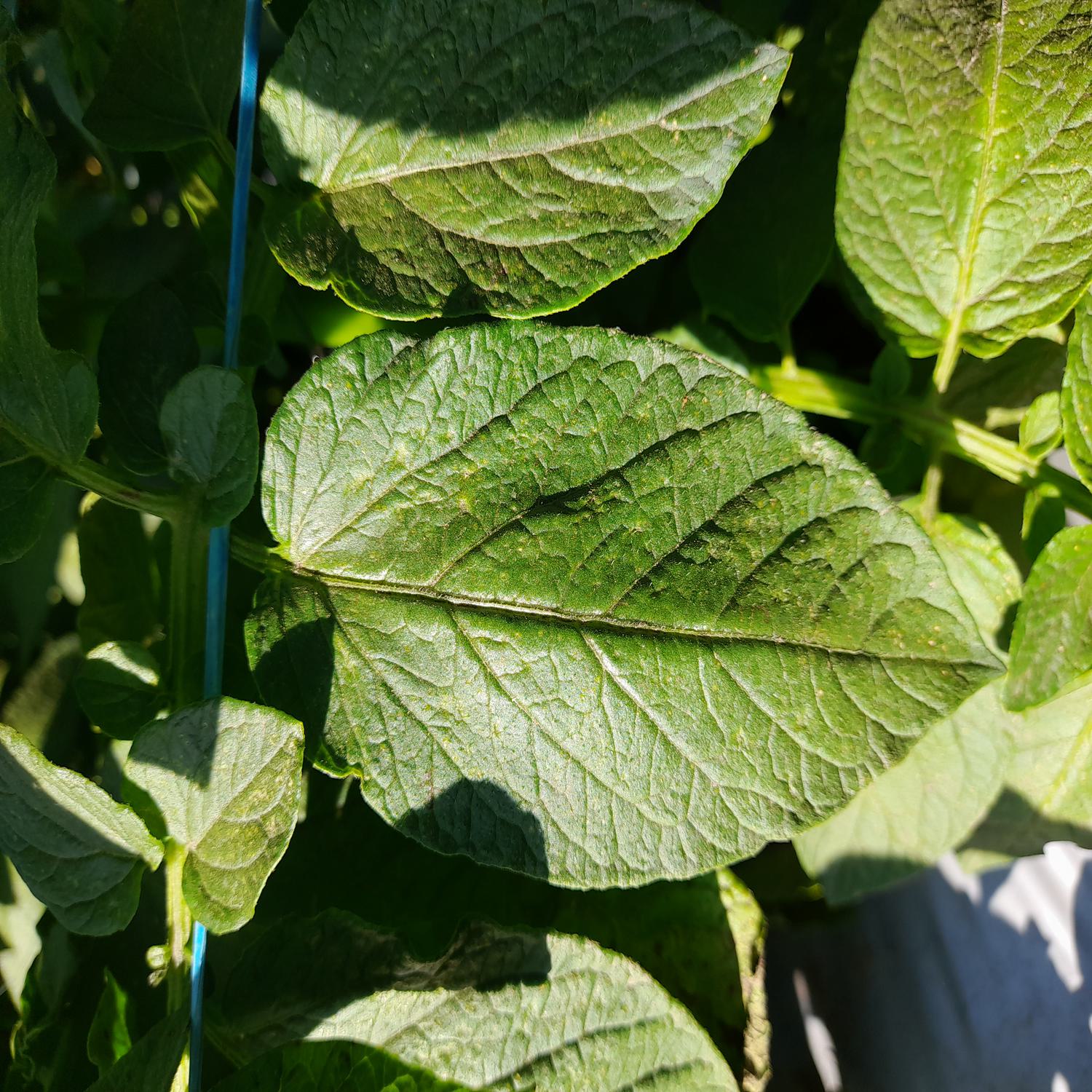}
        \caption{Healthy}
    \end{subfigure}
    \hspace{0.14\linewidth}
    
    \begin{subfigure}{0.22\linewidth}
        \centering
        \includegraphics[width=\linewidth]{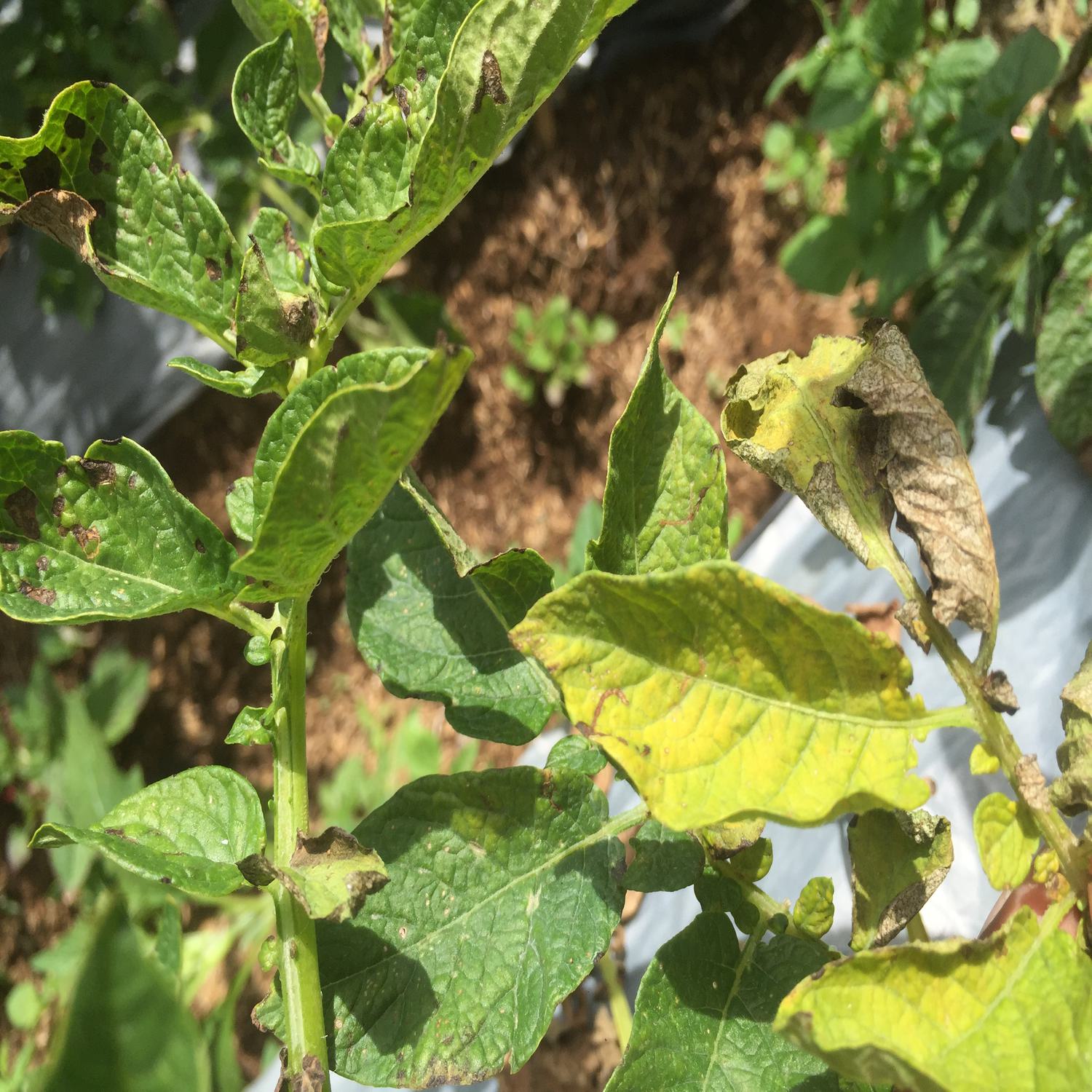}
        \caption{Nematode}
    \end{subfigure}
    \hfill
    \begin{subfigure}{0.22\linewidth}
        \centering
        \includegraphics[width=\linewidth]{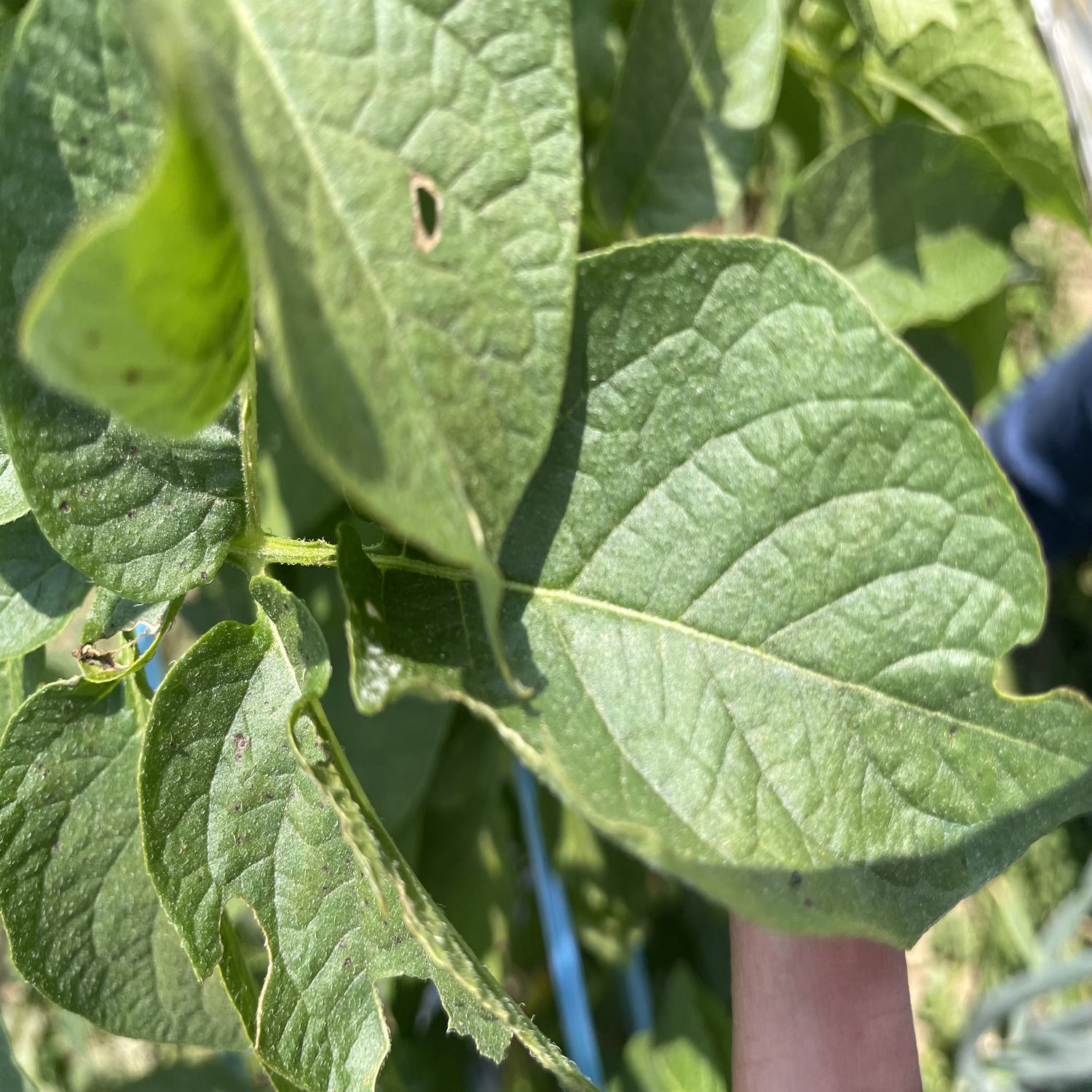}
        \caption{Pest}
    \end{subfigure}
    \hfill
    \begin{subfigure}{0.22\linewidth}
        \centering
        \includegraphics[width=\linewidth]{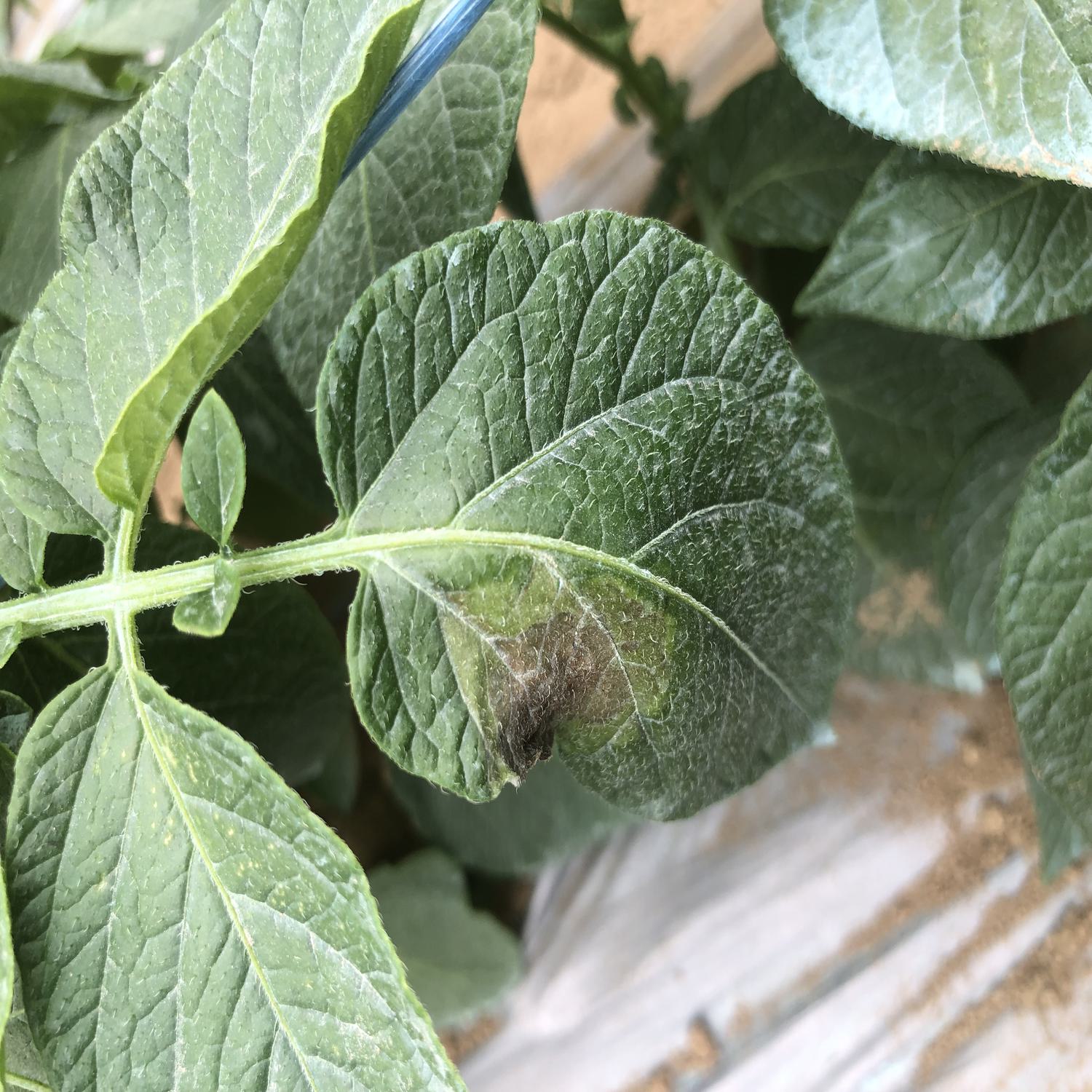}
        \caption{Phytophthora}
    \end{subfigure}
    \hfill
    \begin{subfigure}{0.22\linewidth}
        \centering
        \includegraphics[width=\linewidth]{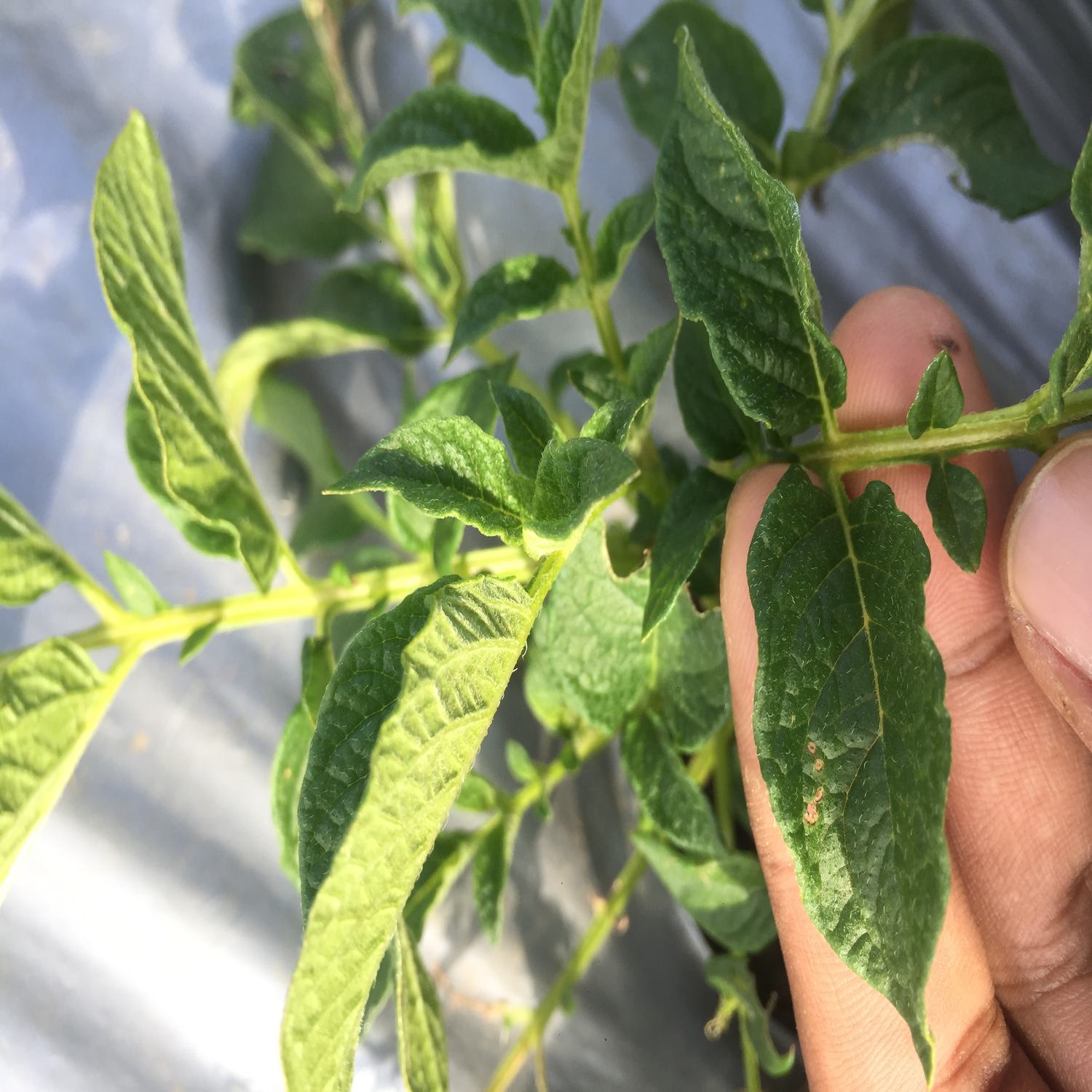}
        \caption{Virus}
    \end{subfigure}

    \vspace{-0.5em}
    \caption{Sample images representing different classes in the potato leaf disease dataset.}
    \label{fig:potato_disease_samples}
\end{figure}

The potato leaf disease dataset consists of 3,076 images categorized into seven classes: Bacteria (569), Fungi (748), Healthy (201), Nematode (68), Pest (611), Phytophthora (347), and Virus (532)~\citep{Shabrina2024}. The images were collected from various agricultural fields in Central Java, Indonesia, using a diverse range of smartphone cameras (including models from Apple, Samsung, Xiaomi, and Vivo) with resolutions ranging from 8 to 50 megapixels. Unlike controlled laboratory settings, the dataset captures natural real-world variations, including cluttered backgrounds, varying lighting conditions, and imaging distances ranging from 5 to 15 cm. All images were standardized to a fixed size of $1500 \times 1500$ pixels. The dataset was split using a stratified sampling strategy with a ratio of 81\%, 9\%, and 10\% for training, validation, and testing sets, respectively, as in Shabrina et al.~\cite{Shabrina2024}, and consistent splits across all experiments with a random seed of 42. The detailed distribution is provided in Table~\ref{tab:potato_distribution}. Representative samples of each class are illustrated in Figure~\ref{fig:potato_disease_samples}.

\begin{table}[H]
\centering
\caption{Distribution of samples across classes in the training, validation, and test sets.}
\label{tab:potato_distribution}
\begin{tabular}{|l | c c c | c|}
\hline
\textbf{Classes} & \textbf{Train} & \textbf{Validation} & \textbf{Test} & \textbf{Total} \\
\hline
Bacteria    & 461  & 51  & 57  & 569  \\
Fungi       & 605  & 67  & 76  & 748  \\
Healthy     & 163  & 18  & 20  & 201  \\
Nematode    & 55   & 6   & 7   & 68   \\
Pest        & 494  & 55  & 62  & 611  \\
Phytophthora & 281  & 31  & 35  & 347  \\
Virus       & 430  & 48  & 54  & 532  \\
\hline
\textbf{Total} & \textbf{2489} & \textbf{276} & \textbf{311} & \textbf{3076} \\
\hline
\end{tabular}
\end{table}

\subsubsection{Durian disease dataset}

\begin{figure}[H]
    \centering
    
    \begin{subfigure}{0.18\linewidth}
        \centering
        \includegraphics[width=\linewidth]{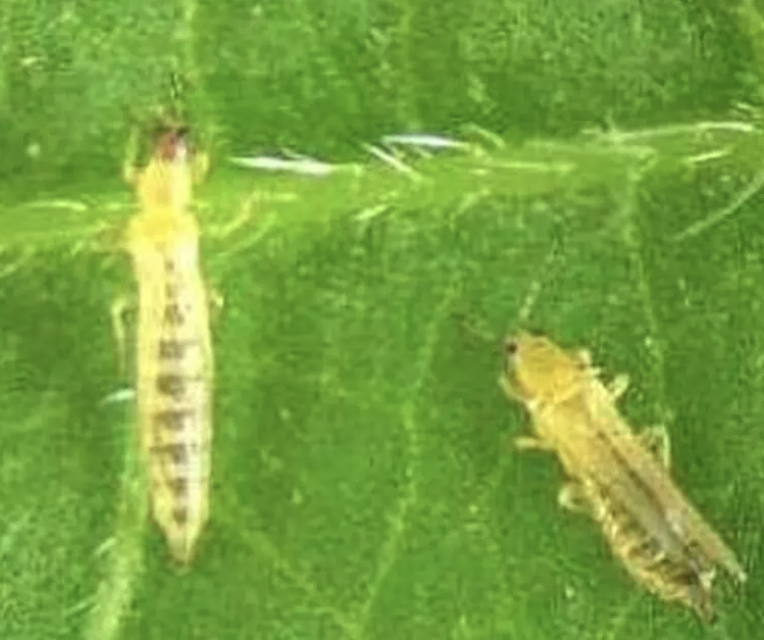}
        \caption{Thrips}
    \end{subfigure}
    \hfill
    \begin{subfigure}{0.18\linewidth}
        \centering
        \includegraphics[width=\linewidth]{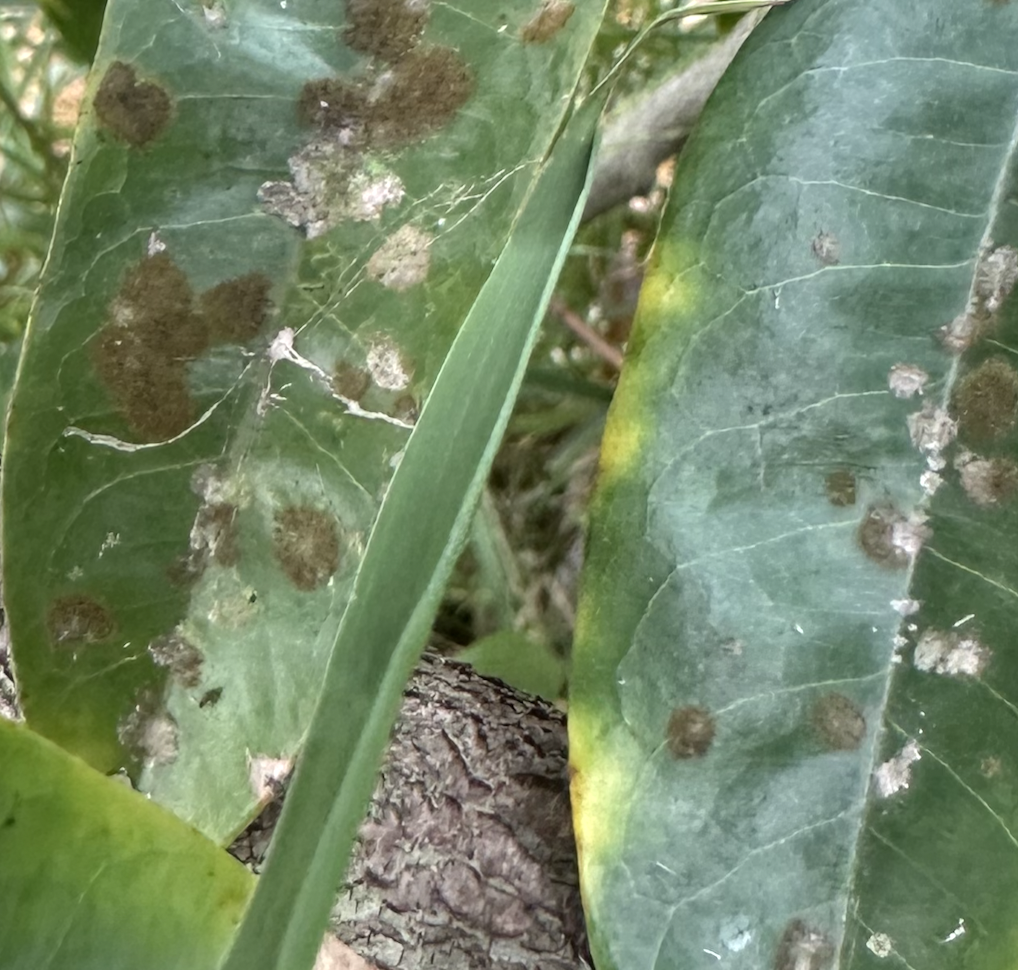}
        \caption{Stem blight}
    \end{subfigure}
    \hfill
    \begin{subfigure}{0.18\linewidth}
        \centering
        \includegraphics[width=\linewidth]{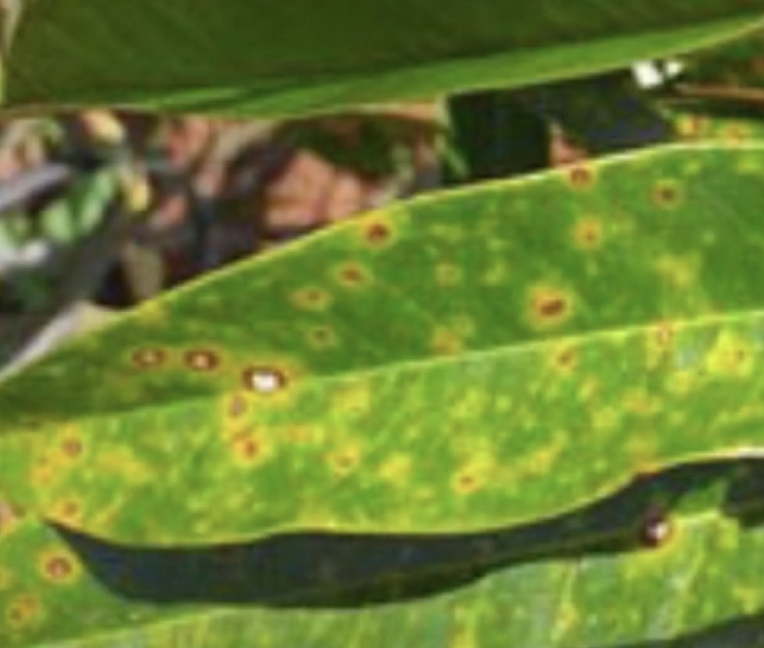}
        \caption{Canker}
    \end{subfigure}
    \hfill
    \begin{subfigure}{0.18\linewidth}
        \centering
        \includegraphics[width=\linewidth]{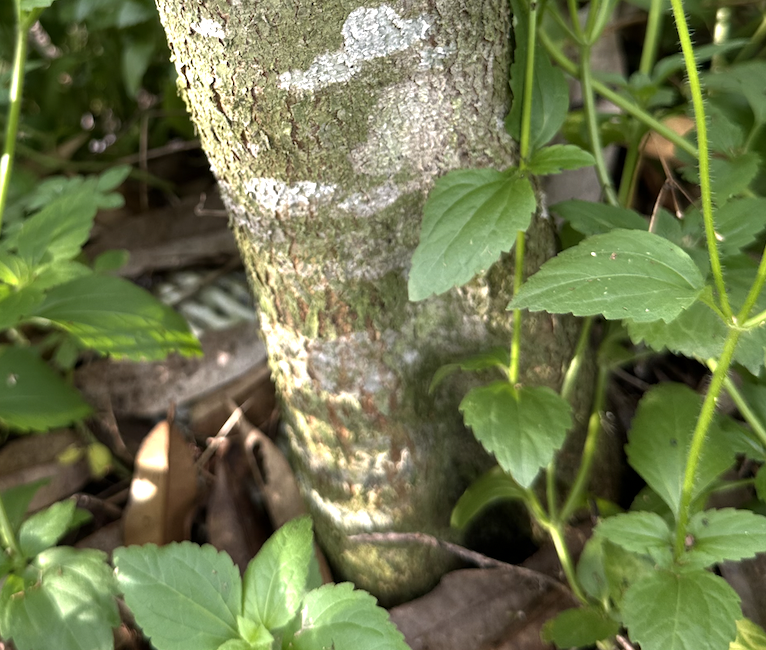}
        \caption{Pink disease}
    \end{subfigure}
    \hfill
    \begin{subfigure}{0.18\linewidth}
        \centering
        \includegraphics[width=\linewidth]{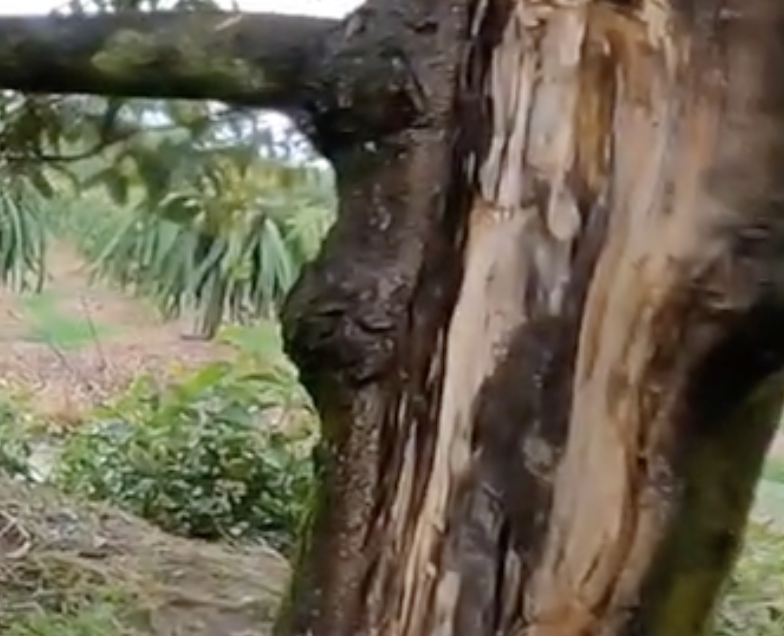}
        \caption{Gummosis}
    \end{subfigure}

    \begin{subfigure}{0.18\linewidth}
        \centering
        \includegraphics[width=\linewidth]{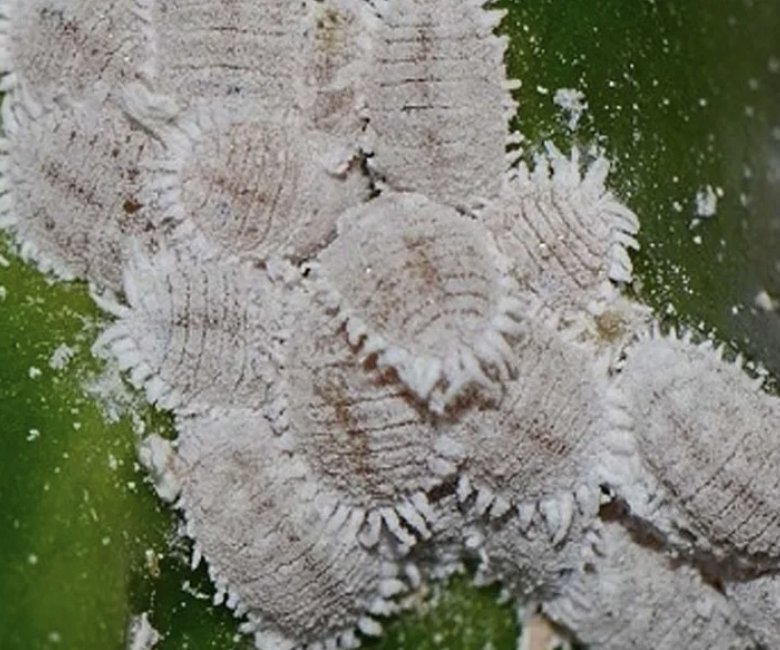}
        \caption{Mealybug}
    \end{subfigure}
    \hfill
    \begin{subfigure}{0.18\linewidth}
        \centering
        \includegraphics[width=\linewidth]{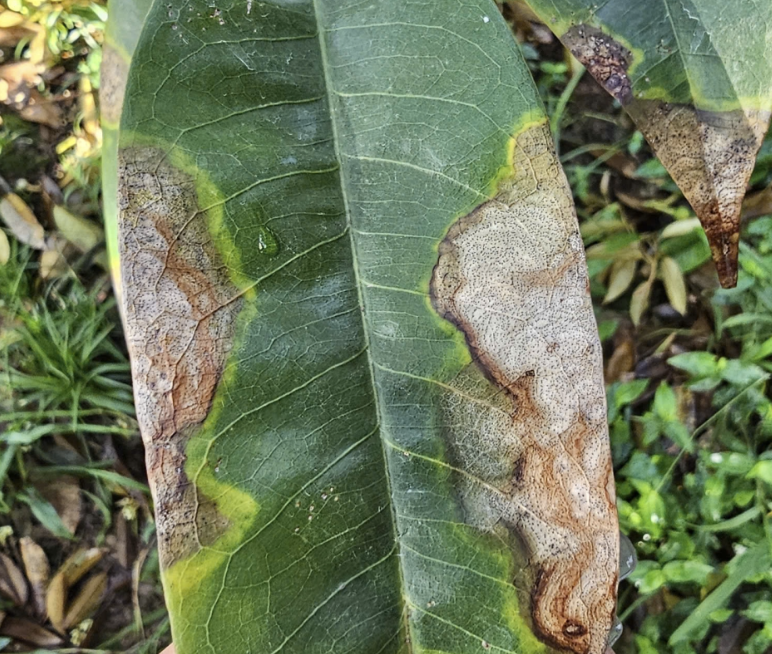}
        \caption{Anthracnose}
    \end{subfigure}
    \hfill
    \begin{subfigure}{0.18\linewidth}
        \centering
        \includegraphics[width=\linewidth]{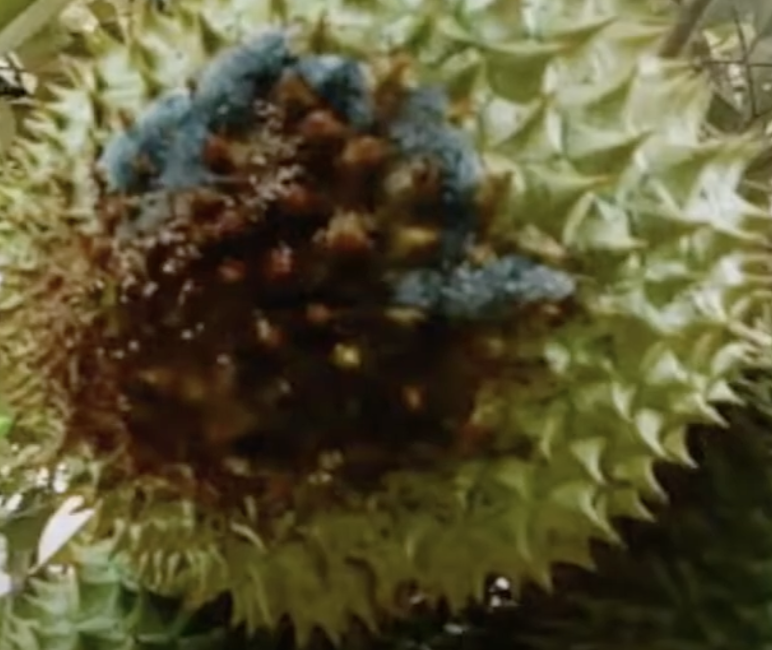}
        \caption{Fruit rot}
    \end{subfigure}
    \hfill
    \begin{subfigure}{0.18\linewidth}
        \centering
        \includegraphics[width=\linewidth]{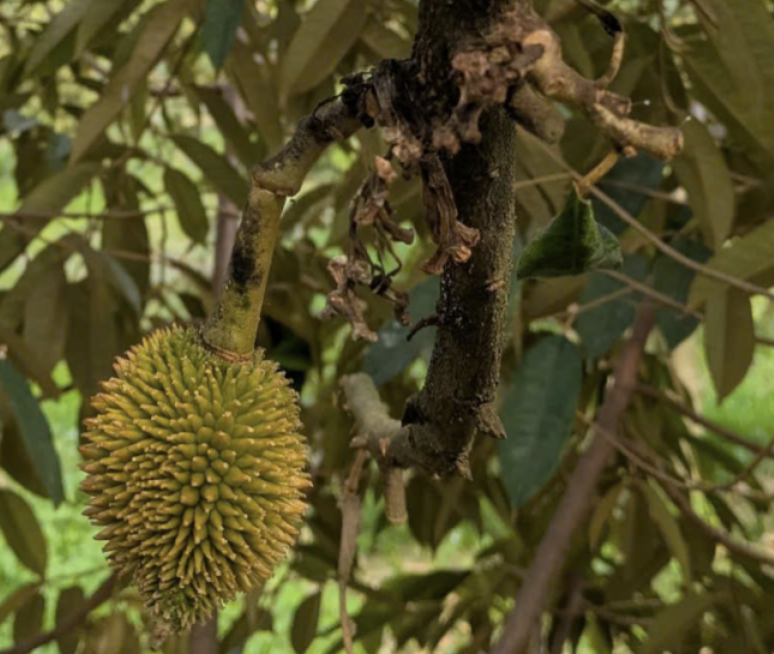}
        \caption{Sooty mold}
    \end{subfigure}
    \hfill
    \begin{subfigure}{0.18\linewidth}
        \centering
        \includegraphics[width=\linewidth]{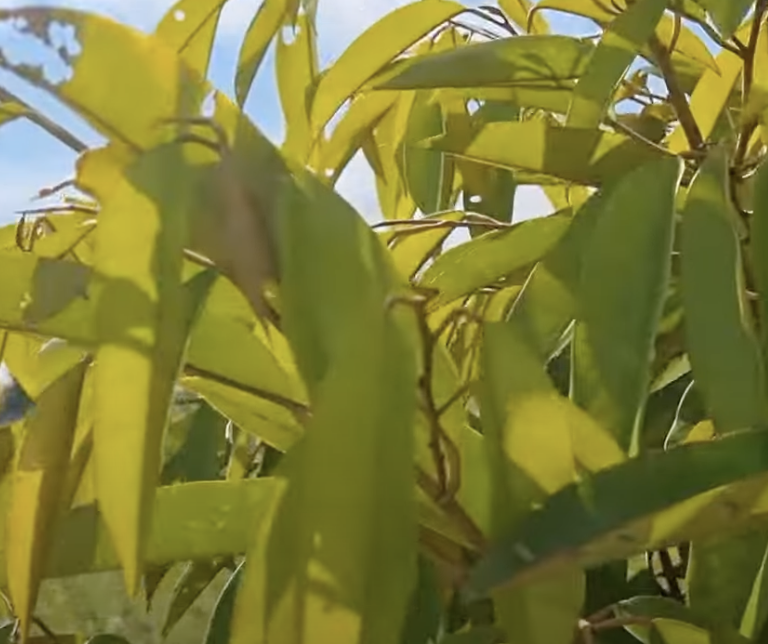}
        \caption{Yellow leaf}
    \end{subfigure}
    
    \vspace{-0.5em}
    \caption{Sample images representing different classes in the durian disease dataset.}
    \label{fig:durian_disease_samples}
\end{figure}

The durian disease dataset consists of 5,451 images categorized into ten classes: Thrips disease (537), Stem blight (554), Canker disease (557), Pink disease (549), Stem cracking and gummosis (560), Mealybug infestation (536), Anthracnose (549), Fruit rot (528), Sooty mold (541), and Yellow leaf disease (540)~\citep{Nguyen2025}. The images were collected from a family-owned durian orchard and four neighboring farms in Vinh Long, Vietnam, under real-field conditions. All images were captured using an iPhone 14 under natural lighting. Image resolutions range from 150 to 1824 pixels in width and from 150 to 1452 pixels in height. The dataset contains various real-world challenges, including motion blur, uneven illumination, diverse viewing angles, and complex backgrounds. The dataset was split using stratified sampling with a ratio of 64\%, 16\%, and 20\% for training, validation, and testing, respectively, with a fixed random seed of 42 to ensure consistent splits across all experiments. Representative samples of each disease class are illustrated in Figure~\ref{fig:durian_disease_samples}.

\subsubsection{Sesame leaf disease dataset}

\begin{figure}[H]
    \centering
    
    \begin{subfigure}[t]{0.23\linewidth}
        \centering
        \includegraphics[width=\linewidth]{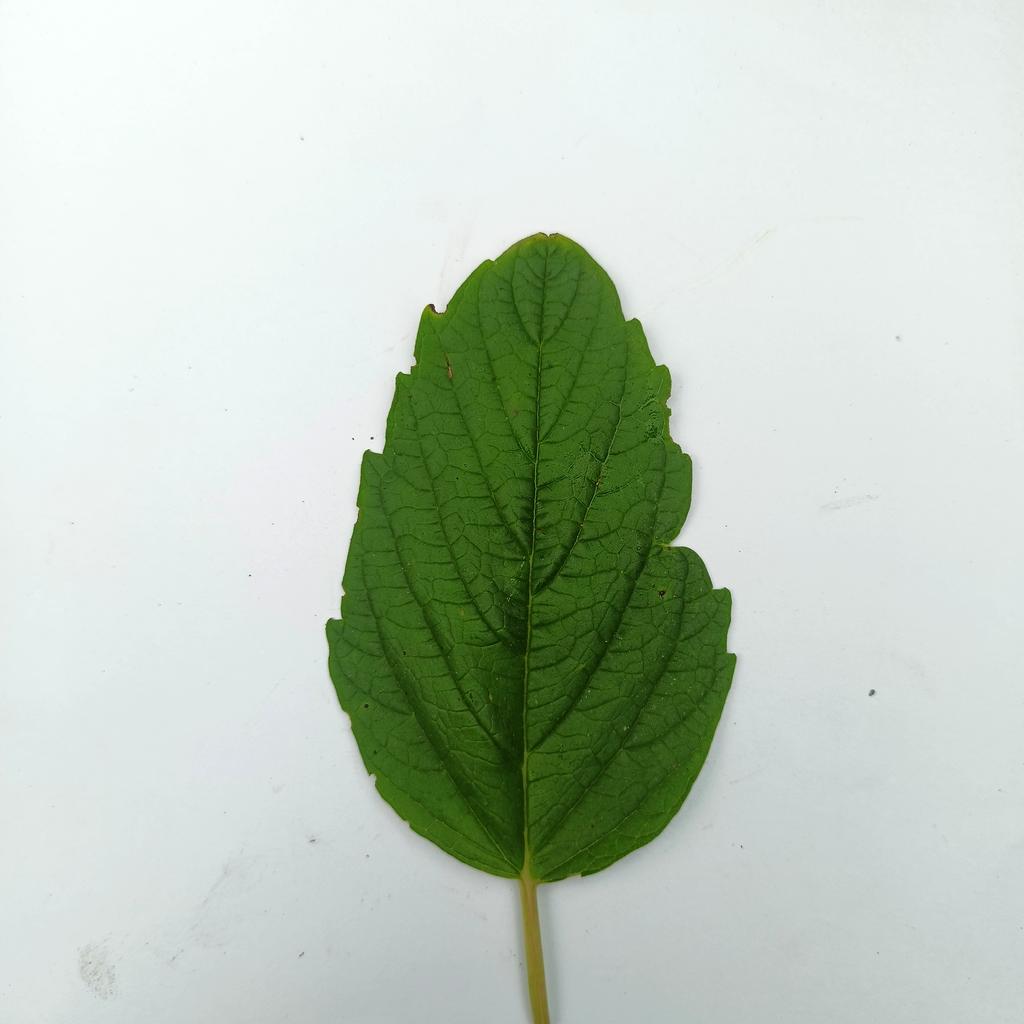}
        \caption{Healthy Leaf}
    \end{subfigure}
    \hfill
    \begin{subfigure}[t]{0.23\linewidth}
        \centering
        \includegraphics[width=\linewidth]{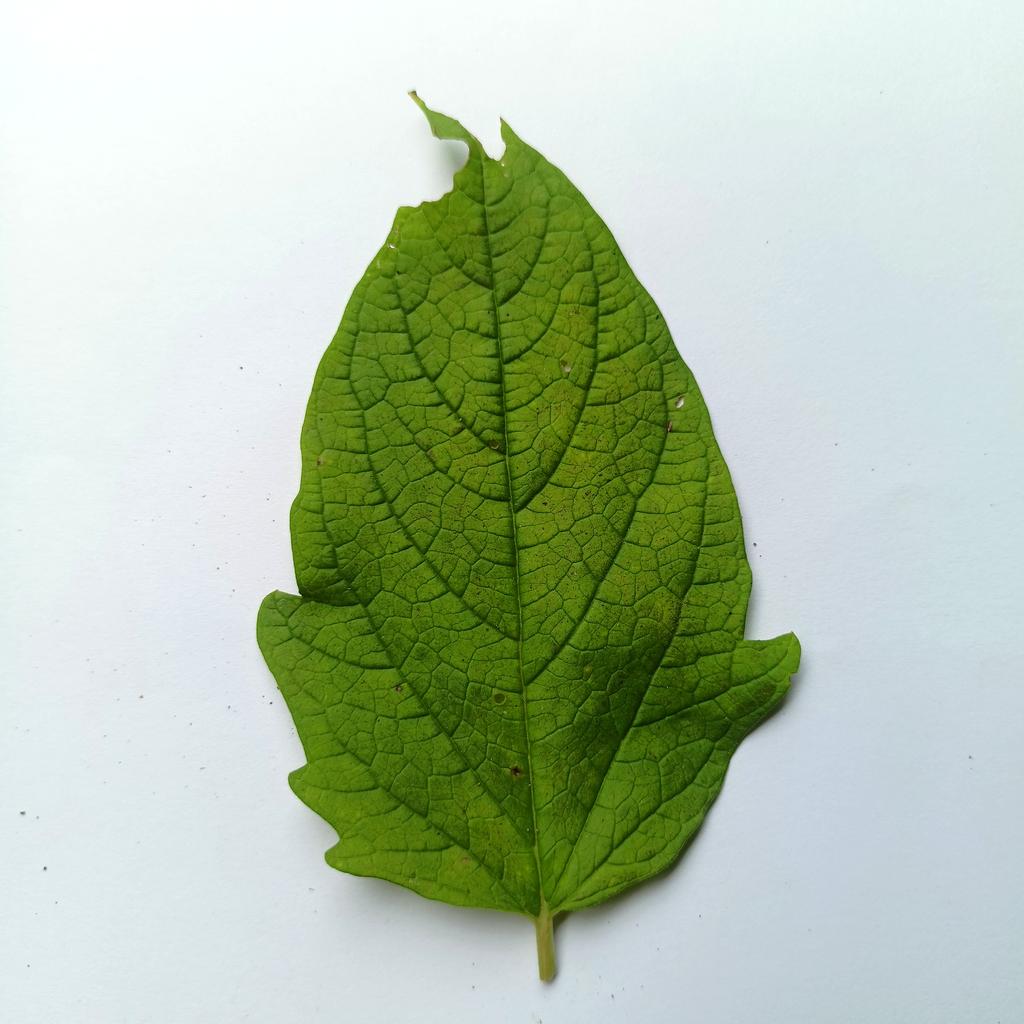}
        \caption{Insect Leaf Damage}
    \end{subfigure}
    \hfill
    \begin{subfigure}[t]{0.23\linewidth}
        \centering
        \includegraphics[width=\linewidth]{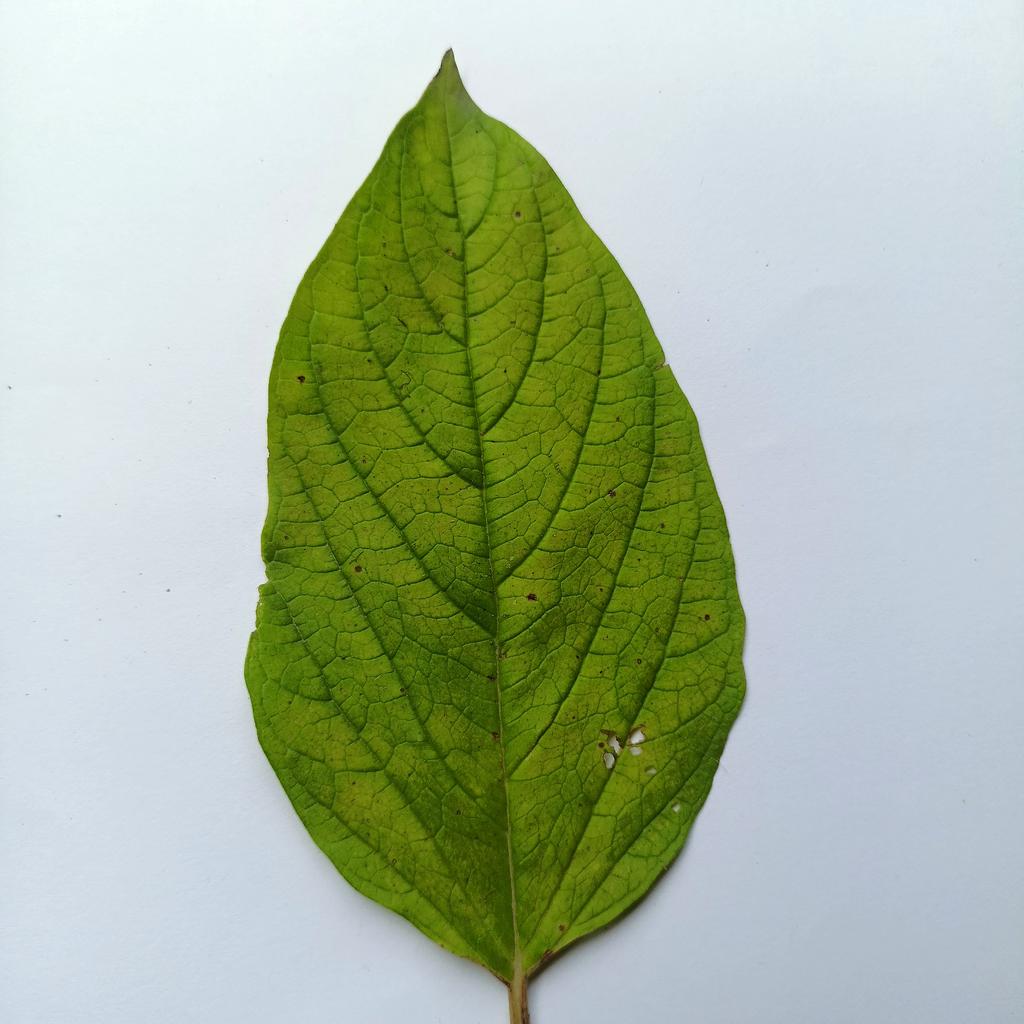}
        \caption{Leaf Spot Disease}
    \end{subfigure}
    \hfill
    \begin{subfigure}[t]{0.23\linewidth}
        \centering
        \includegraphics[width=\linewidth]{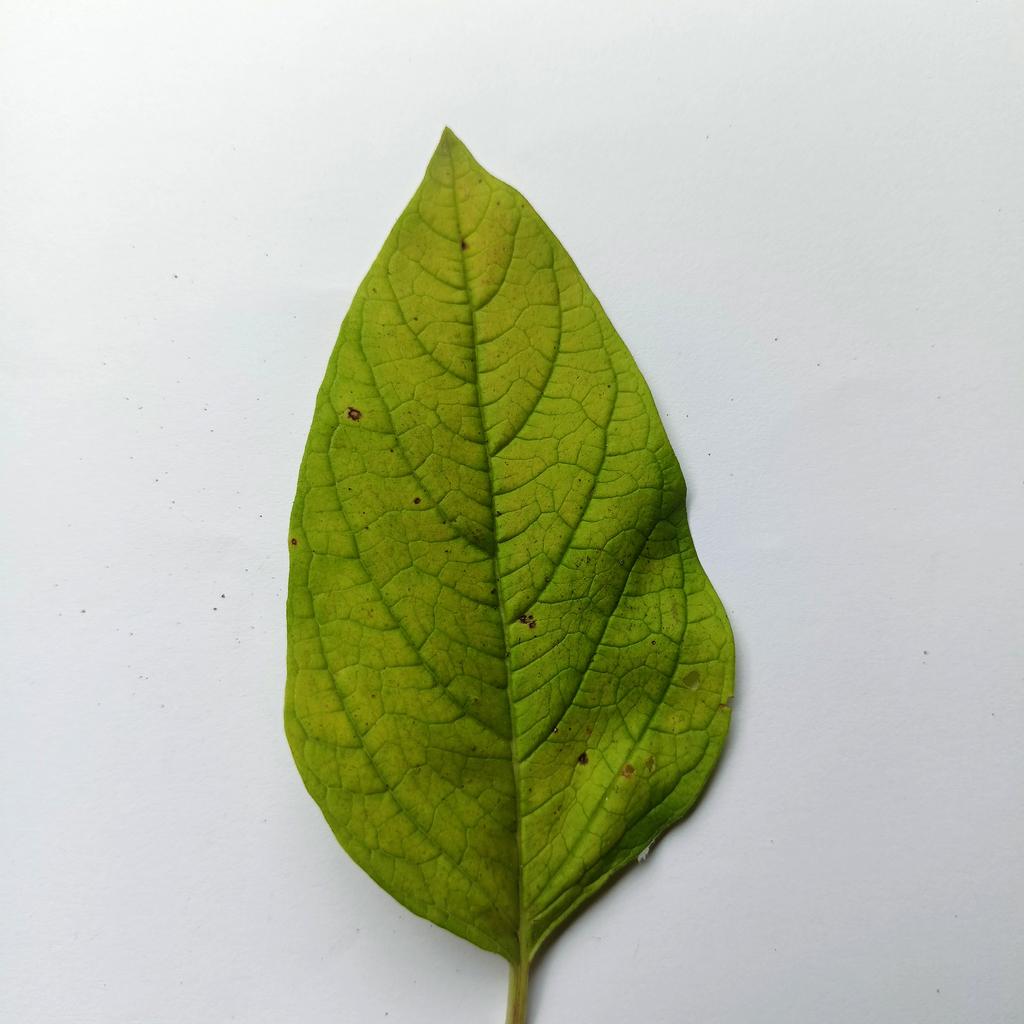}
        \caption{Yellowing Leaf Syndrome}
    \end{subfigure}
    
    \vspace{-0.5em}
    \caption{Sample images representing different classes in the sesame leaf disease dataset.}
    \label{fig:sesame_disease_samples}
\end{figure}

The sesame leaf disease dataset consists of 3,540 images categorized into four classes: Healthy Leaf (1,335), Insect Leaf Damage (724), Leaf Spot Disease (587), and Yellowing Leaf Syndrome (894)~\citep{Rahman2025}. The images were collected from sesame cultivation fields in Pabna District, Bangladesh, and captured under natural field conditions using high-resolution smartphone cameras. Unlike controlled laboratory datasets, the images reflect real-world variations such as illumination changes and complex background clutter. All images were standardized to a fixed size of $1024 \times 1024$ pixels. The dataset was split using stratified sampling with a ratio of 64\%, 16\%, and 20\% for training, validation, and testing, respectively, with a fixed random seed of 42 to ensure consistent splits across all experiments. Representative samples of each class are illustrated in Figure~\ref{fig:sesame_disease_samples}.

\subsection{Experimental setting}

All experiments were conducted on the Kaggle platform using two NVIDIA Tesla T4 GPUs, and the proposed framework was implemented in PyTorch. All backbone networks were initialized with \texttt{ImageNet} pre-trained weights, and input images were resized to $224 \times 224$ pixels to match the backbone input requirements. Model optimization followed the two-stage refinement training strategy described in Section~\ref{sec:training_strategy}. Detailed preprocessing, augmentation, and dataset-specific training configurations are summarized in Table~\ref{tab:configurations}.

\begin{table}[H]
\centering
\caption{Detailed experimental and two-stage refinement training configurations for each dataset.}
\label{tab:configurations}
\footnotesize
\renewcommand{\arraystretch}{1.5}
\resizebox{\textwidth}{!}{
\begin{tabular}{|p{1.2cm}|p{6.8cm}|p{2.8cm}|p{2.8cm}|}
\hline
\textbf{Dataset} & \textbf{Preprocessing \& Augmentation} & \textbf{Phase 1 (Fine-tuning)} & \textbf{Phase 2 (Refinement)} \\ \hline

\textbf{Potato leaf disease} & 
\begin{tabular}[t]{@{}l@{}} 
Color jitter (brightness=0.15)\\
Random horizontal flip ($p=0.5$)\\
Random rotation ($\pm20^\circ$)\\
Random affine (translation=0.2)\\
\end{tabular} & 
\begin{tabular}[t]{@{}l@{}}
Epochs: 100 \\
Batch size: 16 \\
Optimizer: Adam\\
LR: $1 \times 10^{-4}$\\
Dropout: 0.3 \\
Patience: 20
\end{tabular} & 
\begin{tabular}[t]{@{}l@{}}
Epochs: 20 \\
Batch size: 16 \\
Optimizer: Adam\\
LR: $5 \times 10^{-6}$\\
Dropout: 0.3 \\
Patience: 5
\end{tabular} \\ \hline

\textbf{Durian leaf disease} & 
\begin{tabular}[t]{@{}l@{}} 
Color jitter (brightness=0.15, contrast=0.05)\\
Random horizontal flip ($p=0.5$)\\
Random rotation ($\pm15^\circ$)\\
Gaussian blur (kernel=3, $\sigma=0.1$–$1.0$)\\
\end{tabular} & 
\begin{tabular}[t]{@{}l@{}}
Epochs: 100 \\
Batch size: 32 \\
Optimizer: Adam\\
LR: $1 \times 10^{-4}$\\
Dropout: 0.1 \\
Patience: 20
\end{tabular} & 
\begin{tabular}[t]{@{}l@{}}
Epochs: 20 \\
Batch size: 32 \\
Optimizer: Adam\\
LR: $3 \times 10^{-7}$\\
Dropout: 0.1 \\
Patience: 5
\end{tabular} \\ \hline

\textbf{Sesame leaf disease} & 
\begin{tabular}[t]{@{}l@{}} 
Color jitter (brightness=0.15, contrast=0.05)\\
Random horizontal flip ($p=0.5$)\\
Random rotation ($\pm15^\circ$)\\
Gaussian blur (kernel=3, $\sigma=0.1$–$1.0$)\\
\end{tabular} & 
\begin{tabular}[t]{@{}l@{}}
Epochs: 100 \\
Batch size: 32 \\
Optimizer: Adam\\
LR: $1 \times 10^{-4}$\\
Dropout: 0.5 \\
Patience: 20
\end{tabular} & 
\begin{tabular}[t]{@{}l@{}}
Epochs: 20 \\
Batch size: 32 \\
Optimizer: Adam\\
LR: $7 \times 10^{-7}$\\
Dropout: 0.5 \\
Patience: 5
\end{tabular} \\ \hline

\end{tabular}
}
\end{table}

\section{Results and discussion}\label{sec:results_discussion}

This section presents the experimental evaluation of the proposed MoE framework, including performance analysis on the primary potato leaf disease dataset, internal mechanism analysis, cross-dataset validation, comparison with previous studies, and discussion of limitations.

\subsection{Results on the potato leaf disease dataset}

\subsubsection{Performance of individual models}~\label{sec:pre_trained_models_on_potato}

To establish baseline performance for the potato leaf disease classification task and identify suitable experts for the proposed MoE framework, five representative pre-trained models were evaluated: MobileNet-V2, EfficientNet-B0, DenseNet-121, ResNet-50, and Swin-Tiny. These models were selected to cover diverse architectural paradigms, including lightweight convolutional networks (MobileNet-V2), residual CNNs (ResNet-50), densely connected networks (DenseNet-121), compound-scaled efficient CNNs (EfficientNet-B0), and transformer-based architectures with hierarchical self-attention (Swin-Tiny). This architectural diversity enables a broad assessment of complementary representational characteristics for expert selection. While expert selection was primarily guided by validation performance, test performance was additionally evaluated to assess the generalization capability of the candidate architectures on unseen data.

\begin{table}[H]
\caption{Classification performance of pre-trained models on the potato leaf disease dataset across the validation and test sets (\%).}
\label{tab:baseline_potato_results}
\footnotesize
\centering
\renewcommand{\arraystretch}{1.4}
\resizebox{\linewidth}{!}{
\begin{tabular}{|l|c|c|c|c|c|c|c|c|}
\hline
\multirow{2}{*}{\textbf{Model}} 
& \multicolumn{4}{c|}{\textbf{Validation set}} 
& \multicolumn{4}{c|}{\textbf{Test set}} \\
\cline{2-9}
& \textbf{Accuracy} & \textbf{Precision} & \textbf{Recall} & \textbf{F1-score}
& \textbf{Accuracy} & \textbf{Precision} & \textbf{Recall} & \textbf{F1-score} \\
\hline
MobileNet-V2    
& 85.57 & 86.93 & 80.71 & 82.55
& 85.21 & 87.76 & 77.58 & 79.50 \\
\hline
EfficientNet-B0 
& 83.33 & 84.80 & 82.84 & 83.49
& 86.50 & 87.22 & 84.15 & 85.37 \\
\hline
DenseNet-121    
& \textbf{87.68} & 85.32 & 85.38 & 84.80
& 87.78 & 84.65 & 81.93 & 82.72 \\
\hline
ResNet-50       
& 84.78 & 83.36 & 82.61 & 82.93
& 85.21 & 85.99 & 80.44 & 82.42 \\
\hline
Swin-Tiny       
& 86.23 & \textbf{87.93} & \textbf{85.53} & \textbf{86.23}
& \textbf{89.39} & \textbf{91.65} & \textbf{85.77} & \textbf{87.59} \\
\hline
\end{tabular}
}
\end{table}

Table~\ref{tab:baseline_potato_results} summarizes the performance of the candidate architectures. Since expert selection was primarily based on validation performance, the validation results are analyzed first. Given the pronounced class imbalance in the potato dataset, recall and F1-score are particularly important for identifying robust expert candidates. Swin-Tiny achieves the strongest balanced validation performance, obtaining the highest precision (87.93\%), recall (85.53\%), and F1-score (86.23\%), indicating effective contextual modeling under complex background conditions. DenseNet-121 attains the highest validation accuracy (87.68\%) while maintaining competitive recall (85.38\%) and F1-score (84.80\%), reflecting strong fine-grained local feature extraction. EfficientNet-B0 also demonstrates a relatively balanced performance profile, achieving a recall of 82.84\% and an F1-score of 83.49\%, suggesting its potential contribution through robust multi-scale feature representation. In contrast, the remaining architectures show comparatively weaker validation performance, particularly in recall and F1-score, indicating reduced robustness under class imbalance and high intra-class variability.

To further assess generalization capability, test performance was additionally examined as supportive evidence. The overall trends remain largely consistent with the validation results, with Swin-Tiny achieving the strongest generalization performance, obtaining the highest test accuracy (89.39\%) and F1-score (87.59\%). EfficientNet-B0 also maintains relatively balanced performance, achieving an F1-score of 85.37\%. In contrast, the remaining architectures exhibit weaker generalization, particularly in recall and F1-score, indicating reduced robustness under unseen disease variations, class imbalance, and high intra-class variability.

Based on the validation analysis, further supported by the test set generalization results, MobileNet-V2 and ResNet-50 were excluded due to comparatively weaker performance. DenseNet-121, EfficientNet-B0, and Swin-Tiny were therefore selected as expert candidates for the proposed MoE framework, given their complementary strengths in local, multi-scale, and contextual feature representation.

\subsubsection{Performance of the proposed MoE framework}

Table~\ref{tab:moe_potato_results} presents the classification performance of the MoE framework on the potato leaf disease dataset across the validation and test sets.

\begin{table}[H]
\caption{Classification performance of the proposed MoE framework on the potato leaf disease dataset across the validation and test sets (\%).}
\label{tab:moe_potato_results}
\footnotesize
\centering
\renewcommand{\arraystretch}{1.4}
\resizebox{\linewidth}{!}{
\begin{tabular}{|c|c|c|c|c|c|c|c|}
\hline
\multicolumn{4}{|c|}{\textbf{Validation set}} 
& \multicolumn{4}{c|}{\textbf{Test set}} \\
\hline
\textbf{Accuracy} & \textbf{Precision} & \textbf{Recall} & \textbf{F1-score}
& \textbf{Accuracy} & \textbf{Precision} & \textbf{Recall} & \textbf{F1-score} \\
\hline


\textbf{88.77} & \textbf{87.24} & \textbf{87.74} & \textbf{87.05}
& \textbf{92.28} & \textbf{93.97} & \textbf{91.68} & \textbf{92.62} \\
\hline

\end{tabular}
}
\end{table}

 The MoE framework demonstrates clear and consistent improvements over all individual baseline models, with the most substantial gains observed on the unseen test set. While test accuracy improves from 89.39\% to 92.28\%, corresponding to an absolute gain of 2.89\% over Swin-Tiny, the strongest standalone baseline, the improvements in recall and F1-score are considerably more pronounced. Specifically, recall increases from 85.77\% to 91.68\% (+5.91\%), while the F1-score improves from 87.59\% to 92.62\% (+5.03\%). Given the strong class imbalance in the potato dataset, where the largest class contains more than ten times the samples of the smallest class, these gains are particularly meaningful. The marked improvement in recall indicates a substantially stronger ability to correctly identify minority and difficult disease categories, while the corresponding F1-score gain confirms a more balanced trade-off between precision and recall.

These improvements can be attributed to the complementary integration of heterogeneous expert representations. DenseNet-121 contributes fine-grained local texture modeling, EfficientNet-B0 provides robust multi-scale feature extraction, and Swin-Tiny captures broader contextual dependencies through hierarchical self-attention. When operating independently, each architecture exhibits inherent limitations, particularly when handling imbalanced and visually diverse disease categories. Through the soft gating mechanism, the framework dynamically adjusts expert contributions according to input characteristics, effectively combining these strengths into a unified representation and yielding more balanced classification performance than any standalone model.

Consistent gains are also observed on the validation set, where the framework achieves 88.77\% accuracy and 87.05\% F1-score, exceeding the highest baseline validation accuracy by 1.09\% and the best validation F1-score by 0.82\%. This supporting evidence indicates that the observed improvements are not limited to the final test evaluation but are also reflected during model optimization.

\subsubsection{Confusion matrix analysis}

\begin{figure}[H]   
\centering

\begin{subfigure}[t]{0.49\textwidth}
    \centering
    \includegraphics[width=\linewidth]{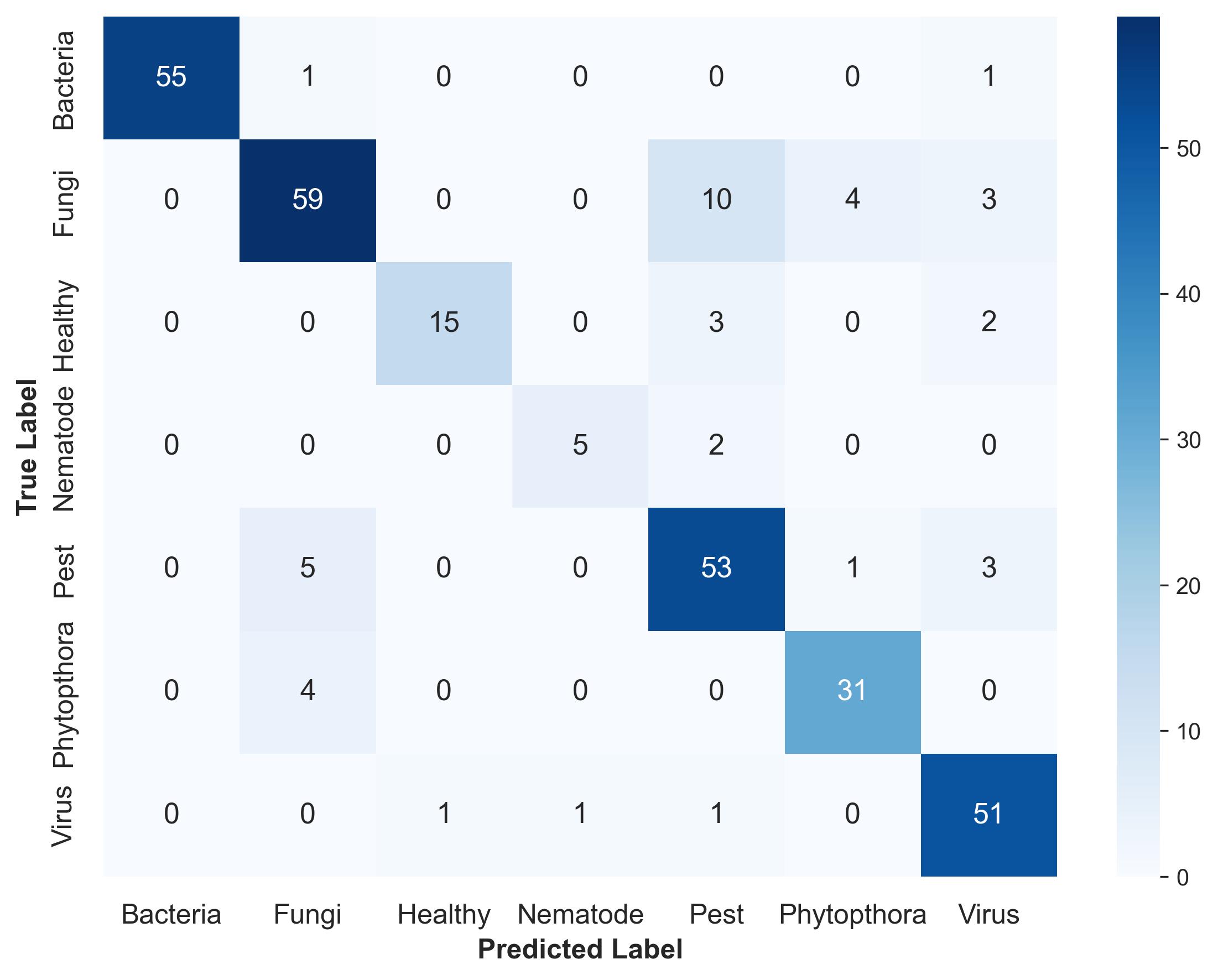}
    \caption{EfficientNet-B0}
\end{subfigure}
\hfill
\begin{subfigure}[t]{0.49\textwidth}
    \centering
    \includegraphics[width=\linewidth]{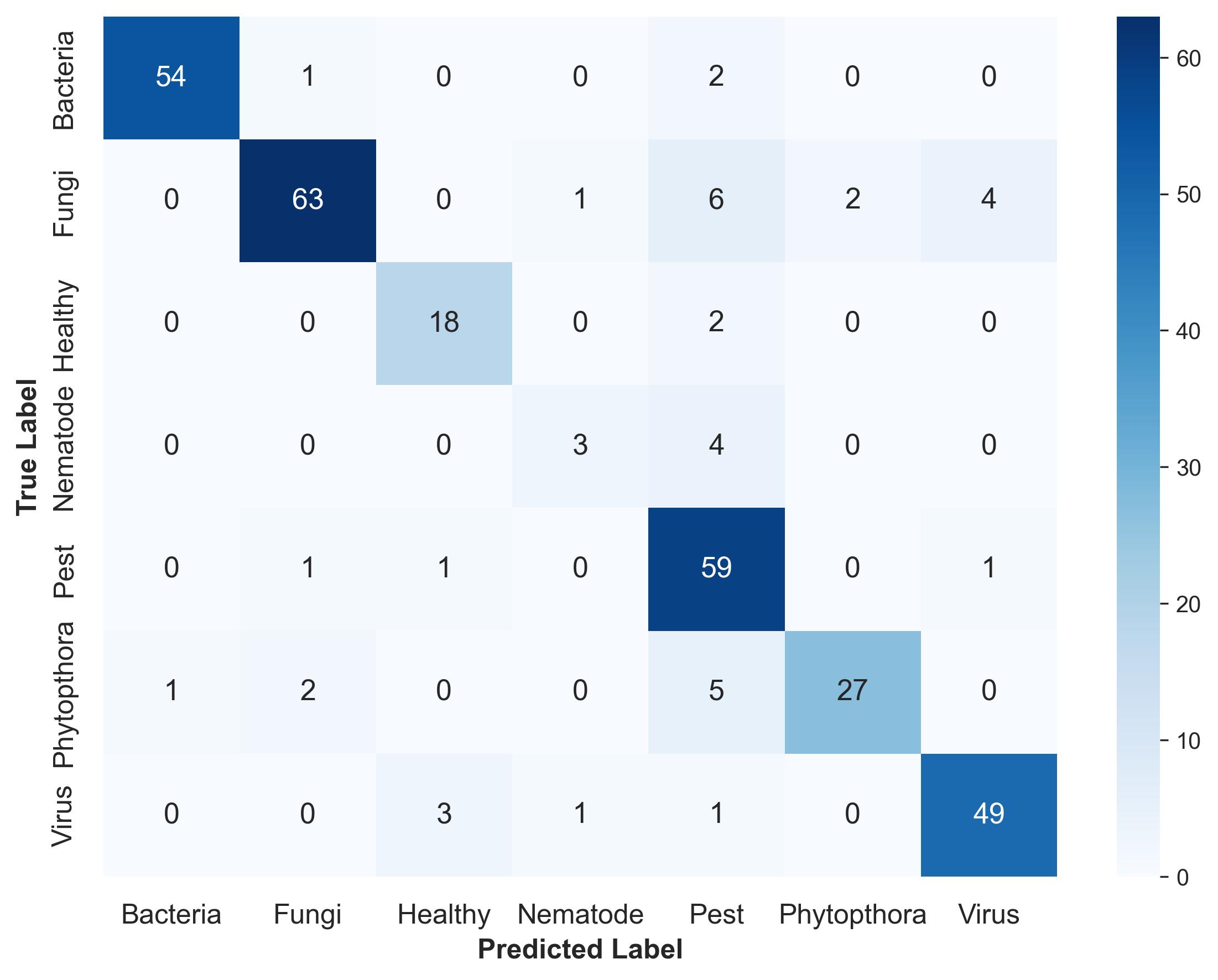}
    \caption{DenseNet-121}
\end{subfigure}

\begin{subfigure}[t]{0.49\textwidth}
    \centering
    \includegraphics[width=\linewidth]{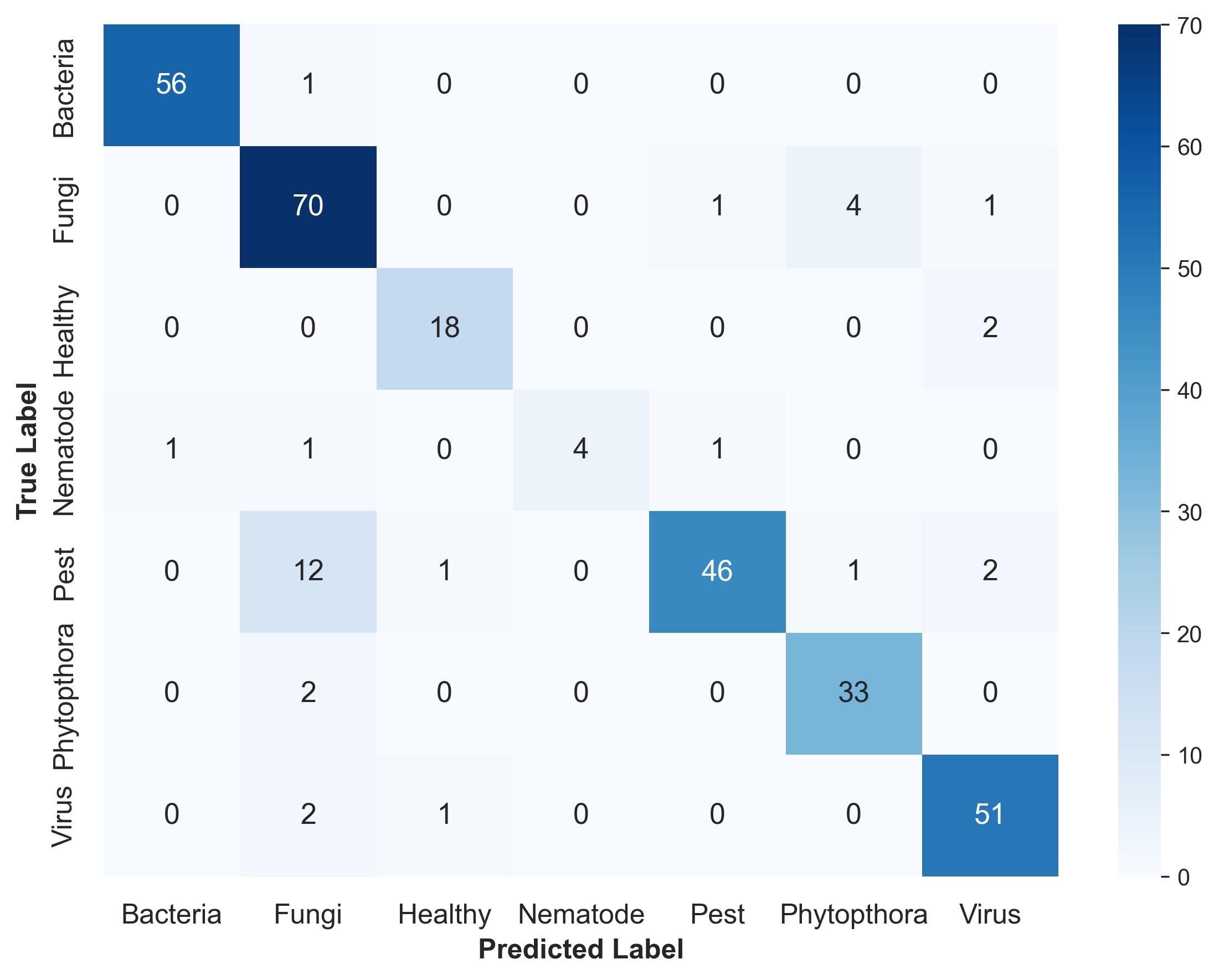}
    \caption{Swin-Tiny}
\end{subfigure}
\hfill
\begin{subfigure}[t]{0.49\textwidth}
    \centering
    \includegraphics[width=\linewidth]{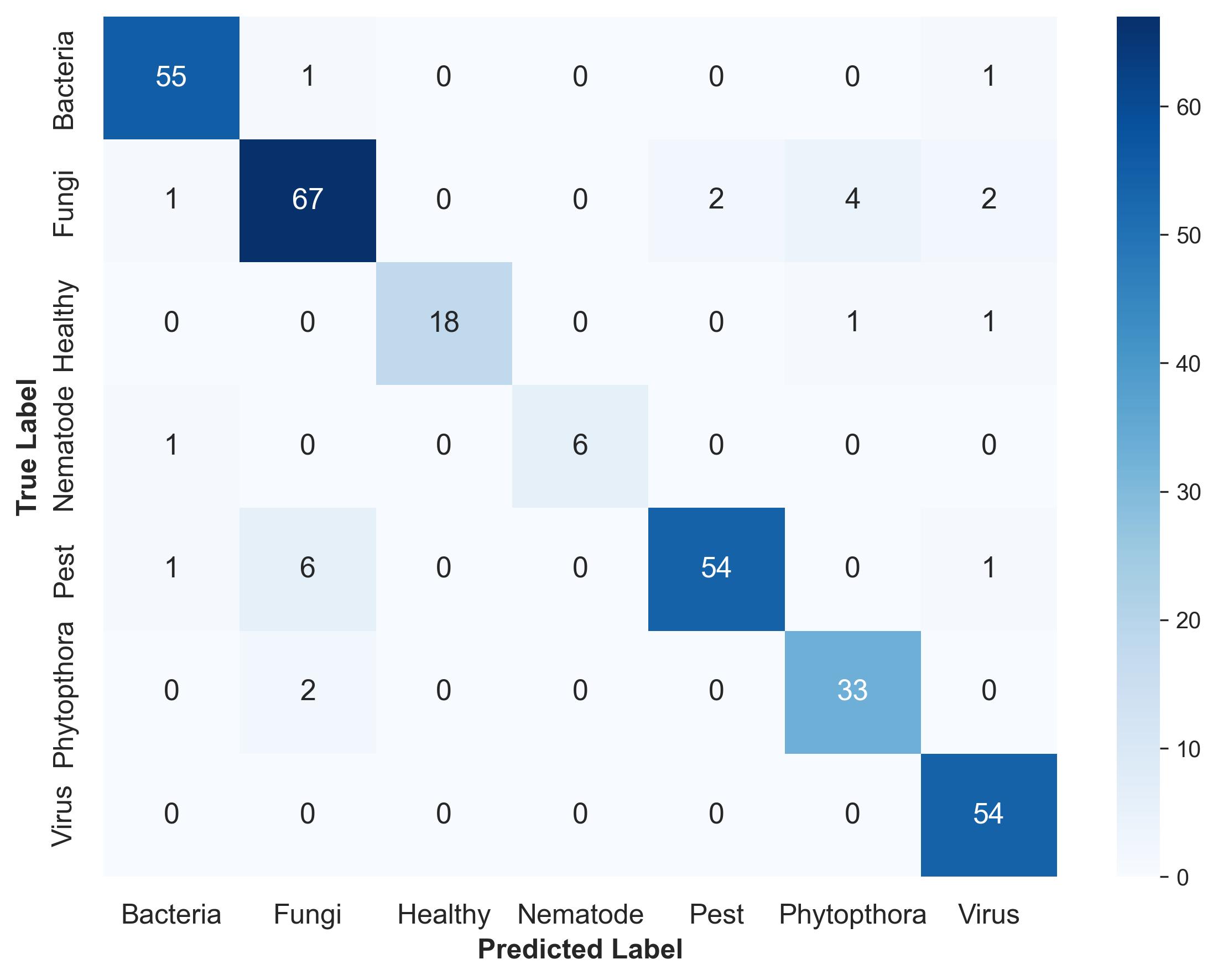}
    \caption{MoE}
\end{subfigure}

\vspace{-0.5em}
\caption{Confusion matrices on the test set of EfficientNet-B0, DenseNet-121, Swin-Tiny, and the proposed MoE framework for potato leaf disease classification.}

\label{fig:confusion_matrix_comparison}
\end{figure}

To better understand class-level errors under visual similarity and class imbalance, confusion matrices on the potato leaf disease test set are analyzed, as shown in Figure~\ref{fig:confusion_matrix_comparison}.

A key challenge in this task is the high inter-class similarity, particularly among Pest, Fungi, and Phytophthora. Individual baseline models exhibit noticeable confusion between these categories. For example, Swin-Tiny misclassifies 12 Pest samples as Fungi, while EfficientNet-B0 confuses 10 Fungi samples with Pest. DenseNet-121 also misclassifies several Phytophthora samples as Pest. In contrast, the MoE framework substantially reduces these errors, decreasing Pest→Fungi confusion to 6 samples, Fungi→Pest confusion to 2 samples, and completely eliminating Phytophthora→Pest misclassification. These results indicate stronger separation of visually similar disease patterns.

The confusion matrices also highlight improved robustness under severe class imbalance. Minority classes such as Nematode and Healthy remain challenging for individual models, with only 4-5 correctly classified Nematode samples in the baseline architectures. In comparison, the MoE framework correctly identifies 6 out of 7 Nematode samples and 18 out of 20 Healthy samples, indicating a stronger ability to recognize minority and difficult classes. This suggests that the adaptive routing mechanism reduces the bias toward majority classes by dynamically weighting expert contributions based on input characteristics rather than dataset-level priors.

These observations are consistent with the improvements in recall and F1-score reported in Table~\ref{tab:moe_potato_results}, further confirming the effectiveness of the MoE framework in reducing critical class-level misclassification errors.

\subsubsection{Configuration ablation study of the MoE framework}

A configuration ablation study was conducted on the validation set of the potato leaf disease dataset to assess the impact of key design choices within the MoE framework, including classifier head structure, normalization strategy, dropout configuration, and loss formulation. The detailed experimental results are summarized in Table~\ref{tab:ablation_head}.

To evaluate the effectiveness of the fused MoE representation, different classifier head configurations were investigated. Specifically, the proposed single-layer classifier was compared with two-layer MLP heads using progressively reduced hidden dimensions ($512 \rightarrow 384$, $512 \rightarrow 256$, and $512 \rightarrow 128$). Given the class imbalance of the potato dataset, F1-score and recall are particularly important for assessing balanced classification performance. The results show that increasing classifier depth does not provide consistent benefits. Although some two-layer configurations achieve comparable F1-scores, the single-layer design delivers the best overall trade-off, achieving the highest accuracy (88.77\%), the highest recall (87.74\%), and a tied best F1-score (87.05\%) with substantially lower architectural complexity. This suggests that the fused representation is already sufficiently discriminative, making additional hidden layers unnecessary.

\begin{table}[H]
\centering
\caption{Ablation study of the MoE framework under different configurations on the validation set of the potato leaf disease dataset.}
\label{tab:ablation_head}
\resizebox{\textwidth}{!}{%
\begin{tabular}{@{}lllccccc@{}}
\toprule
\textbf{Normalization} & \textbf{Classifier Head Structure (Depth)} & \textbf{Loss} & \textbf{Dropout} & \textbf{Accuracy} & \textbf{Precision} & \textbf{Recall} & \textbf{F1-Score} \\
 & & \textbf{Function} & \textbf{Rates} & \textbf{(\%)} & \textbf{(\%)} & \textbf{(\%)} & \textbf{(\%)} \\ \midrule
LayerNorm & 512 $\rightarrow$ 384 $\rightarrow$ classes (2 layers) & CE & 0.3, 0.3 & 87.68 & 88.85 & 85.04 & 86.21 \\
LayerNorm & 512 $\rightarrow$ 384 $\rightarrow$ classes (2 layers) & CE & 0.5, 0.3 & 87.68 & 89.09 & 85.40 & 87.02 \\
LayerNorm & 512 $\rightarrow$ 256 $\rightarrow$ classes (2 layers) & CE & 0.3, 0.3 & 87.32 & 88.58 & 85.92 & 87.05 \\
LayerNorm & 512 $\rightarrow$ 256 $\rightarrow$ classes (2 layers) & CE & 0.5, 0.3 & 87.32 & 85.85 & 86.34 & 85.86 \\
LayerNorm & 512 $\rightarrow$ 128 $\rightarrow$ classes (2 layers) & CE & 0.3, 0.3 & 86.96 & 87.99 & 86.38 & 86.79 \\
LayerNorm & 512 $\rightarrow$ 128 $\rightarrow$ classes (2 layers) & CE & 0.5, 0.3 & 86.59 & 88.05 & 83.65 & 85.28 \\
LayerNorm & 512 $\rightarrow$ classes (1 layer) & Weighted CE & 0.3 & 86.96 & 87.67 & 87.18 & 86.83 \\ 
BatchNorm & 512 $\rightarrow$ classes (1 layer) & CE & 0.3 & 87.68 & 86.64 & 86.04 & 86.20 \\
\textbf{LayerNorm} & \textbf{512 $\rightarrow$ classes (1 layer - Ours)} & \textbf{CE} & \textbf{0.3} & \textbf{88.77} & \textbf{87.24} & \textbf{87.74} & \textbf{87.05} \\ \bottomrule
\end{tabular}
}
\end{table}

The impact of loss formulation was also examined, particularly given the pronounced class imbalance of the potato dataset. Weighted Cross-Entropy (Weighted CE) was evaluated to determine whether explicit class reweighting could improve minority-class recognition. However, replacing standard Cross-Entropy (CE) reduced performance, lowering accuracy to 86.96\% and F1-score to 86.83\%. This suggests that the adaptive expert weighting mechanism already provides implicit balancing benefits, as reflected in the improved recall and stronger recognition of minority classes, making additional static class reweighting unnecessary.

The normalization strategy within the Feature Projection Module was also evaluated by comparing LayerNorm and BatchNorm. LayerNorm achieved better performance, improving accuracy from 87.68\% to 88.77\%. This result is consistent with the heterogeneous nature of the framework, which integrates both CNN-based experts and a transformer-based architecture, as well as the relatively small training batch size (16), where BatchNorm may be more sensitive to unstable batch statistics. In contrast, LayerNorm provides more stable feature alignment by operating independently of the batch dimension.

The ablation results support the final configuration consisting of a single-layer classifier, standard Cross-Entropy loss, LayerNorm, and a dropout rate of 0.3, which provides the most effective balance between preserving complementary feature information and maintaining optimization stability.

\subsubsection{Analysis of the MoE routing mechanism}

While the quantitative results confirm the effectiveness of the MoE framework, further analysis is needed to understand its internal routing behavior. This subsection examines three aspects: global expert importance, class-wise expert specialization, and gating entropy, providing insight into how the framework adapts expert contributions under varying input complexities.

\vspace{0.2cm}
\noindent \textbf{a) Expert importance and decision-level selection behavior}
\vspace{0.2cm}

\begin{figure}[H]
    \centering
    \includegraphics[width=\textwidth]{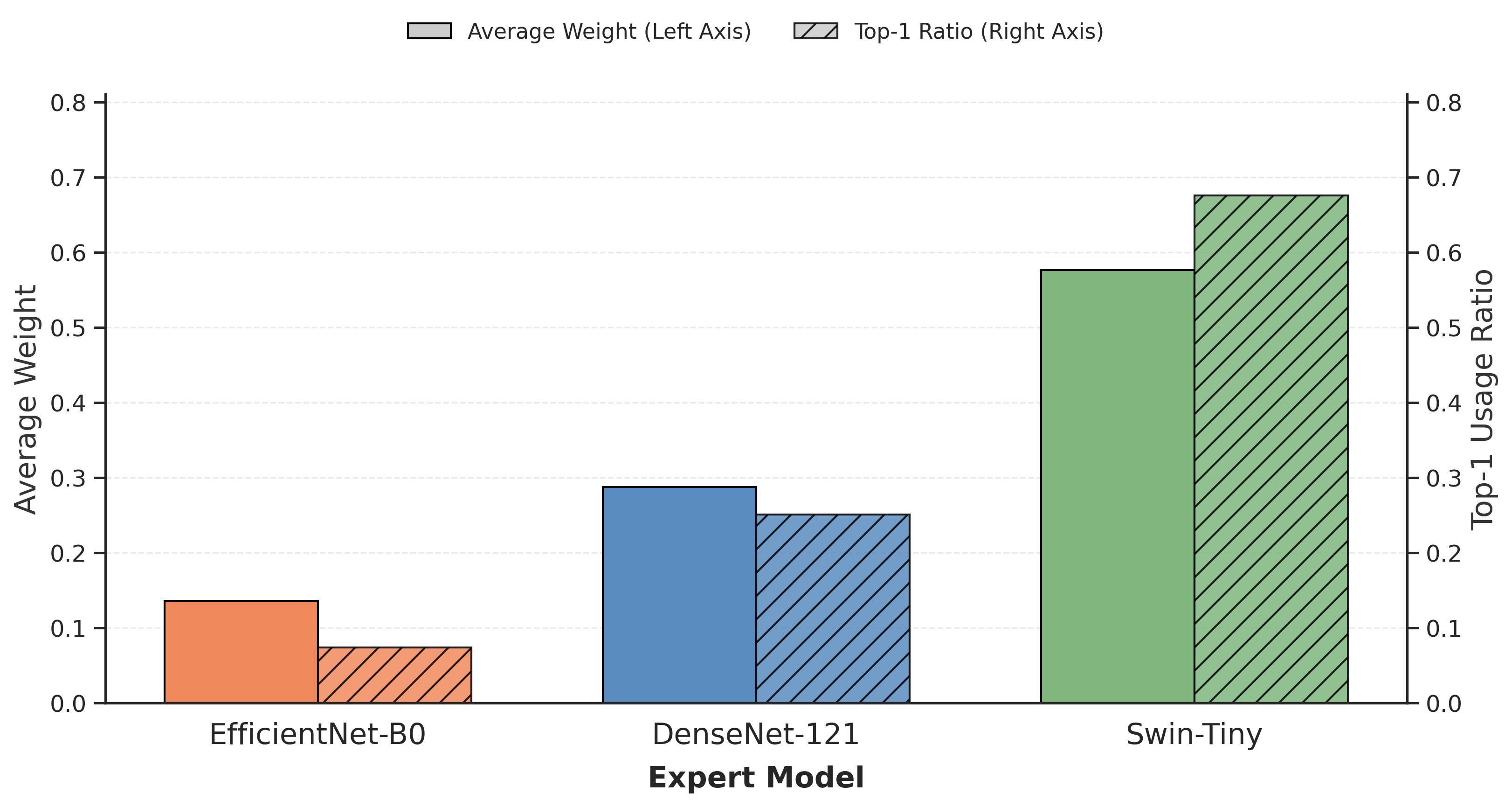}
    \caption{Global expert importance and decision-level selection patterns in the proposed MoE framework.}
    \label{fig:expert_importance}
\end{figure}

Figure~\ref{fig:expert_importance} presents two global routing metrics derived from the soft gating network: the average gating weight assigned to each expert across the test set and the top-1 selection ratio, which indicates how frequently an expert receives the highest routing weight.

Swin-Tiny exhibits the strongest overall contribution, with an average gating weight of approximately 0.58 and a top-1 selection ratio of about 68\%. This indicates that the gating network frequently prioritizes its representations. This behavior is consistent with Swin-Tiny’s strength in modeling broader contextual dependencies, which are particularly beneficial under visually complex field conditions.
DenseNet-121 serves as the second most influential expert, with an average gating weight of approximately 0.29 and a top-1 ratio of roughly 25\%. Its contribution reflects a complementary role, supporting the framework through fine-grained local texture modeling that complements the global contextual representations provided by Swin-Tiny.
In contrast, EfficientNet-B0 contributes more modestly, with an average gating weight of approximately 0.14 and a top-1 ratio below 9\%. Its lower but non-zero contribution suggests a supportive complementary role, providing additional multi-scale feature cues for specific input conditions.

The clearly non-uniform routing distribution indicates that the framework learns an input-adaptive expert allocation strategy rather than behaving as a conventional ensemble with uniform expert averaging.

\vspace{0.2cm}
\noindent \textbf{b) Class-wise expert specialization}
\vspace{0.2cm}

\begin{figure}[H]
    \centering
    \includegraphics[width=\textwidth]{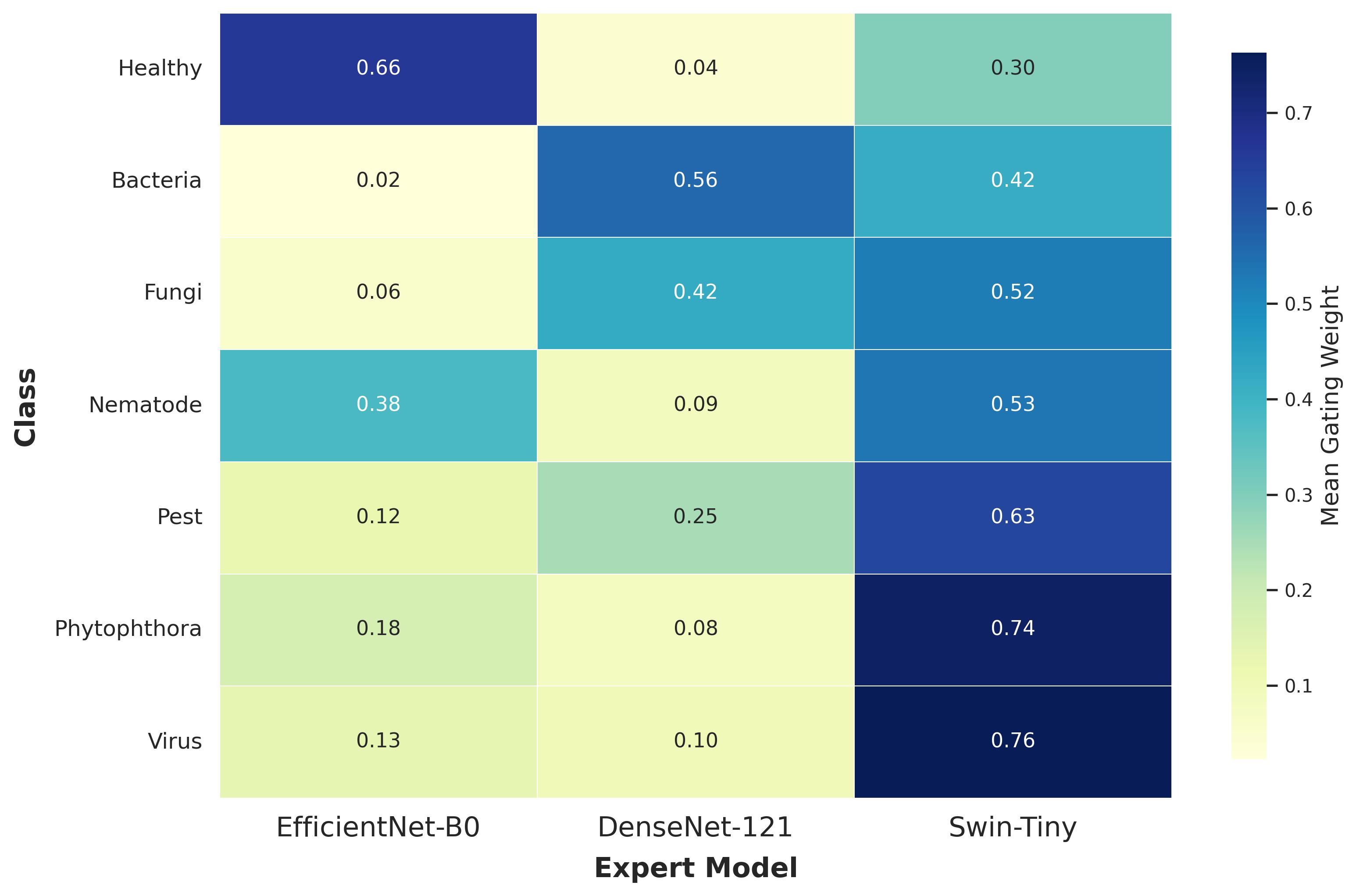}
    \caption{Class-wise expert specialization patterns across disease categories.}
    \label{fig:class_specialization}
\end{figure}

To further understand the routing behavior of the soft gating network at the class level, class-wise expert specialization is analyzed by examining the average gating weights assigned to each expert across different disease categories, as shown in Figure~\ref{fig:class_specialization}. The results reveal clear specialization patterns, indicating that the model adaptively prioritizes different experts according to class-specific visual characteristics.

Swin-Tiny receives the highest gating weights across most classes. It dominates the classification process for Virus, Phytophthora, and Pest, with average weights of 0.76, 0.74, and 0.63, respectively. It also provides the largest contribution for the Nematode (0.53) and Fungi (0.52) classes, indicating its strong capability in handling complex and spatially distributed disease patterns.
In contrast, DenseNet-121 exhibits a clear specialization for the Bacteria class, where it achieves the highest gating weight of 0.56 among all experts. It also plays a significant secondary role in the Fungi class, with a weight of 0.42, reflecting its effectiveness in capturing fine-grained local texture features.
EfficientNet-B0, on the other hand, shows a highly concentrated contribution to the Healthy class, where it obtains a dominant weight of 0.66. Its secondary contribution is observed in the Nematode class (0.38), suggesting its usefulness in handling variations in feature scale and overall leaf structure.

The heatmap also reveals that certain classes rely on collaborative expert contributions rather than a single dominant expert. For example, Fungi is jointly influenced by Swin-Tiny (0.52) and DenseNet-121 (0.42), while Nematode receives substantial contributions from both Swin-Tiny (0.53) and EfficientNet-B0 (0.38). These patterns confirm that the framework performs adaptive class-dependent expert allocation rather than fixed or uniform expert fusion.

\vspace{0.2cm}
\noindent \textbf{c) Adaptive routing behavior via gating entropy}
\vspace{0.2cm}

\begin{figure}[H]
    \centering
    \includegraphics[width=\textwidth]{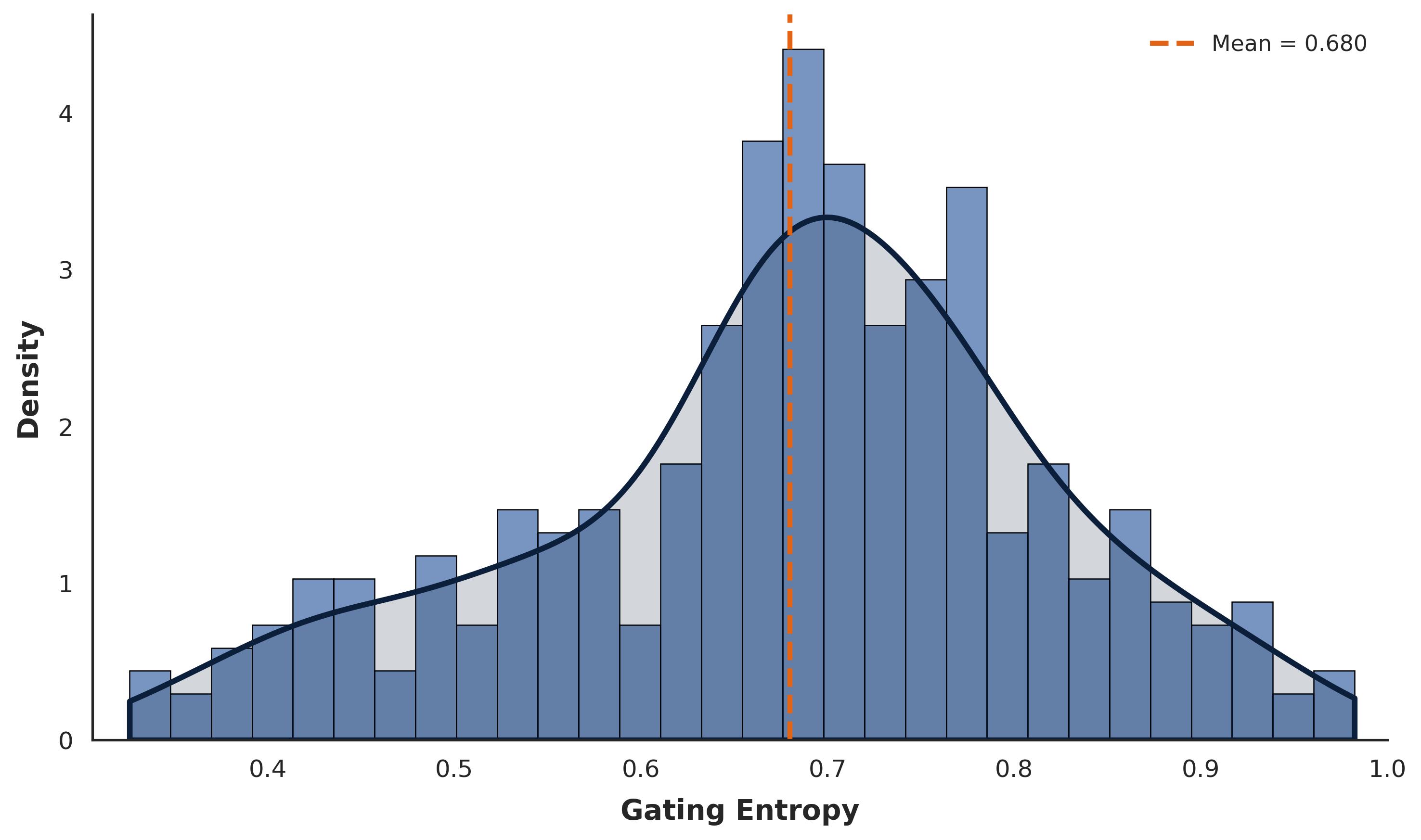}
    \caption{Gating entropy distribution characterizing adaptive expert routing behavior.}
    \label{fig:gating_entropy}
\end{figure}

Beyond class-level specialization, the adaptive routing behavior of the soft gating mechanism is further examined through the distribution of gating entropy across the test set, as illustrated in Figure~\ref{fig:gating_entropy}. In the proposed framework, gating entropy quantifies how concentrated or distributed the routing weights are among experts for each input sample. Lower entropy indicates stronger concentration toward a dominant expert, whereas higher entropy reflects more collaborative routing across multiple experts.

As shown in Figure~\ref{fig:gating_entropy}, entropy values span a broad range from approximately 0.3 to 1.0, with a mean of 0.680. This wide distribution indicates substantial variation in routing behavior across samples, confirming that the framework does not rely on a fixed expert allocation strategy.

For low-entropy samples (e.g., $<0.5$), routing is strongly concentrated on a single expert, reflecting more decisive expert selection. In contrast, for high-entropy samples (e.g., $>0.8$), routing weights are distributed more evenly, indicating collaborative integration of multiple expert representations.

These observations suggest that the MoE framework dynamically switches between selective expert routing and collaborative expert fusion depending on the characteristics of each input, further distinguishing it from conventional ensemble methods with static or uniform expert aggregation.

\subsubsection{Interpretability analysis using Grad-CAM visualizations}

\begin{figure}[H]
\centering
\tiny

\begin{subfigure}[t]{0.19\linewidth}
    \centering
    \includegraphics[width=\linewidth]{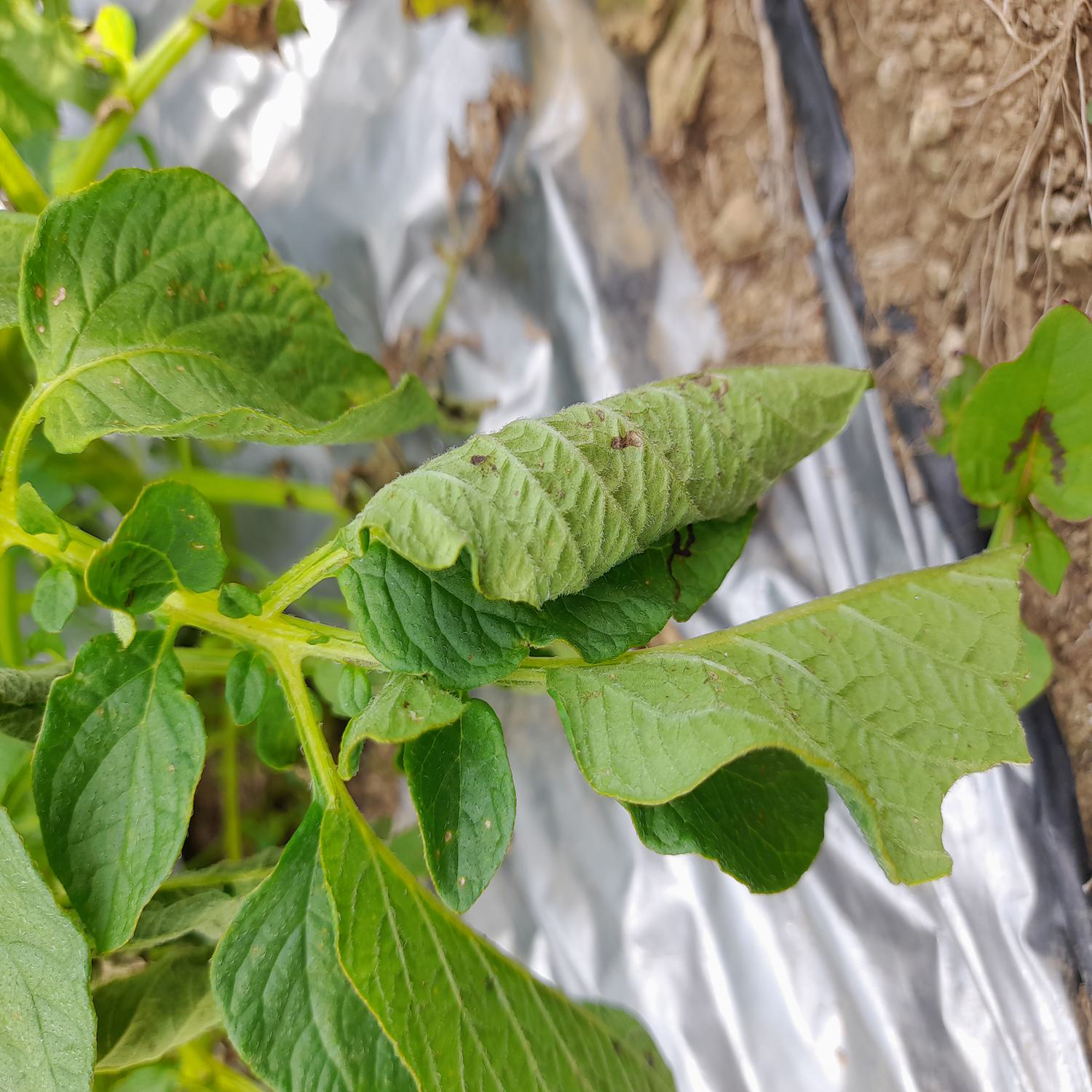}
    \caption{Original (Bacteria)}
    \label{fig:gradcam_bacteria_original}
\end{subfigure}
\hfill
\begin{subfigure}[t]{0.19\linewidth}
    \centering
    \includegraphics[width=\linewidth]{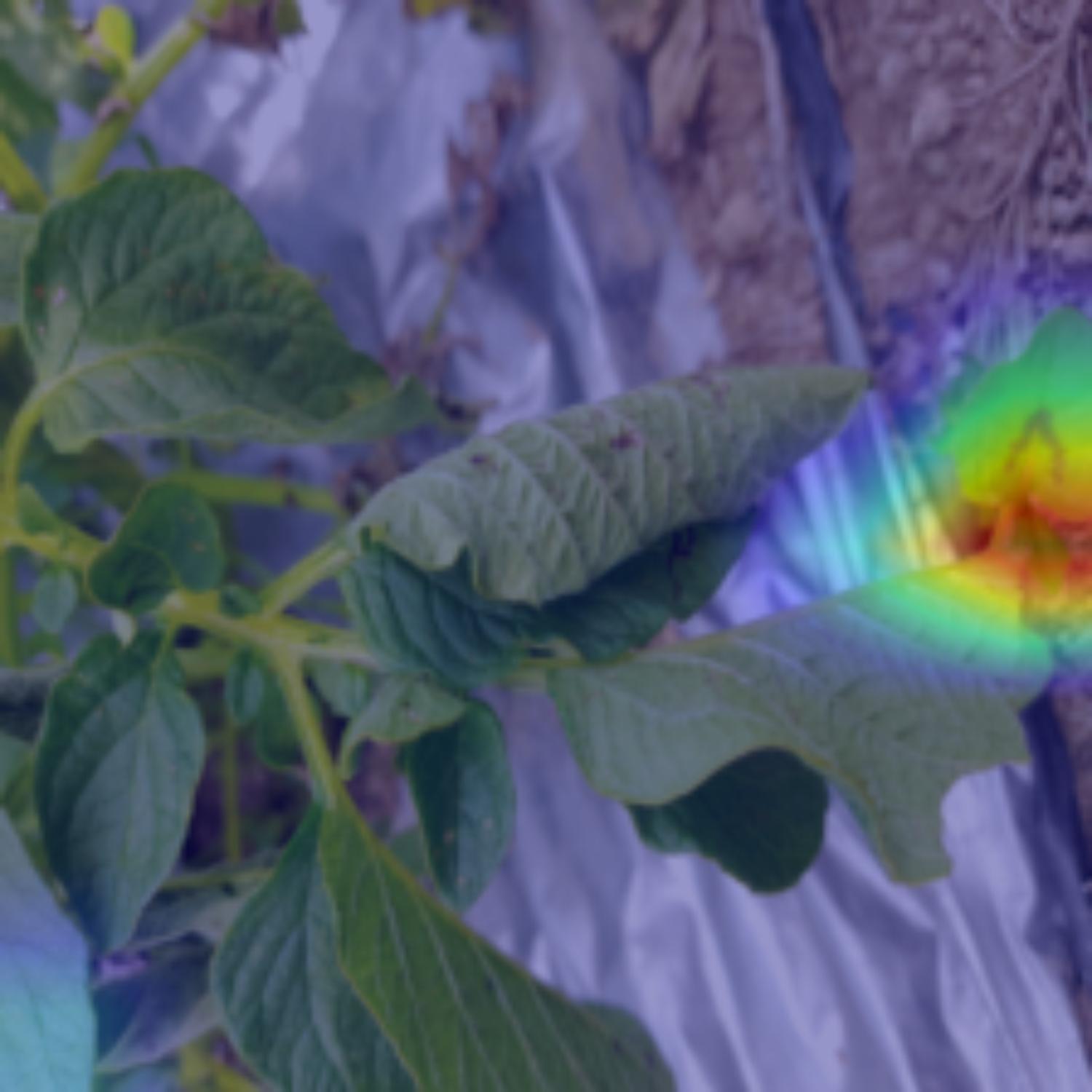}
    \caption{EfficientNet-B0 Expert}
    \label{fig:gradcam_bacteria_efficientnet}
\end{subfigure}
\hfill
\begin{subfigure}[t]{0.19\linewidth}
    \centering
    \includegraphics[width=\linewidth]{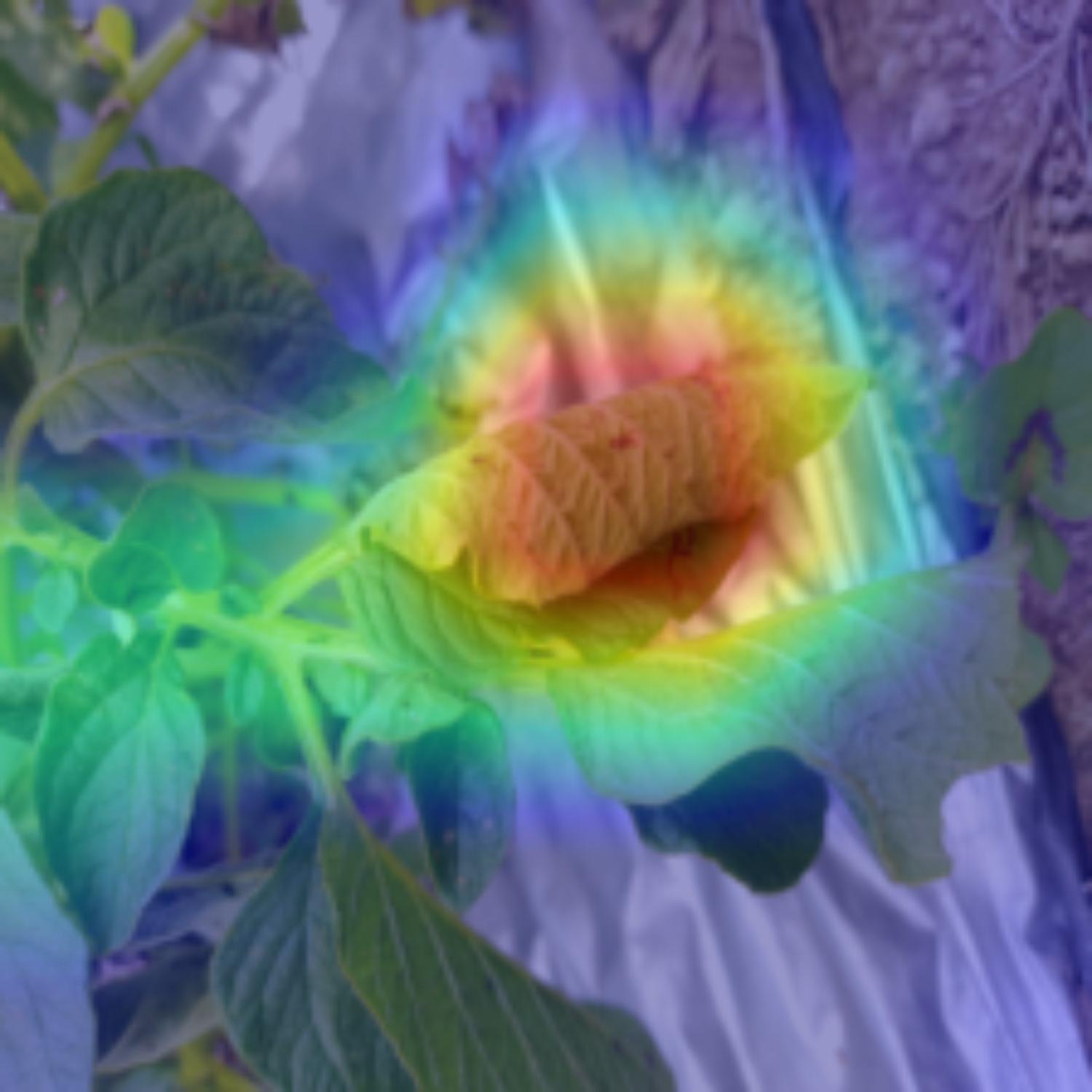}
    \caption{DenseNet-121 Expert}
    \label{fig:gradcam_bacteria_densenet}
\end{subfigure}
\hfill
\begin{subfigure}[t]{0.19\linewidth}
    \centering
    \includegraphics[width=\linewidth]{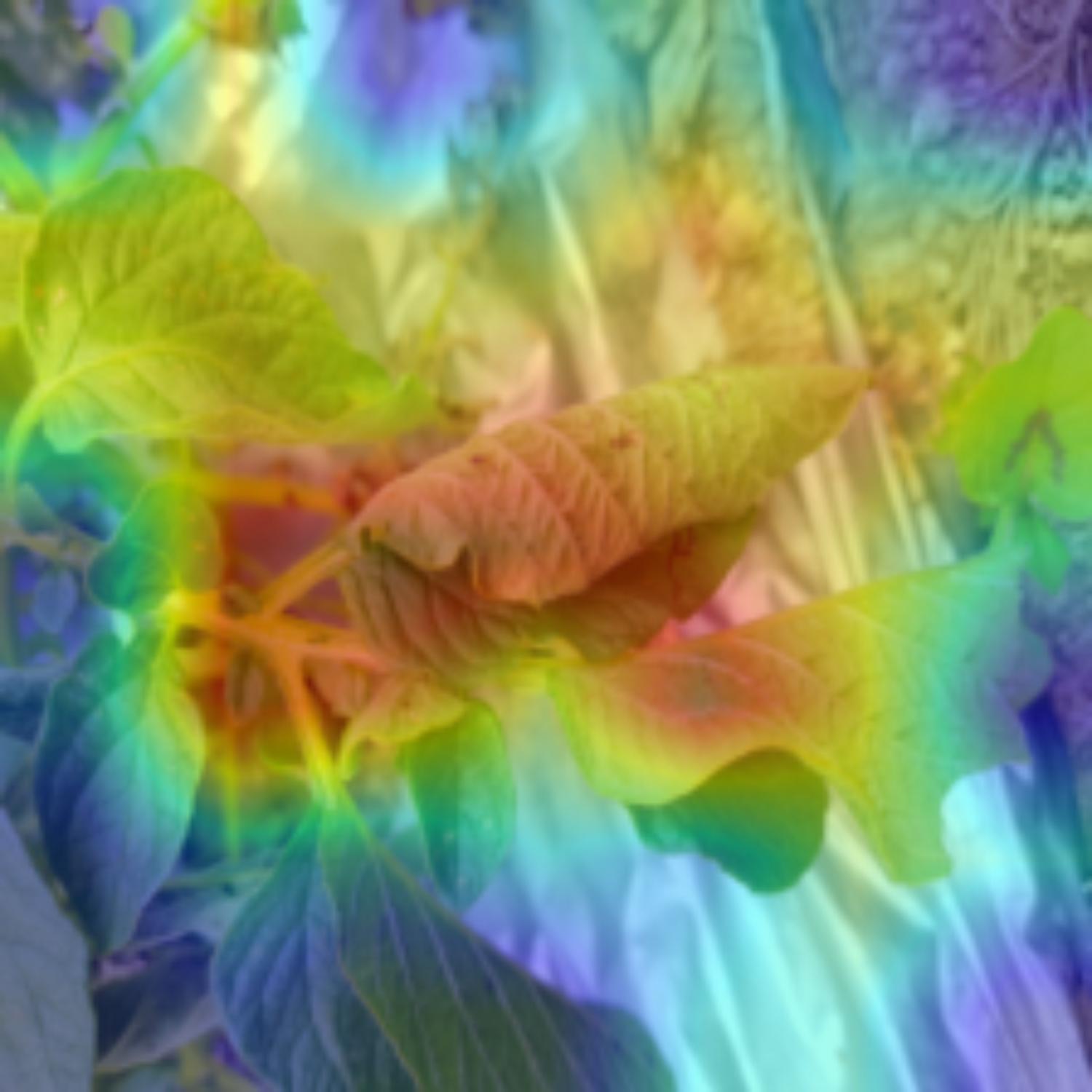}
    \caption{Swin-Tiny Expert}
    \label{fig:gradcam_bacteria_swin}
\end{subfigure}
\hfill
\begin{subfigure}[t]{0.19\linewidth}
    \centering
    \includegraphics[width=\linewidth]{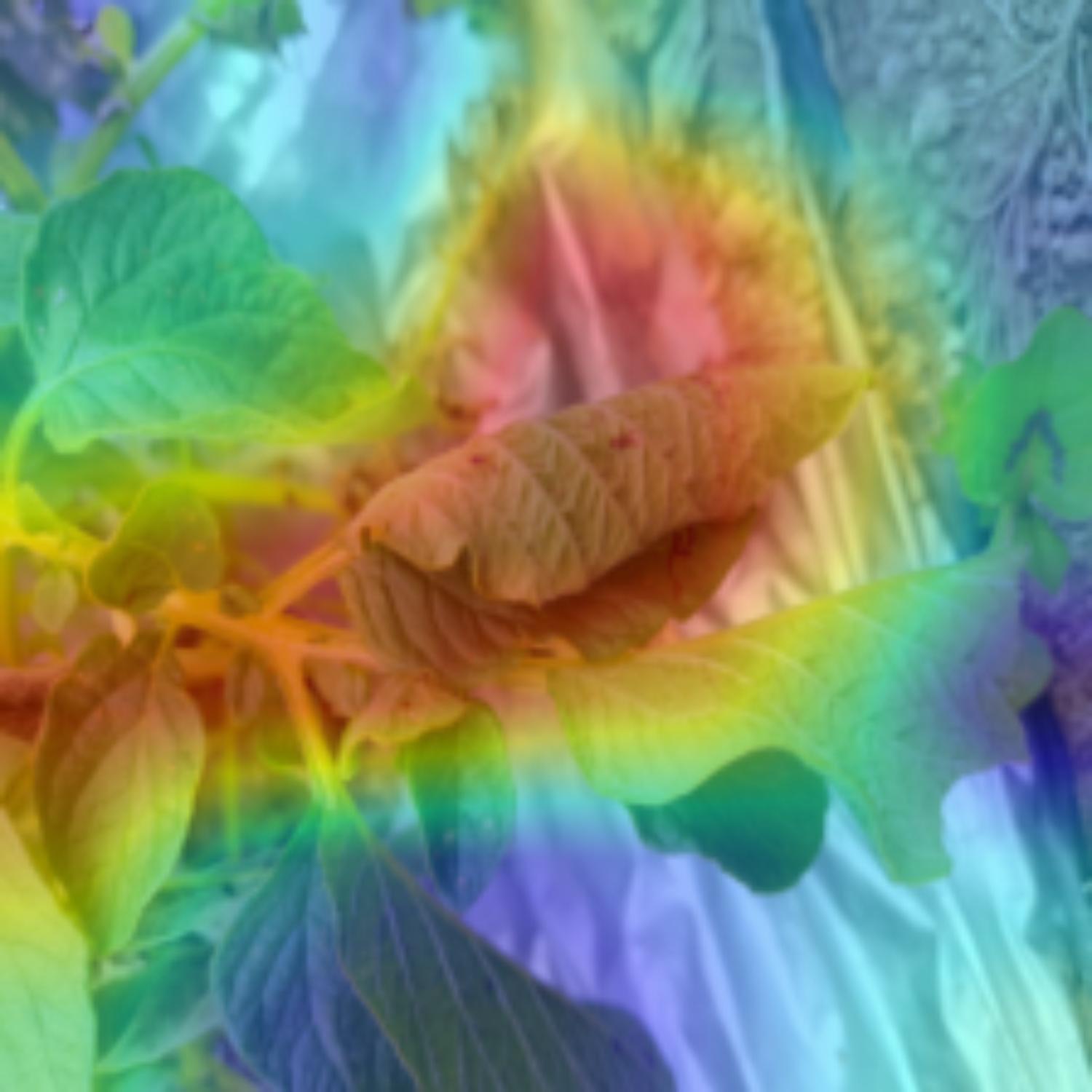}
    \caption{MoE}
    \label{fig:gradcam_bacteria_moe}
\end{subfigure}

\begin{subfigure}[t]{0.19\linewidth}
    \centering
    \includegraphics[width=\linewidth]{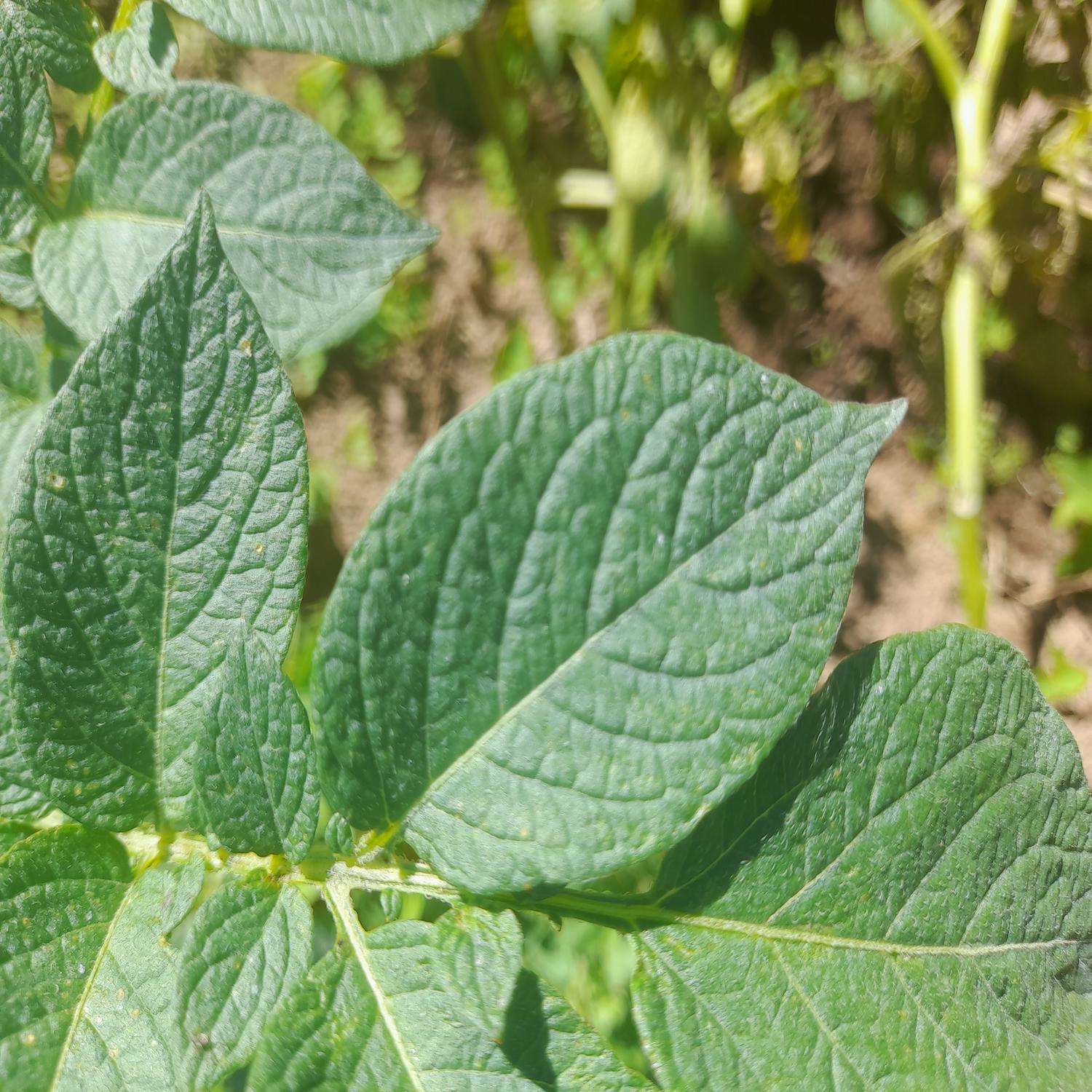}
    \caption{Original (Healthy)}
    \label{fig:gradcam_healthy_original}
\end{subfigure}
\hfill
\begin{subfigure}[t]{0.19\linewidth}
    \centering
    \includegraphics[width=\linewidth]{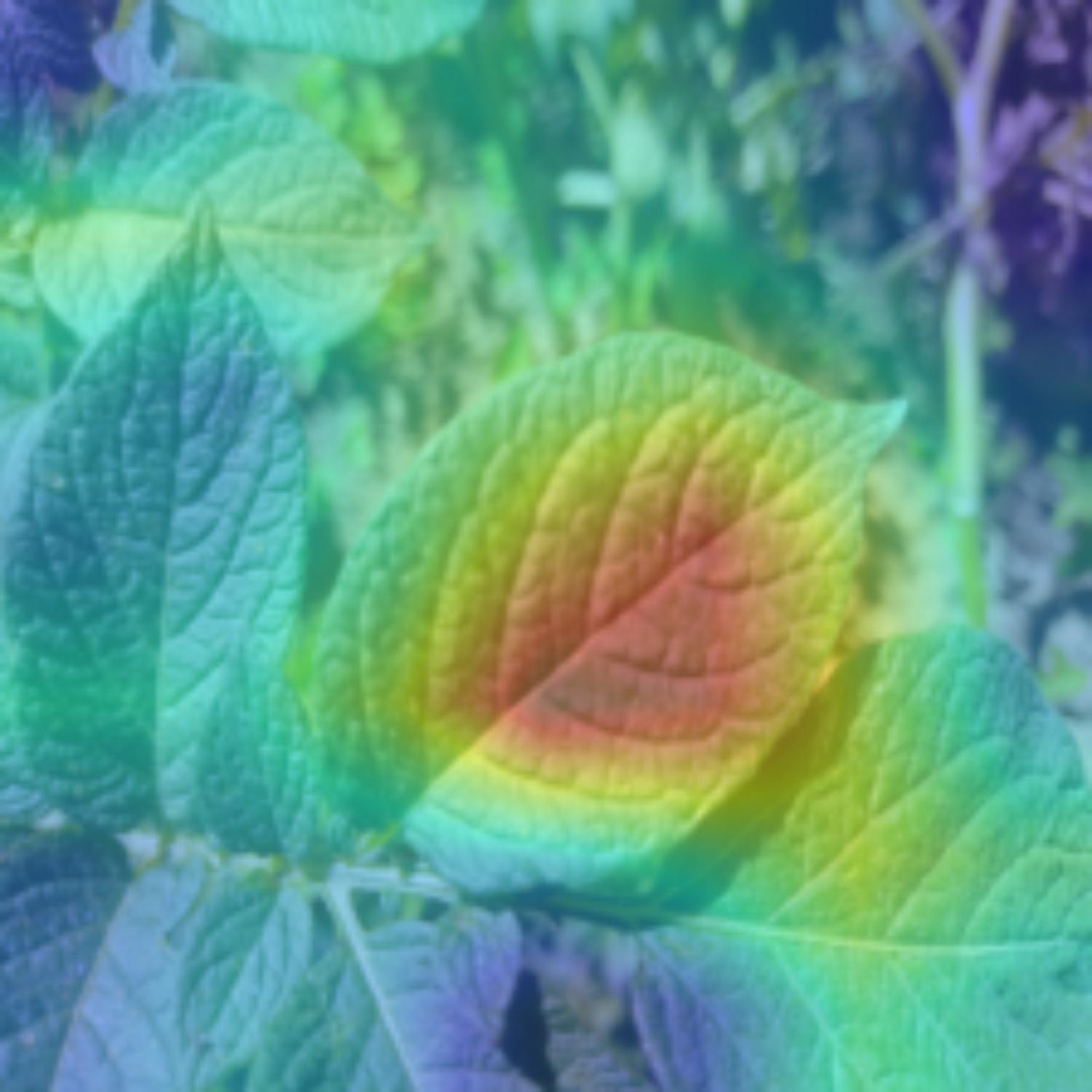}
    \caption{EfficientNet-B0 Expert}
    \label{fig:gradcam_healthy_efficientnet}
\end{subfigure}
\hfill
\begin{subfigure}[t]{0.19\linewidth}
    \centering
    \includegraphics[width=\linewidth]{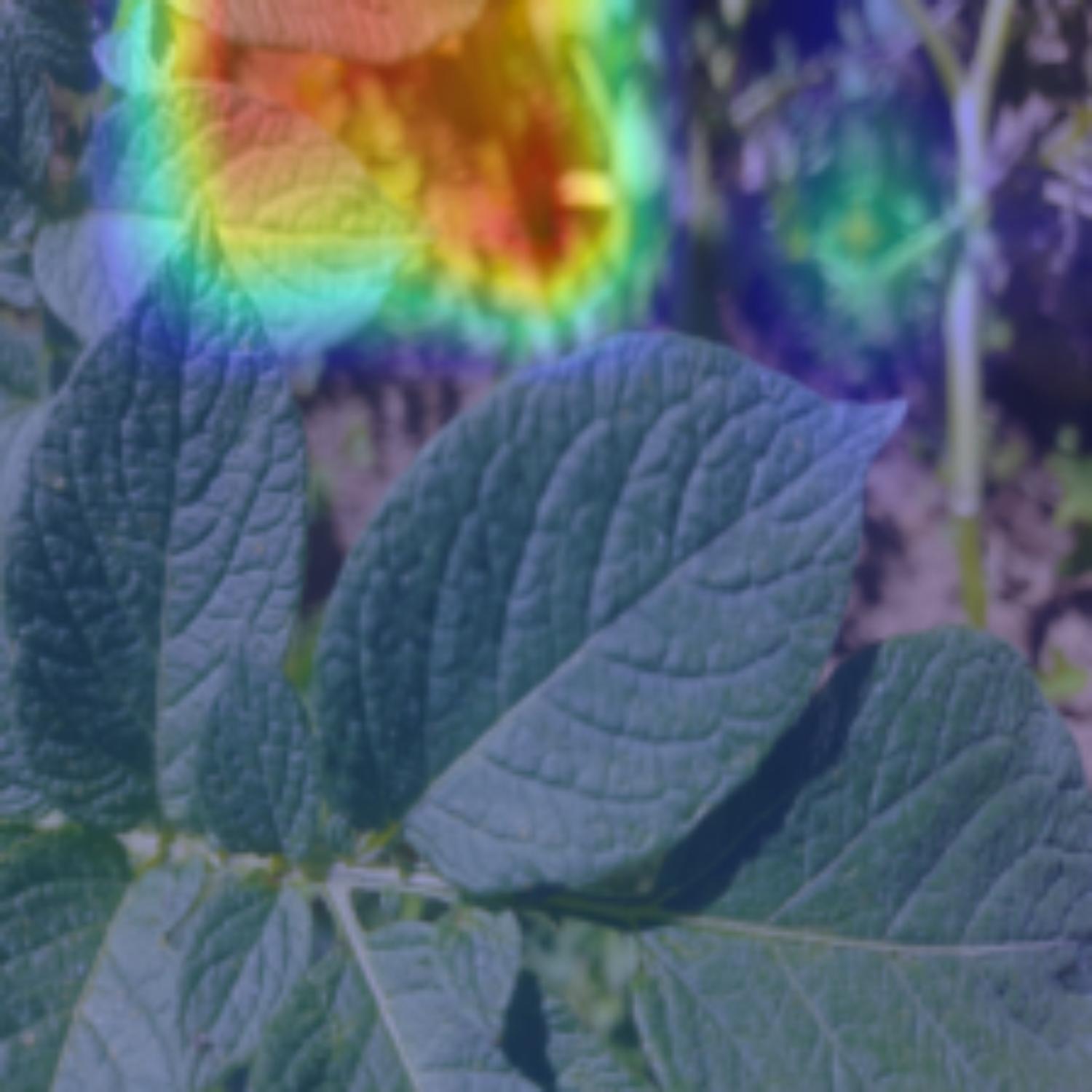}
    \caption{DenseNet-121 Expert}
    \label{fig:gradcam_healthy_densenet}
\end{subfigure}
\hfill
\begin{subfigure}[t]{0.19\linewidth}
    \centering
    \includegraphics[width=\linewidth]{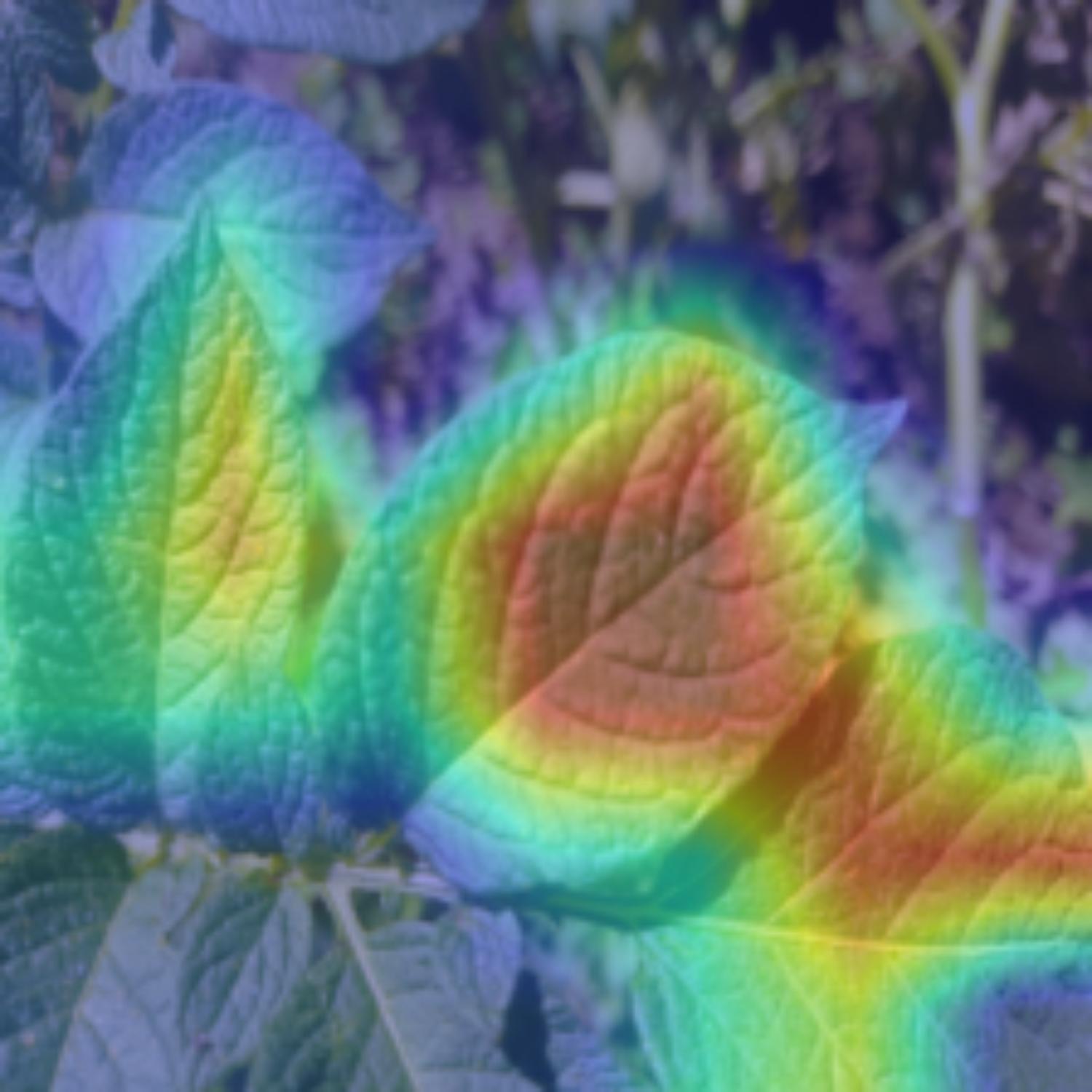}
    \caption{Swin-Tiny Expert}
    \label{fig:gradcam_healthy_swin}
\end{subfigure}
\hfill
\begin{subfigure}[t]{0.19\linewidth}
    \centering
    \includegraphics[width=\linewidth]{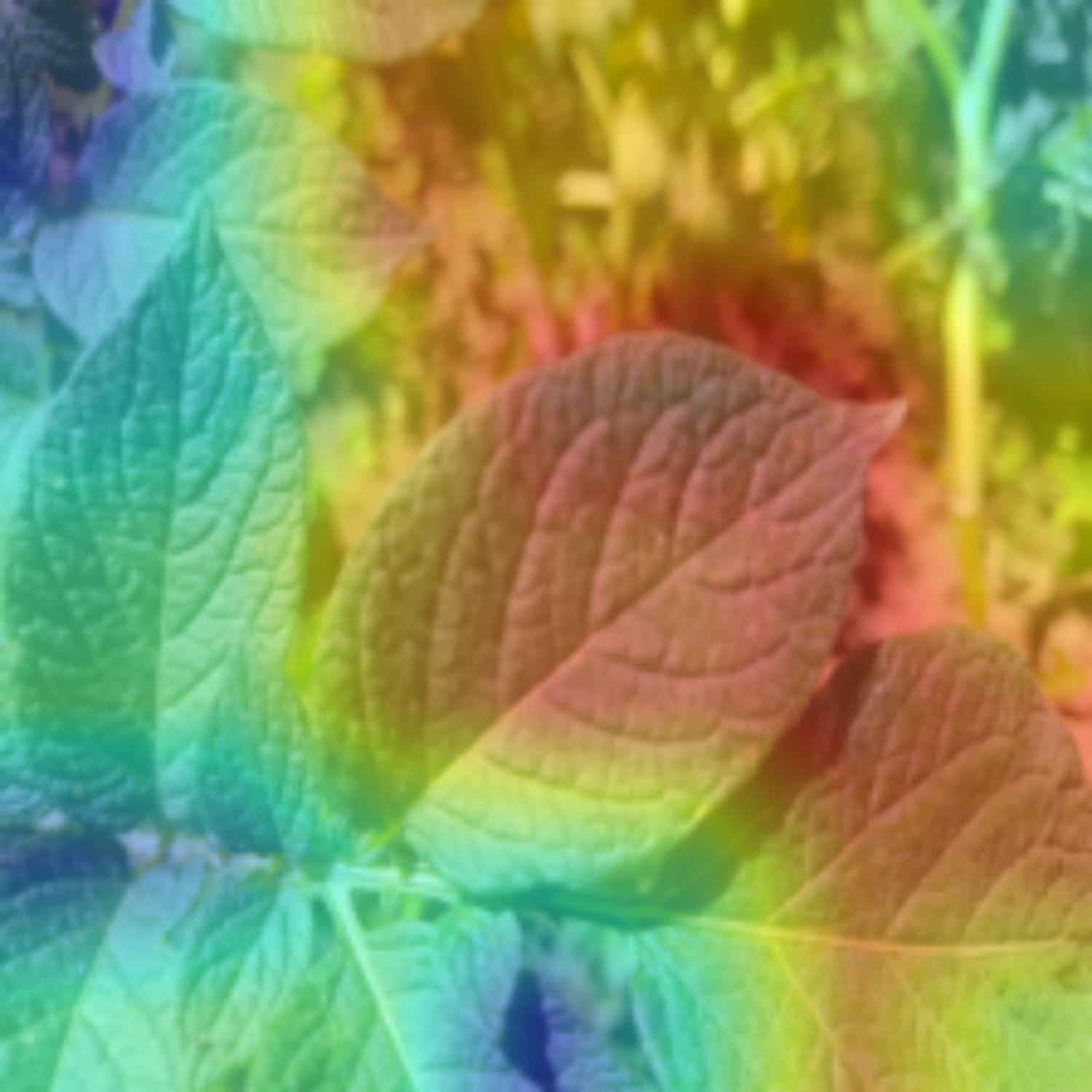}
    \caption{MoE}
    \label{fig:gradcam_healthy_moe}
\end{subfigure}

\begin{subfigure}[t]{0.19\linewidth}
    \centering
    \includegraphics[width=\linewidth]{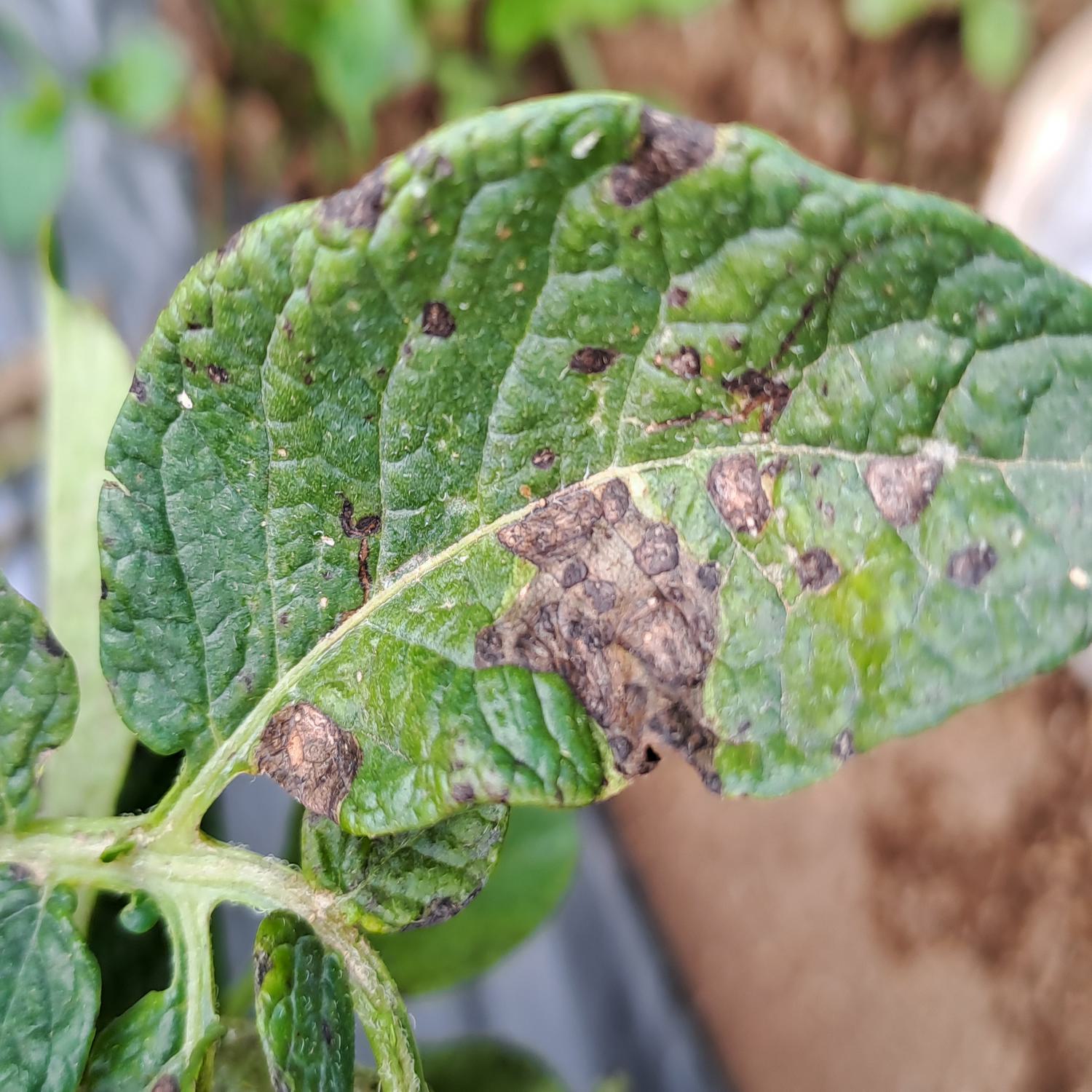}
    \caption{Original (Fungi)}
    \label{fig:gradcam_fungi_original}
\end{subfigure}
\hfill
\begin{subfigure}[t]{0.19\linewidth}
    \centering
    \includegraphics[width=\linewidth]{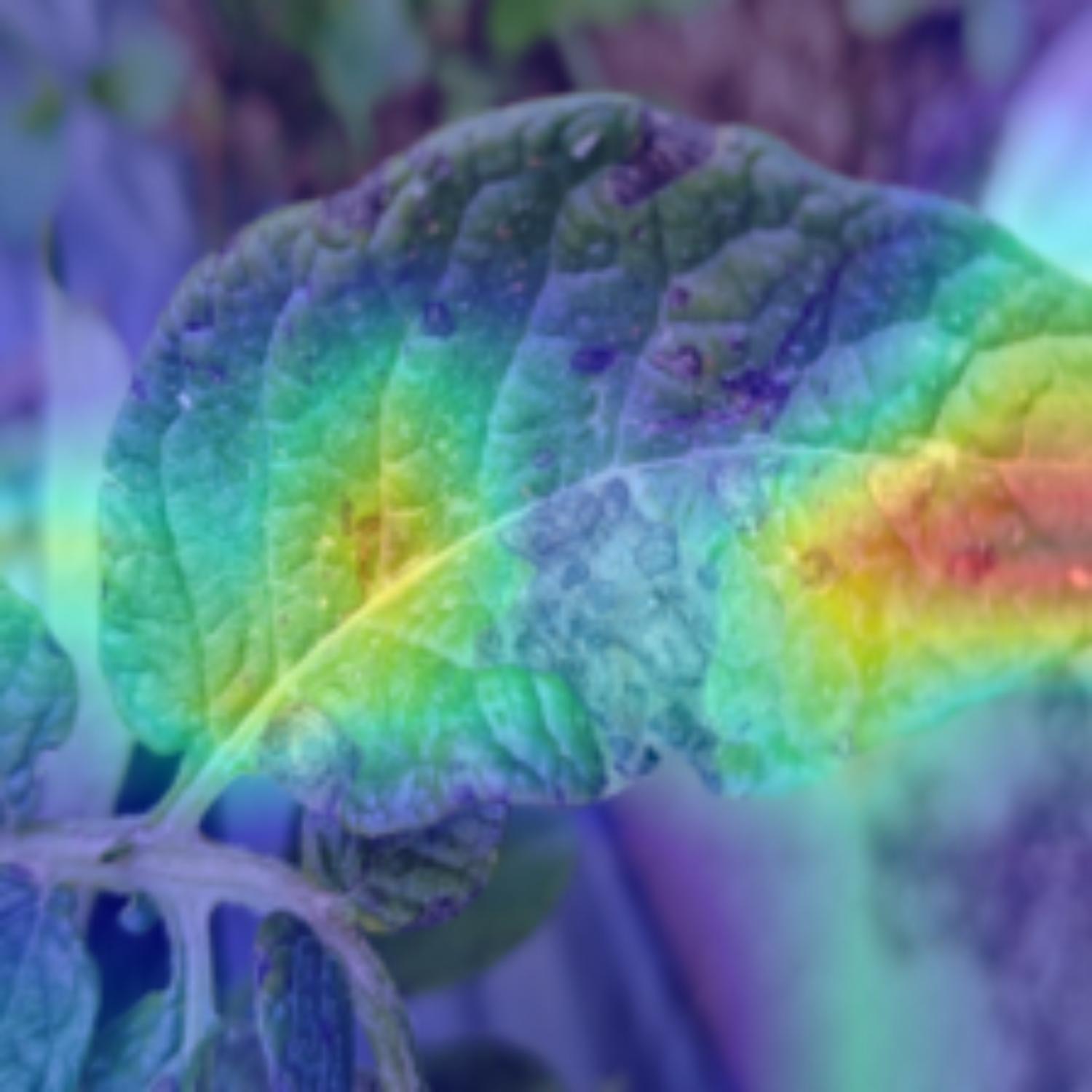}
    \caption{EfficientNet-B0 Expert}
    \label{fig:gradcam_fungi_efficientnet}
\end{subfigure}
\hfill
\begin{subfigure}[t]{0.19\linewidth}
    \centering
    \includegraphics[width=\linewidth]{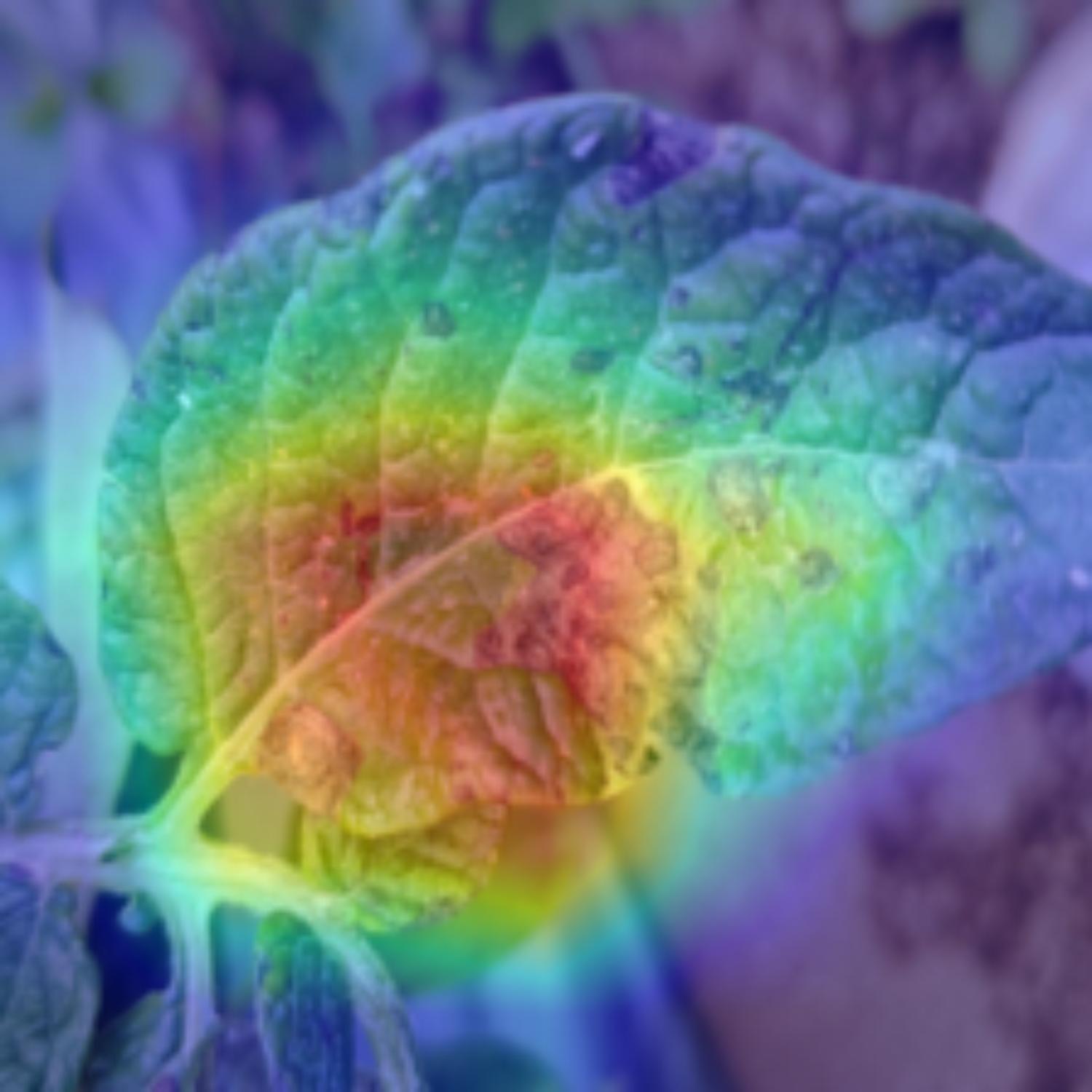}
    \caption{DenseNet-121 Expert}
    \label{fig:gradcam_fungi_densenet}
\end{subfigure}
\hfill
\begin{subfigure}[t]{0.19\linewidth}
    \centering
    \includegraphics[width=\linewidth]{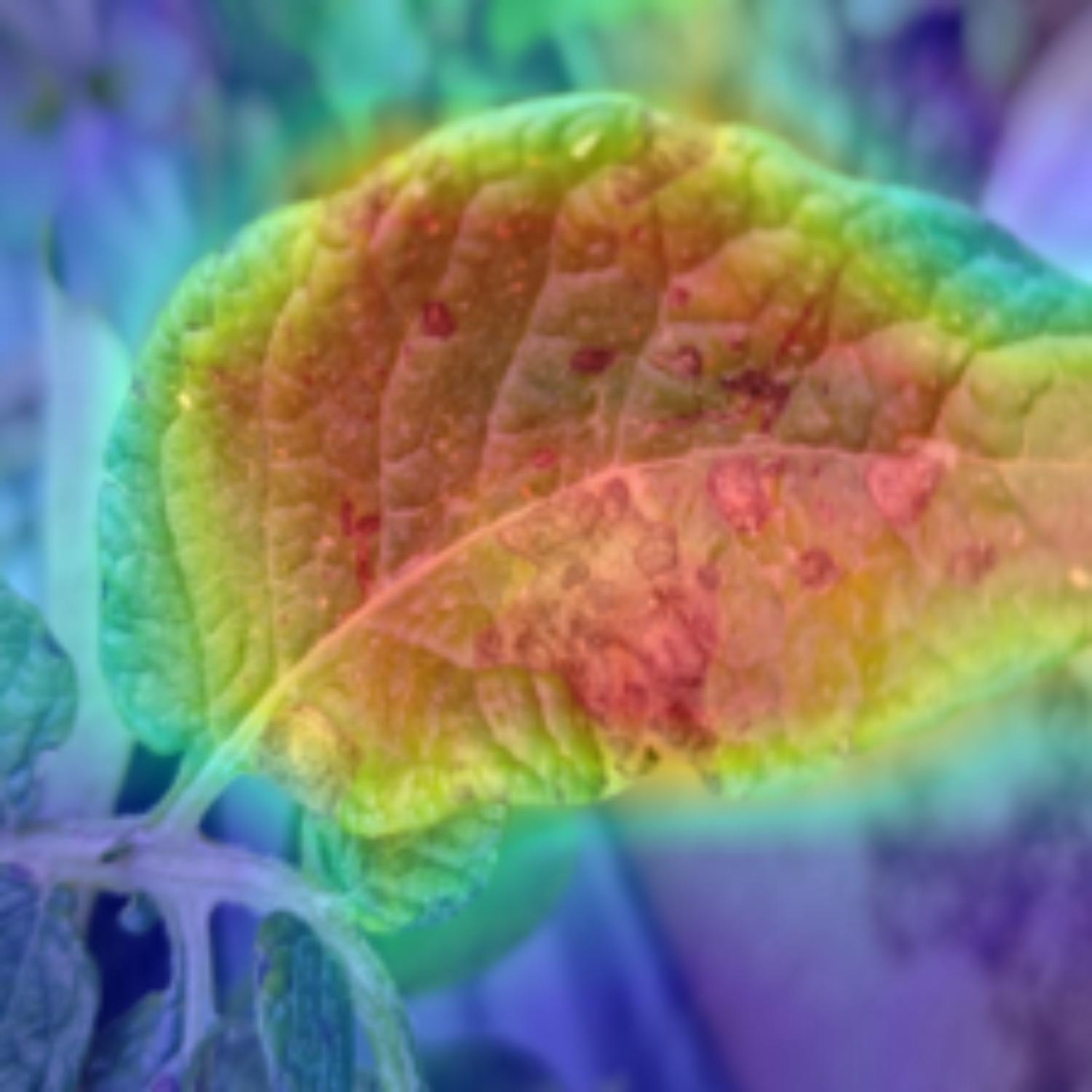}
    \caption{Swin-Tiny Expert}
    \label{fig:gradcam_fungi_swin}
\end{subfigure}
\hfill
\begin{subfigure}[t]{0.19\linewidth}
    \centering
    \includegraphics[width=\linewidth]{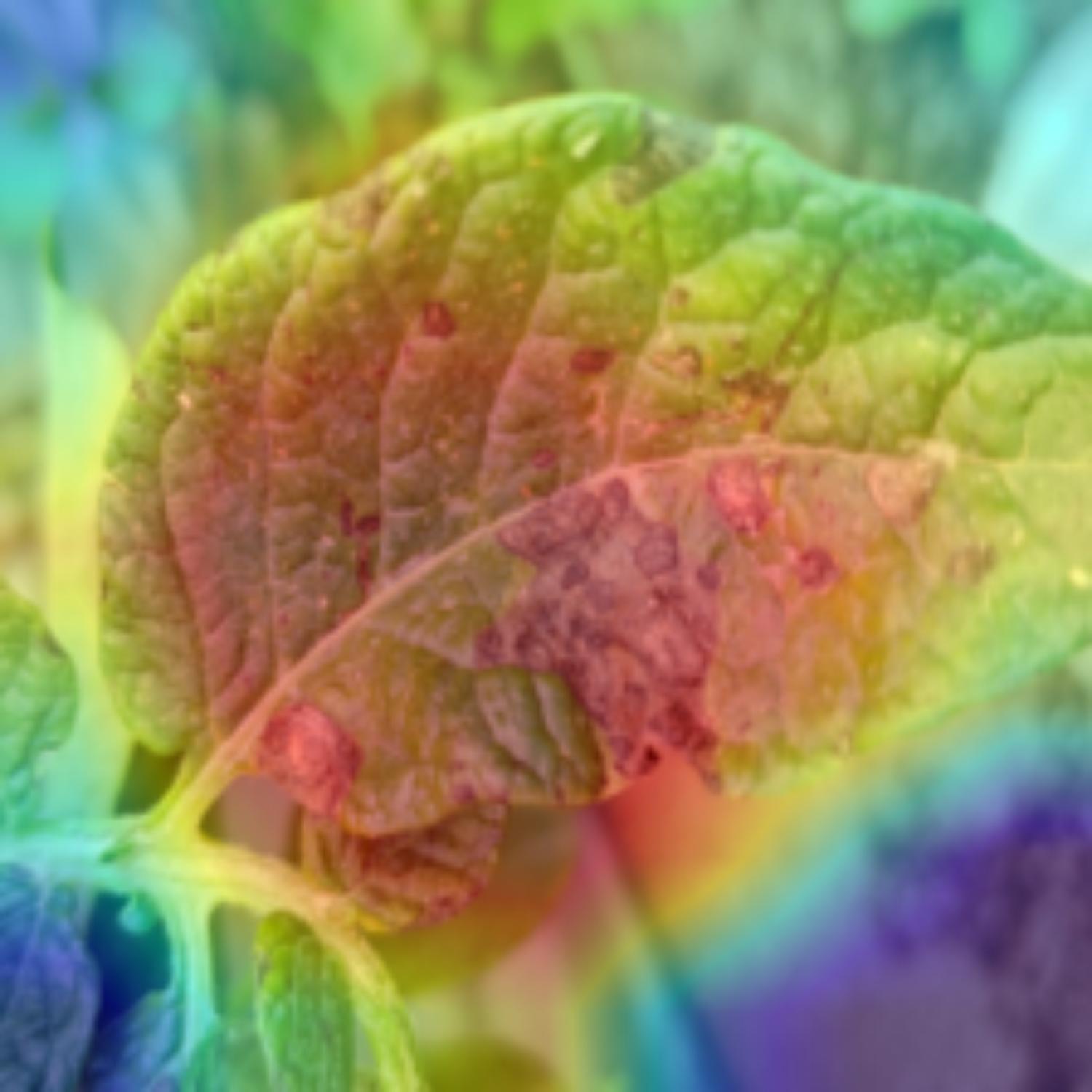}
    \caption{MoE}
    \label{fig:gradcam_fungi_moe}
\end{subfigure}

\begin{subfigure}[t]{0.19\linewidth}
    \centering
    \includegraphics[width=\linewidth]{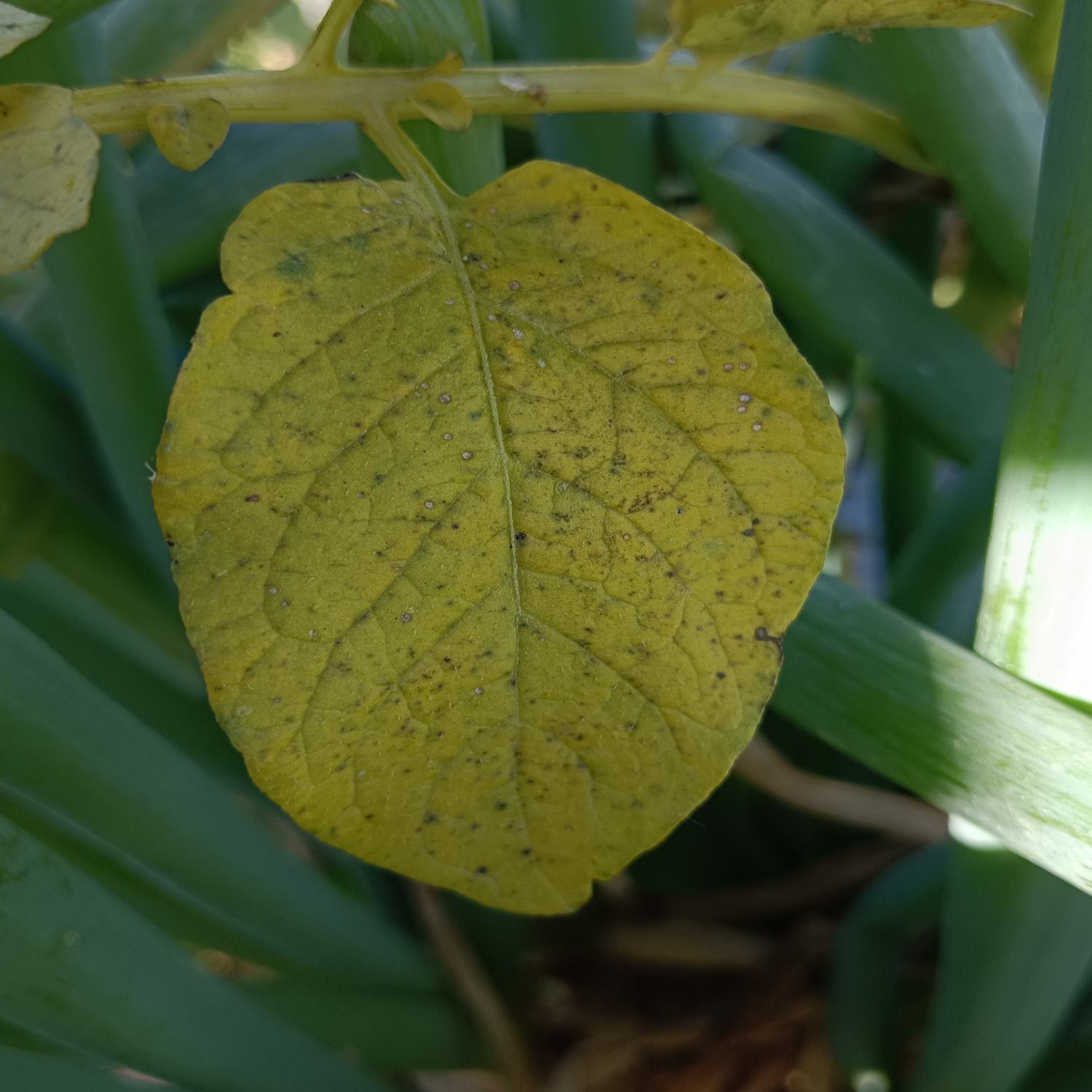}
    \caption{Original (Nematode)}
    \label{fig:gradcam_nematode_original}
\end{subfigure}
\hfill
\begin{subfigure}[t]{0.19\linewidth}
    \centering
    \includegraphics[width=\linewidth]{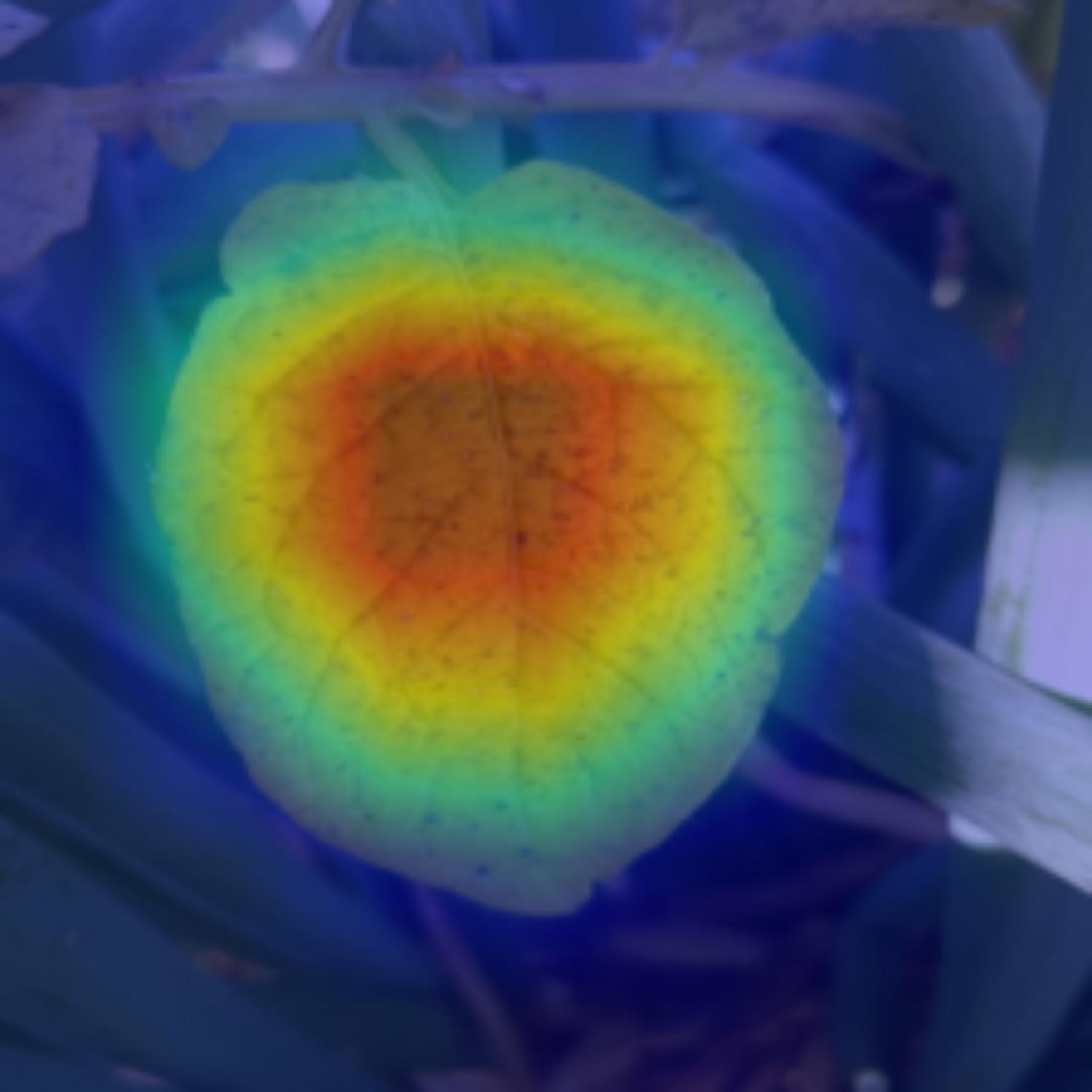}
    \caption{EfficientNet-B0 Expert}
    \label{fig:gradcam_nematode_efficientnet}
\end{subfigure}
\hfill
\begin{subfigure}[t]{0.19\linewidth}
    \centering
    \includegraphics[width=\linewidth]{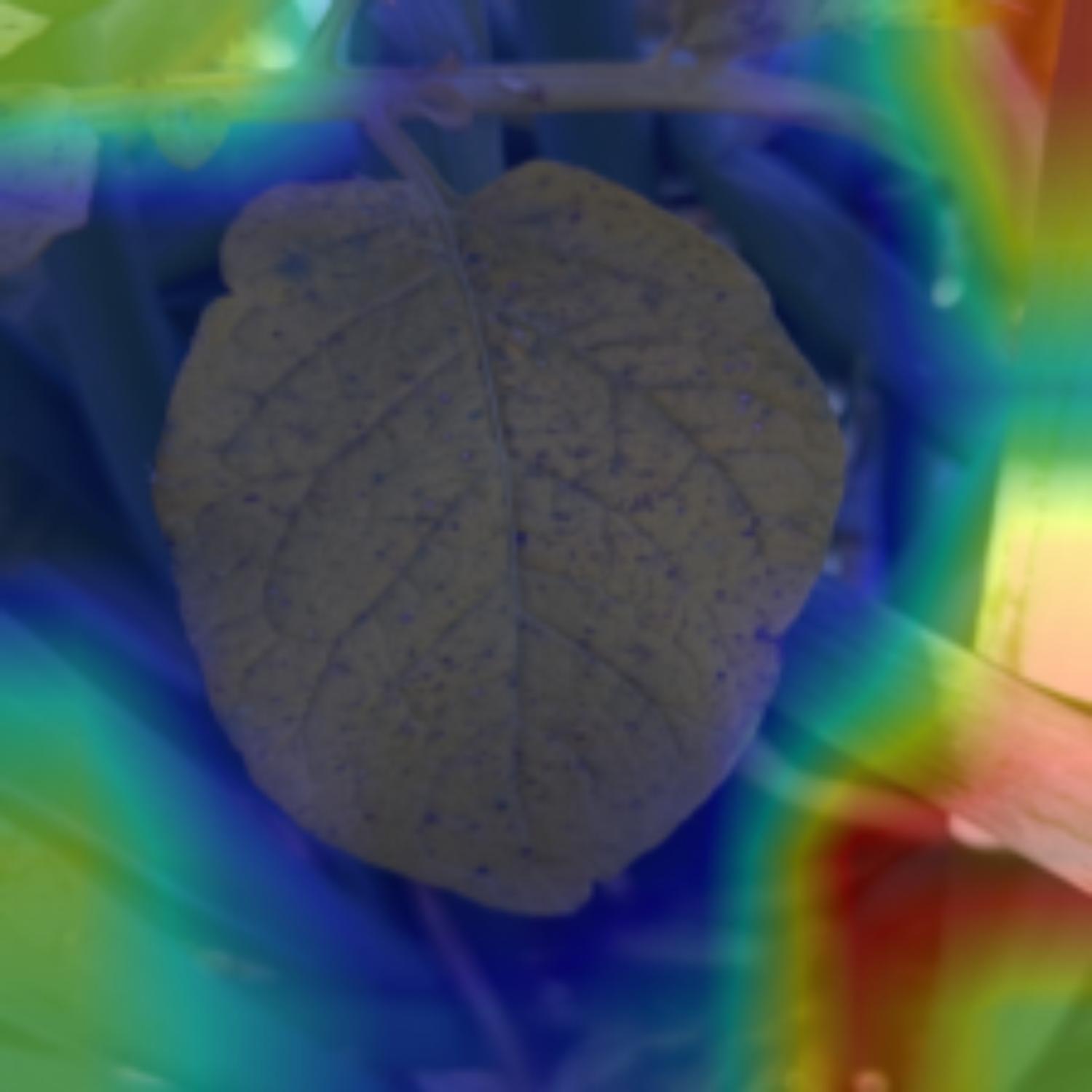}
    \caption{DenseNet-121 Expert}
    \label{fig:gradcam_nematode_densenet}
\end{subfigure}
\hfill
\begin{subfigure}[t]{0.19\linewidth}
    \centering
    \includegraphics[width=\linewidth]{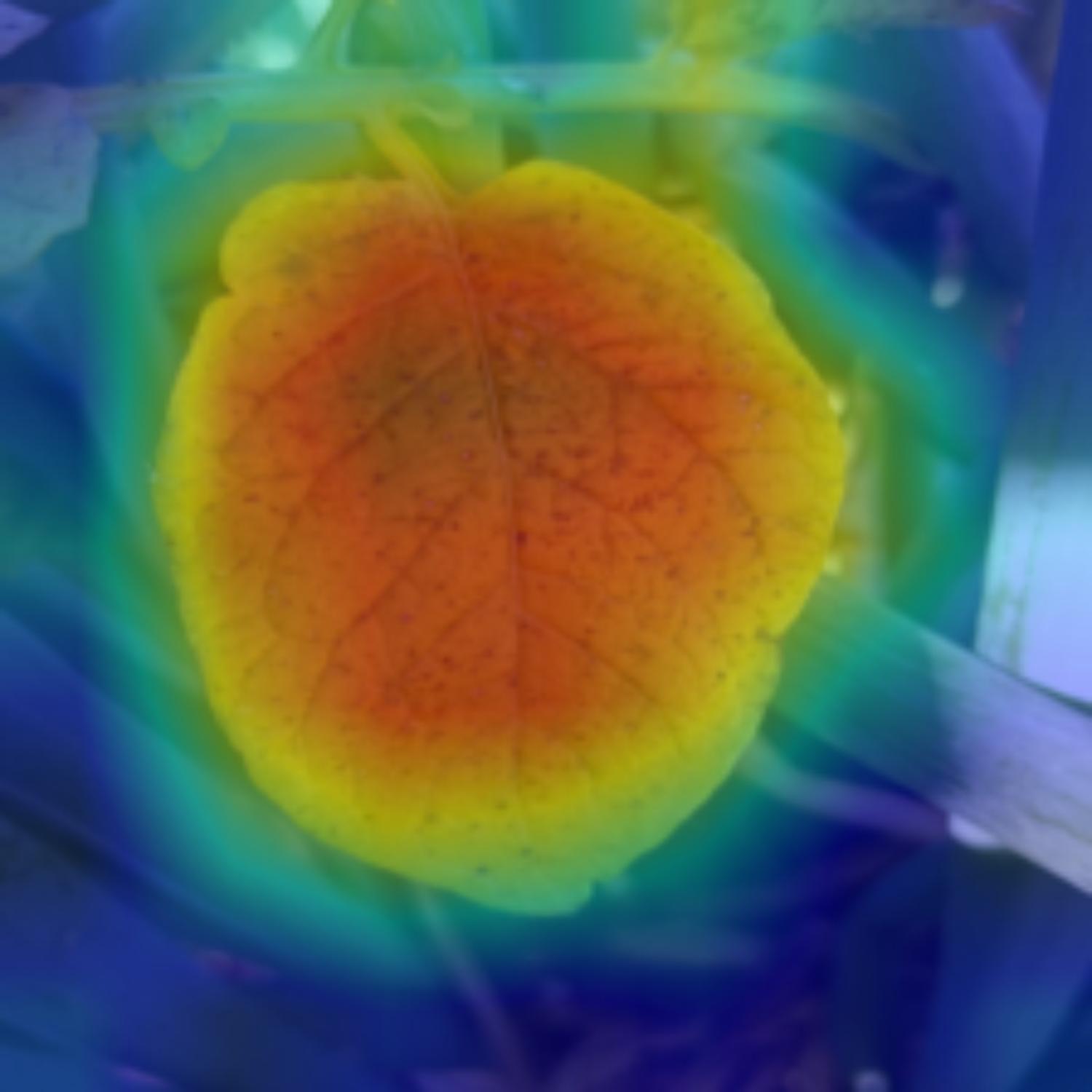}
    \caption{Swin-Tiny Expert}
    \label{fig:gradcam_nematode_swin}
\end{subfigure}
\hfill
\begin{subfigure}[t]{0.19\linewidth}
    \centering
    \includegraphics[width=\linewidth]{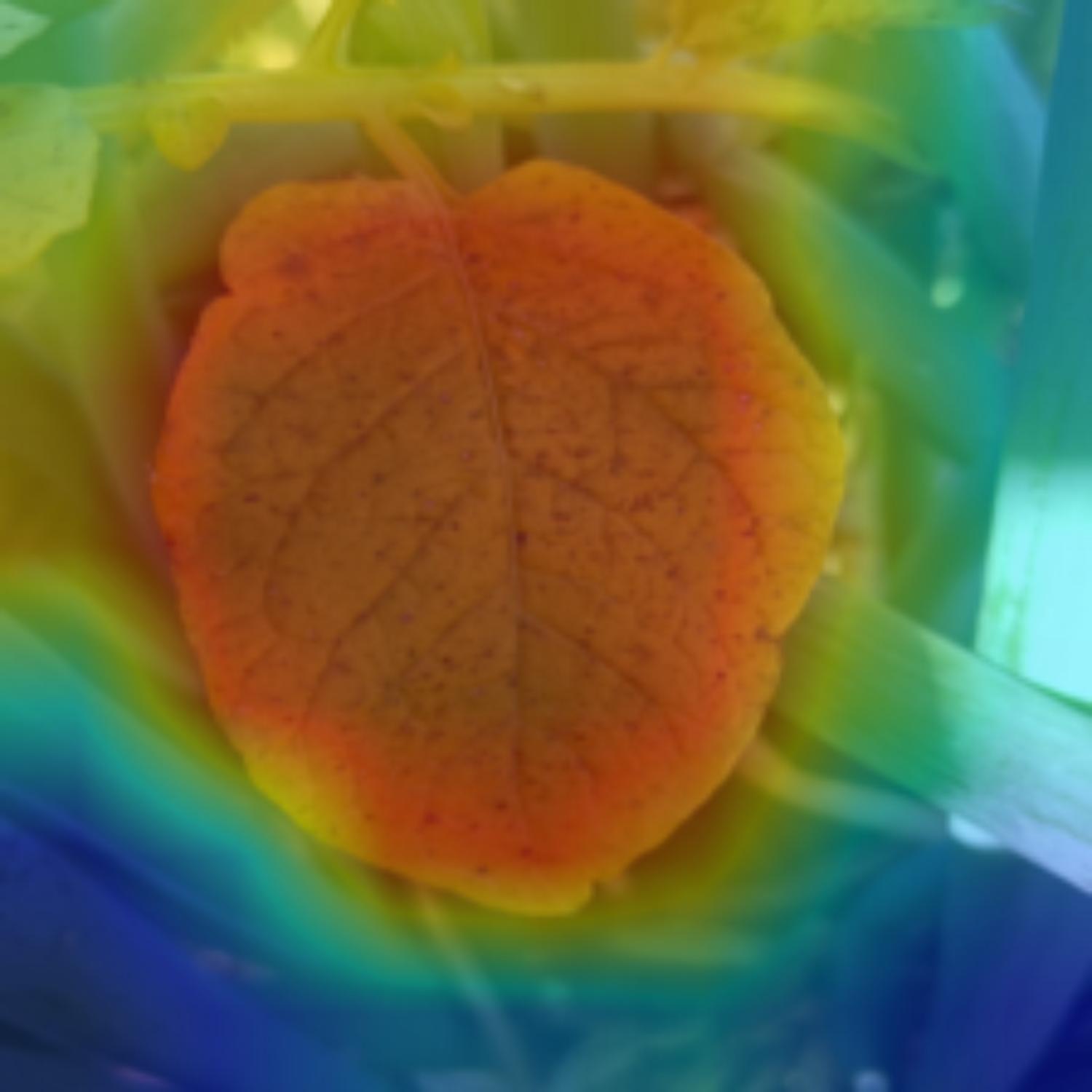}
    \caption{MoE}
    \label{fig:gradcam_nematode_moe}
\end{subfigure}

\begin{subfigure}[t]{0.19\linewidth}
    \centering
    \includegraphics[width=\linewidth]{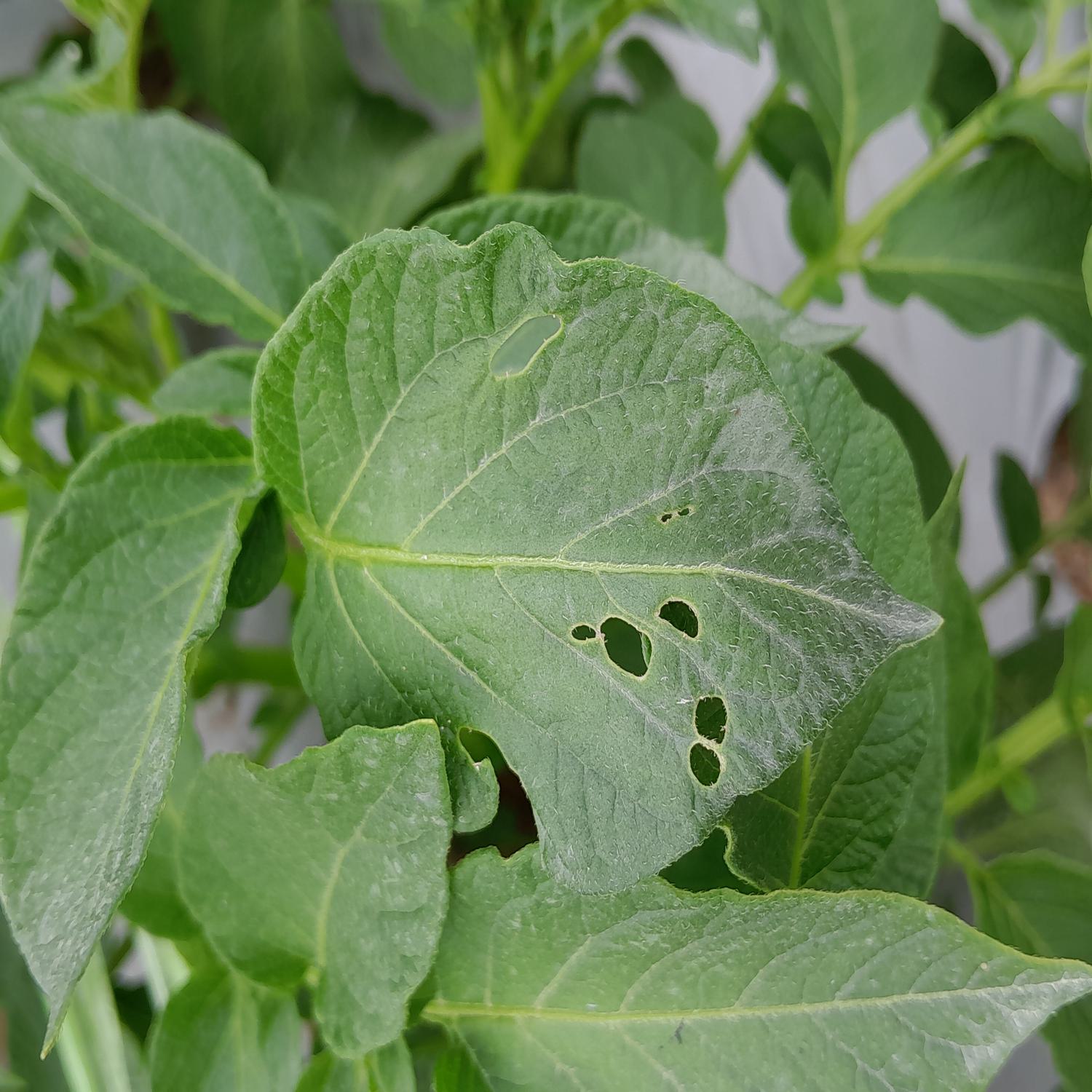}
    \caption{Original (Pest)}
    \label{fig:gradcam_pest_original}
\end{subfigure}
\hfill
\begin{subfigure}[t]{0.19\linewidth}
    \centering
    \includegraphics[width=\linewidth]{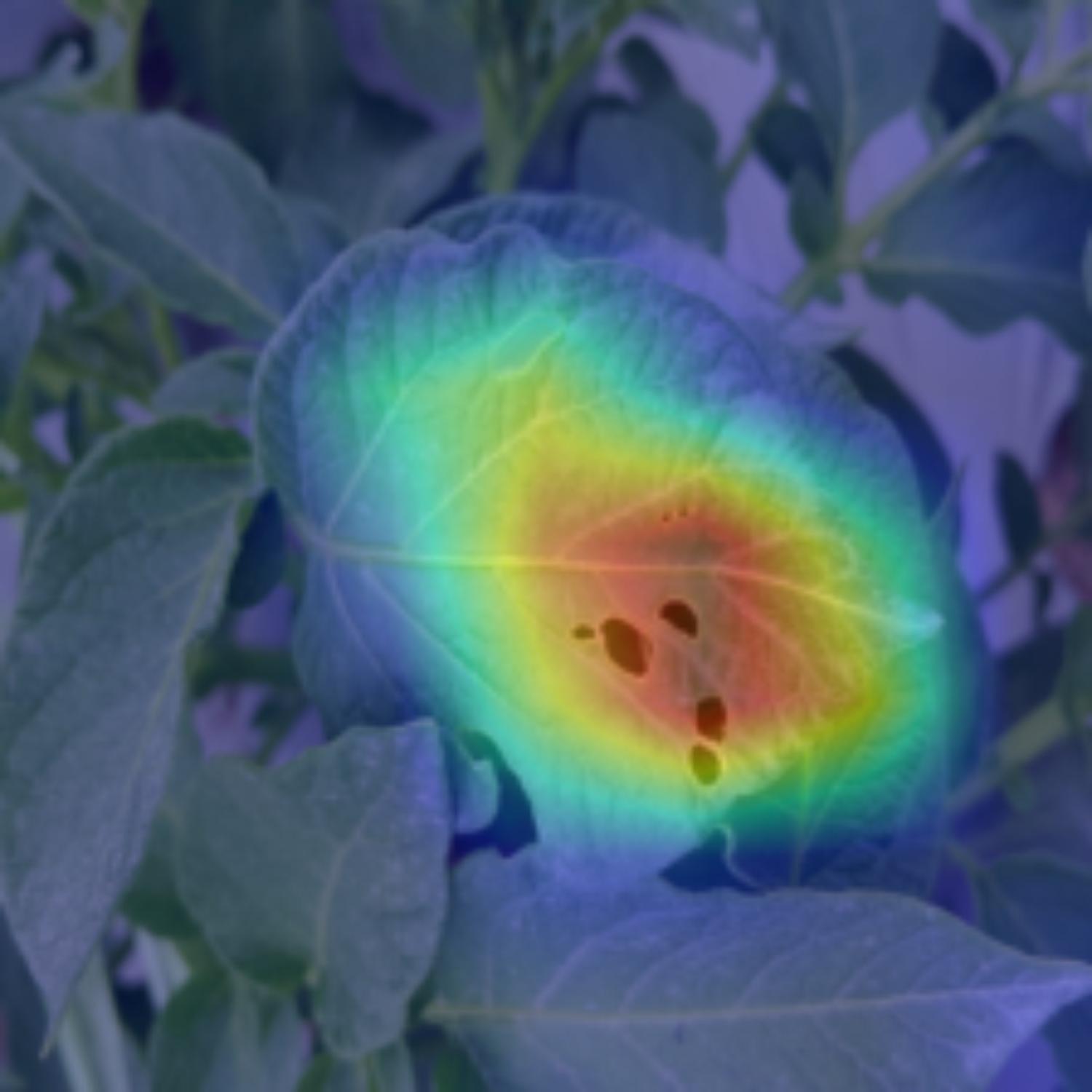}
    \caption{EfficientNet-B0 Expert}
    \label{fig:gradcam_pest_efficientnet}
\end{subfigure}
\hfill
\begin{subfigure}[t]{0.19\linewidth}
    \centering
    \includegraphics[width=\linewidth]{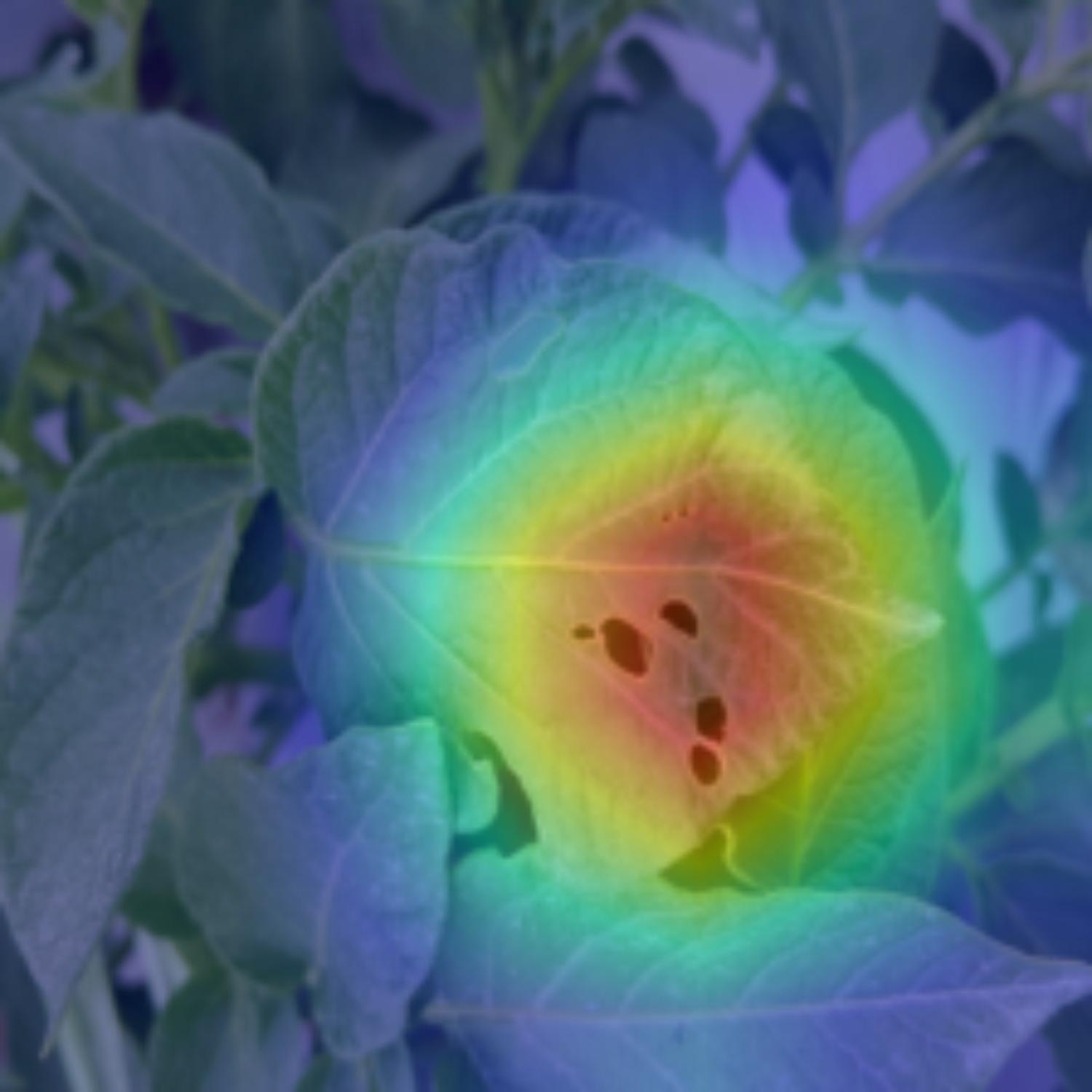}
    \caption{DenseNet-121 Expert}
    \label{fig:gradcam_pest_densenet}
\end{subfigure}
\hfill
\begin{subfigure}[t]{0.19\linewidth}
    \centering
    \includegraphics[width=\linewidth]{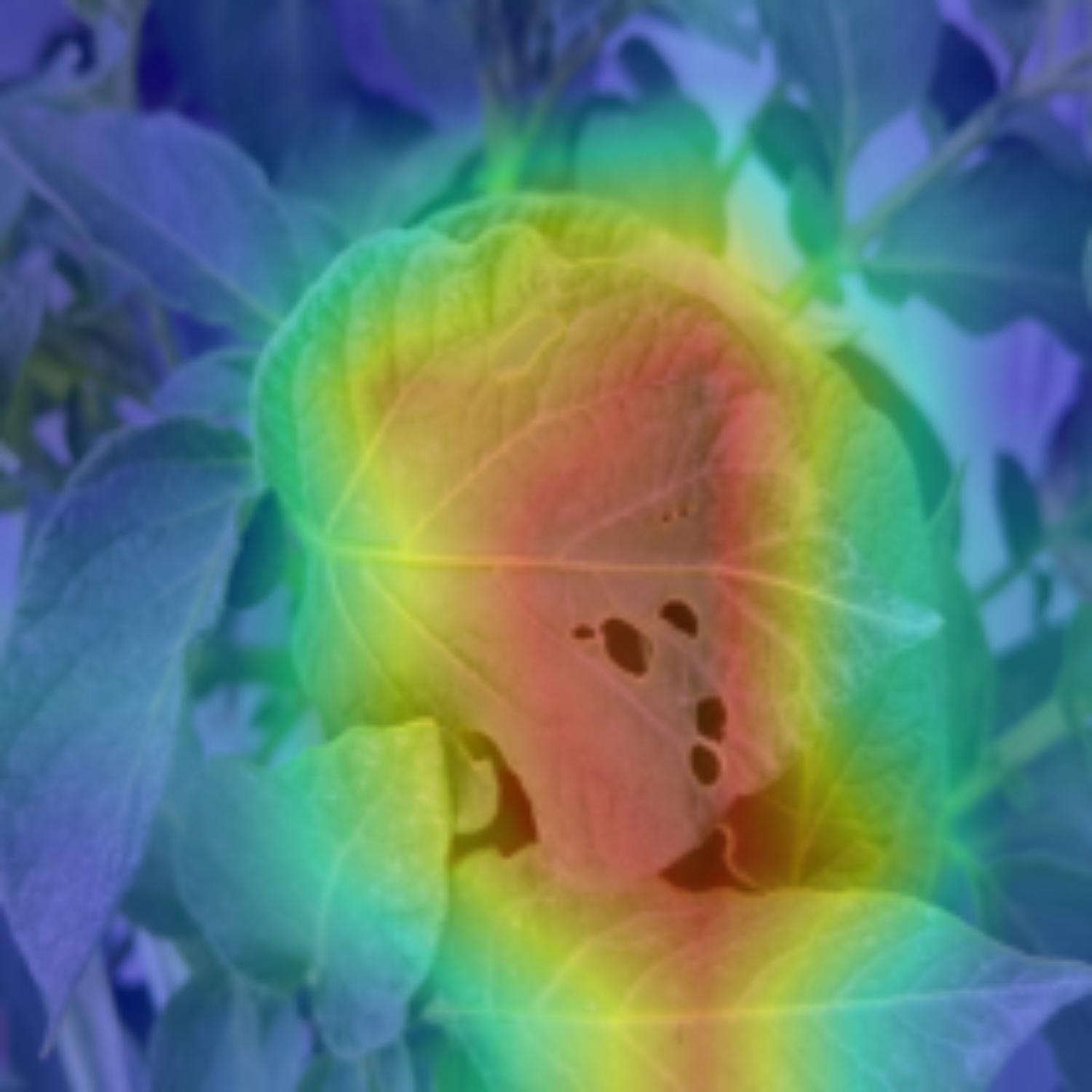}
    \caption{Swin-Tiny Expert}
    \label{fig:gradcam_pest_swin}
\end{subfigure}
\hfill
\begin{subfigure}[t]{0.19\linewidth}
    \centering
    \includegraphics[width=\linewidth]{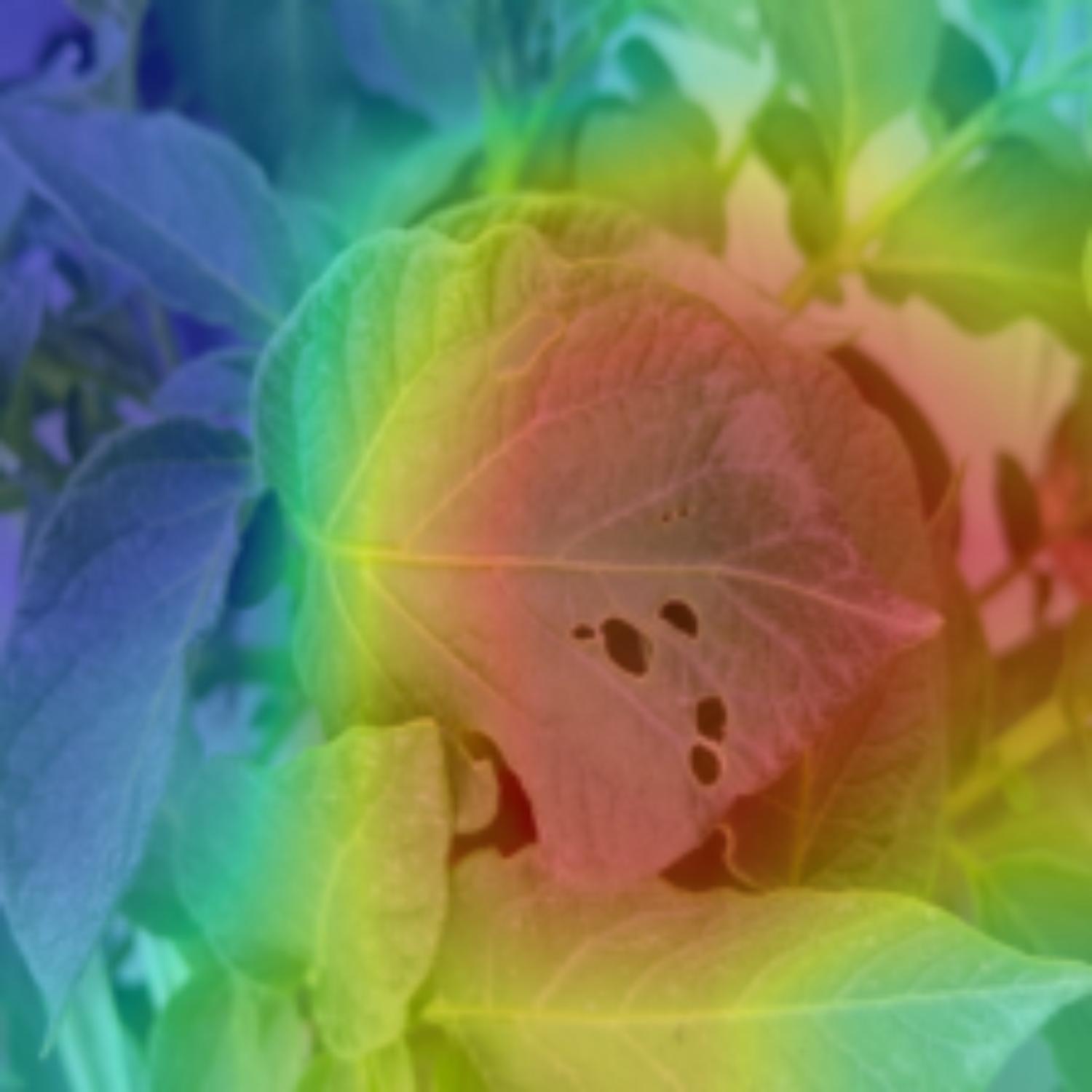}
    \caption{MoE}
    \label{fig:gradcam_pest_moe}
\end{subfigure}

\vspace{-0.5em}
\caption{Grad-CAM visualizations for representative samples across five potato leaf classes (Bacteria, Healthy, Fungi, Nematode, and Pest). Each row corresponds to a class, while columns present the original image, individual expert models, and the proposed MoE framework.}
\label{fig:gradcam}

\end{figure}

The attention behavior of the proposed framework and the visual evidence of expert specialization were further examined using Grad-CAM-based attention heatmaps. Figure~\ref{fig:gradcam} presents representative visualizations for five classes from the potato leaf disease dataset (Bacteria, Healthy, Fungi, Nematode, and Pest). These comparisons provide qualitative insight into how the individual experts (EfficientNet-B0, DenseNet-121, and Swin-Tiny) and the MoE framework emphasize different regions of interest, revealing how complementary visual features are integrated during prediction.

The visual observations reveal a clear relationship between the architectural characteristics of each expert and their responses to different pathological patterns. In the Bacteria class, characterized by small and scattered lesion regions, DenseNet-121 provides the most localized activation around the affected area, consistent with its strength in capturing fine-grained local texture characteristics. In contrast, Swin-Tiny exhibits more diffuse activations over a broader region, while EfficientNet-B0 partially attends to less relevant leaf boundaries. The MoE heatmap appears to preserve the localized sensitivity observed in DenseNet-121 while suppressing less informative activations, resulting in more focused attention around the pathological region.

For disease classes involving broader structural damage, the experts exhibit complementary behaviors. In the Pest sample, where damage appears as clustered physical holes, DenseNet-121 focuses on localized damaged regions, whereas Swin-Tiny captures a broader structural context surrounding the affected area. EfficientNet-B0 also attends to the lesion region but with less precise localization. The MoE framework integrates these complementary responses, maintaining concentrated activation around the lesion while preserving relevant contextual information.

A similar collaborative behavior is observed in the Nematode class, where disease symptoms appear as widespread discoloration across the leaf surface. Both Swin-Tiny and EfficientNet-B0 emphasize broader structural regions rather than localized details, and the MoE heatmap reflects a similar whole-region attention pattern, suggesting collaborative integration of these broader representations.

The effectiveness of complementary feature integration is particularly evident in the Fungi class, where the pathology includes both a necrotic center and a surrounding affected region. DenseNet-121 concentrates on the central lesion area, while Swin-Tiny captures a broader surrounding context. The MoE heatmap combines these responses, preserving strong activation at the lesion center while extending attention across the affected surrounding region, resulting in a more comprehensive visual representation.

For Healthy leaf samples, where no localized lesions are present, the classification task depends more on modeling overall leaf appearance rather than detecting specific pathological regions. EfficientNet-B0 emphasizes broader leaf appearance, while the MoE framework further refines this behavior by focusing attention more consistently on the leaf region while reducing irrelevant background influence.

Overall, the Grad-CAM visualizations provide qualitative evidence that the soft gating mechanism adaptively combines complementary expert representations, leading to more coherent and interpretable attention behavior than any individual expert alone.

\subsection{Validation on additional datasets}

To examine the generalization capability of the proposed MoE framework beyond the primary potato benchmark, additional experiments were conducted on the durian and sesame disease datasets. Their differing crop characteristics, class distributions, and visual patterns provide a more challenging setting for cross-dataset validation.

\subsubsection{Durian disease dataset}

Table~\ref{tab:durian_results} presents the classification performance of the baseline models and the proposed framework on the durian disease dataset. Compared with the potato benchmark, this dataset introduces a substantially different evaluation setting, including a different crop domain, a larger number of disease categories, and distinct visual symptom patterns collected under real-field conditions. Evaluating performance under these conditions provides a stronger assessment of the framework’s cross-dataset generalization capability.

\begin{table}[H]
\caption{Performance of baseline models and the proposed MoE framework on the durian disease dataset (\%).}
\label{tab:durian_results}
\footnotesize
\centering
\begin{tabular}{|l|c|c|c|c|}
\hline
\textbf{Model} & \textbf{Accuracy} & \textbf{Precision} & \textbf{Recall} & \textbf{F1-score} \\
\hline
MobileNet-V2        & 84.57 & 85.12 & 84.70 & 84.47 \\
EfficientNet-B0     & 88.89 & 89.31 & 89.00 & 88.79 \\
DenseNet-121        & 89.20 & 89.35 & 89.37 & 89.19 \\
ResNet-50           & 88.73 & 90.54 & 88.83 & 89.25 \\
Swin-Tiny           & 89.35 & 89.87 & 89.44 & 89.28 \\
\hline
\textbf{Proposed framework} & \textbf{93.98} & \textbf{94.18} & \textbf{94.02} & \textbf{94.03} \\
\hline
\end{tabular}
\end{table}

Among the standalone baseline architectures, Swin-Tiny achieves the strongest performance, obtaining an accuracy of 89.35\%, precision of 89.87\%, recall of 89.44\%, and an F1-score of 89.28\%. This result suggests that transformer-based contextual modeling remains effective for capturing the broader visual dependencies present in durian disease symptoms. DenseNet-121 and EfficientNet-B0 also achieve competitive performance, with F1-scores of 89.19\% and 88.79\%, respectively, indicating that each architectural paradigm retains meaningful representational capability when used independently.
However, the proposed framework consistently surpasses all standalone architectures across every evaluation metric. Specifically, it achieves an accuracy of 93.98\%, precision of 94.18\%, recall of 94.02\%, and an F1-score of 94.03\%. Compared with the strongest single baseline (Swin-Tiny), this corresponds to absolute improvements of 4.63\% in accuracy, 4.31\% in precision, 4.58\% in recall, and 4.75\% in F1-score.
The consistency of these gains is particularly significant. Rather than improving only overall accuracy, the framework demonstrates balanced improvement across precision, recall, and F1-score, indicating that the performance gains are systematic rather than metric-specific. This suggests that the adaptive soft gating mechanism successfully preserves and integrates complementary strengths from heterogeneous expert architectures, allowing the framework to generalize effectively beyond the dataset used for expert selection.

These findings provide evidence that the proposed MoE framework is not simply tailored to the primary potato benchmark but exhibits promising generalization across datasets with differing crop domains, disease distributions, and visual symptom characteristics.

\subsubsection{Sesame leaf disease dataset}

The classification results on the sesame dataset are reported in Table~\ref{tab:sesame_results}. In contrast to the durian dataset, all standalone architectures achieve notably strong performance, with F1-scores exceeding 95\%, suggesting that this dataset presents a comparatively more separable classification setting with less severe visual ambiguity.
Among the individual baselines, DenseNet-121 achieves the strongest overall balanced performance, obtaining an accuracy of 96.75\%, precision of 96.67\%, recall of 96.54\%, and an F1-score of 96.58\%. Swin-Tiny also performs competitively, achieving the highest recall among standalone models (96.88\%) and an F1-score of 96.28\%. These results indicate that both fine-grained local texture modeling and broader contextual feature extraction remain effective in this dataset.

Despite the already strong baseline performance, the proposed framework still achieves the best overall results, reaching an accuracy of 97.32\%, precision of 96.96\%, recall of 97.16\%, and an F1-score of 97.04\%. Compared with the strongest standalone baseline, this corresponds to consistent absolute improvements of 0.57\% in accuracy, 0.29\% in precision, 0.62\% in recall, and 0.46\% in F1-score.
Although the performance margin is smaller than that observed on the durian dataset, this outcome is expected given the already high baseline performance, which leaves less room for improvement. Importantly, the framework continues to deliver consistent gains across all evaluation metrics, indicating that the benefit of adaptive expert fusion is not limited to challenging or highly ambiguous datasets. Instead, the results suggest that the proposed routing strategy provides consistent gains on the evaluated datasets even when individual expert models already achieve near-ceiling performance.

\begin{table}[H]
\caption{Performance of baseline models and the proposed MoE framework on the Sesame leaf disease dataset (\%).}
\label{tab:sesame_results}
\footnotesize
\centering
\begin{tabular}{|l|c|c|c|c|}
\hline
\textbf{Model} & \textbf{Accuracy} & \textbf{Precision} & \textbf{Recall} & \textbf{F1-score} \\
\hline
MobileNet-V2        & 95.76 & 95.26 & 95.12 & 95.19 \\
EfficientNet-B0     & 95.90 & 95.61 & 95.46 & 95.53 \\
DenseNet-121        & 96.75 & 96.67 & 96.54 & 96.58 \\
ResNet-50           & 96.19 & 95.57 & 96.04 & 95.74 \\
Swin-Tiny           & 96.47 & 95.83 & 96.88 & 96.28 \\
\hline
\textbf{Proposed framework}        & \textbf{97.32} & \textbf{96.96} & \textbf{97.16} & \textbf{97.04} \\
\hline
\end{tabular}
\end{table}

\subsection{Comparison with previous studies}

\begin{figure}[H]
    \centering
    \includegraphics[width=\textwidth]{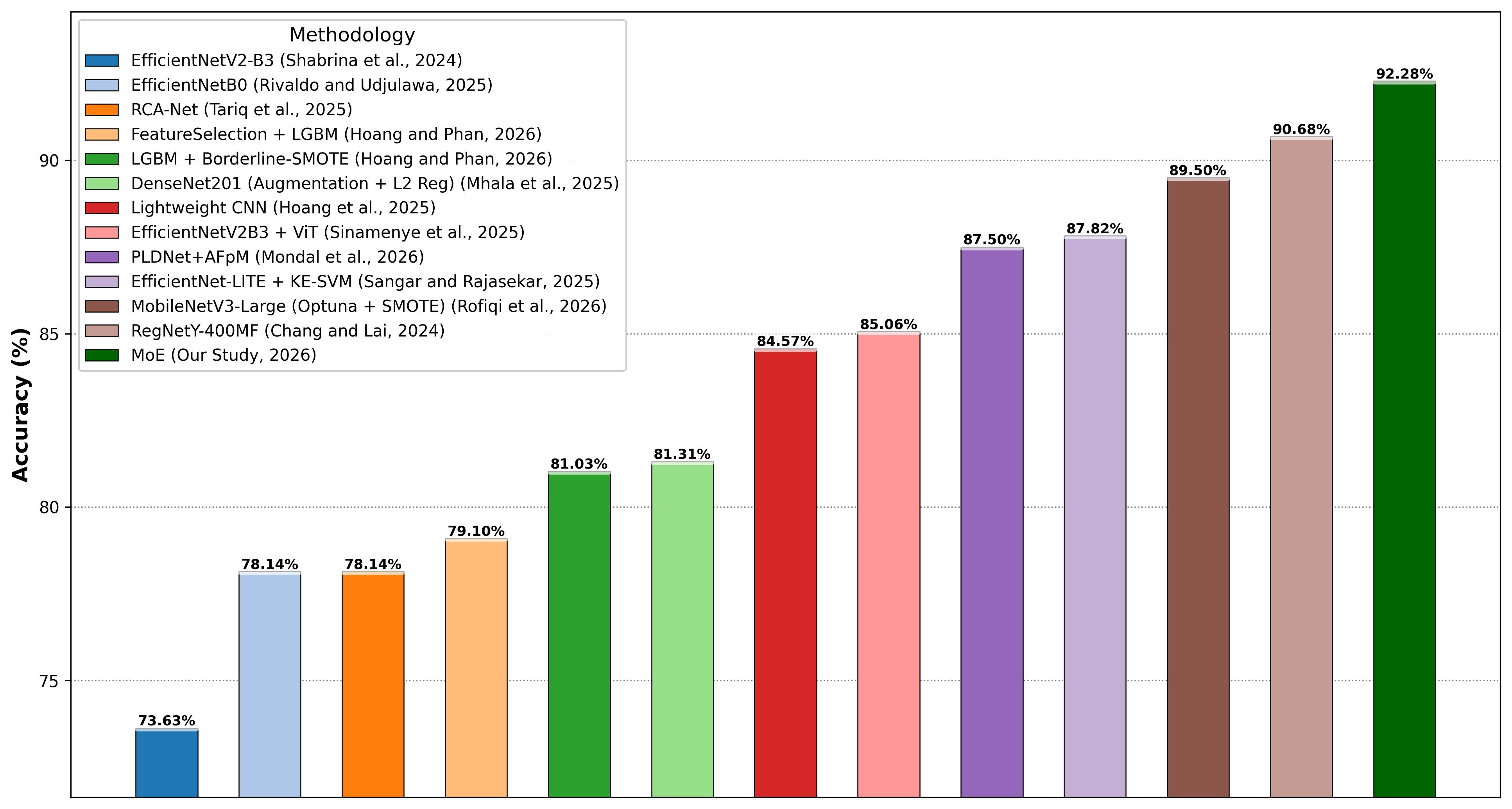}
    \caption{Performance comparison of the proposed MoE framework with existing methods on the potato leaf disease dataset.}
    \label{fig:comparative_previous_study}
\end{figure}

To assess the effectiveness of the proposed framework, its performance was compared with 12 recent studies evaluated on the same primary potato leaf disease dataset. As illustrated in Figure~\ref{fig:comparative_previous_study}, existing approaches can be broadly grouped into three categories: single-architecture deep learning models, feature engineering and imbalance mitigation approaches, and hybrid or ensemble-based methods.

Early studies primarily explored single CNN architectures on this dataset. For example, Shabrina et al.~\cite{Shabrina2024} reported 73.63\% using EfficientNetV2-B3, while subsequent works improved performance through architectural refinement and training optimization. Rivaldo and Udjulawa~\cite{Rivaldo2025} achieved 78.14\% with EfficientNet-B0, Mhala et al.~\cite{Mhala2025} reported 81.31\% using DenseNet201 with augmentation and regularization, and Chang and Lai~\cite{Chang2024} achieved 90.68\% using RegNetY-400MF. Although these methods demonstrate progressively stronger performance, they remain constrained by the fixed representational bias of a single architecture.

Traditional machine learning approaches have also been explored to address class imbalance and feature optimization. Hoang and Phan (2026b)~\cite{Hoang2026FeatureSelection} achieved 79.10\% using LGBM with handcrafted features, improving to 81.03\% with Borderline-SMOTE~\citep{Hoang2026BorderlineSMOTE}. Similarly, Rofiqi et al.\cite{Rofiqi2026} combined MobileNetV3-Large with Optuna tuning and SMOTE to reach 89.50\%. While effective, these methods rely on feature engineering or synthetic resampling rather than adaptive representation learning.

More recent studies have attempted to integrate multiple feature representations through hybrid or ensemble-based designs. Examples include RCA-Net~\citep{Tariq2025} (78.14\%), lightweight CNN-based fusion~\citep{Hoang2025} (84.57\%), EfficientNetV2-B3 + ViT~\citep{Sinamenye2025} (85.06\%), PLDNet~\citep{Mondal2026} (87.50\%), and EfficientNet-LITE + KE-SVM~\citep{Sangar2025} (87.82\%). Although these approaches combine diverse representational strengths, they generally rely on predefined integration strategies that do not adapt dynamically at the sample level.

In contrast, the proposed framework achieves the highest reported accuracy of 92.28\% on this dataset. This improvement is attributed to the adaptive soft gating mechanism, which dynamically adjusts the contribution of DenseNet-121, EfficientNet-B0, and Swin-Tiny according to input characteristics. Unlike conventional ensemble strategies with fixed integration behavior, the proposed design enables sample-dependent expert routing, while also mitigating class imbalance effects through adaptive expert allocation rather than explicit data-level balancing.

\subsection{Limitations}

Despite its strong classification performance, the proposed framework presents several limitations. The primary limitation is computational efficiency. As shown in Table~\ref{tab:efficiency}, integrating three heterogeneous expert networks substantially increases computational cost compared with standalone architectures, resulting in higher FLOPs, parameter count, training time, and inference latency. The proposed framework requires 12.73 GFLOPs and 40.19 million parameters, with an inference latency of 39.85 ms/sample, making direct deployment more challenging on resource-constrained edge devices commonly used in practical agricultural monitoring systems.

\begin{table}[H]
\centering
\caption{Computational efficiency comparison of baseline deep learning models and the proposed method on the potato leaf disease dataset in terms of FLOPs, parameter count, training time, and inference latency.}
\label{tab:efficiency}
\footnotesize   
\renewcommand{\arraystretch}{1.0}
    \resizebox{\textwidth}{!}{
        \begin{tabular}{lcccc}
        \toprule
        Model & FLOPs (G) & Params (M) & Training Time (m:s) & Latency (ms/sample) \\
        \midrule
        Swin-Tiny       & 5.95 & 27.52 & 31:44.96 & $16.04 \pm 2.13$ \\
        ResNet-50       & 8.26 & 23.52 & 38:24.63 & $8.30 \pm 1.14$ \\
        DenseNet-121    & 5.79 & 6.96 & 30:17.48 & $18.56 \pm 2.63$ \\
        EfficientNet-B0 & 0.83 & 4.02 & 26:43.17 & $11.92 \pm 1.62$ \\
        MobileNet-V2    & 0.65 & 2.23 & 32:25.88 & $7.61 \pm 1.55$ \\
        \midrule
        Proposed (MoE)  & 12.73 & 40.19 & 75:29.40 &  $39.85 \pm 3.79$ \\
        \bottomrule
        \end{tabular}
}
\end{table}

\begin{figure}[H]
    \centering
    
    \begin{subfigure}{\linewidth}
        \centering
        \includegraphics[width=\linewidth]{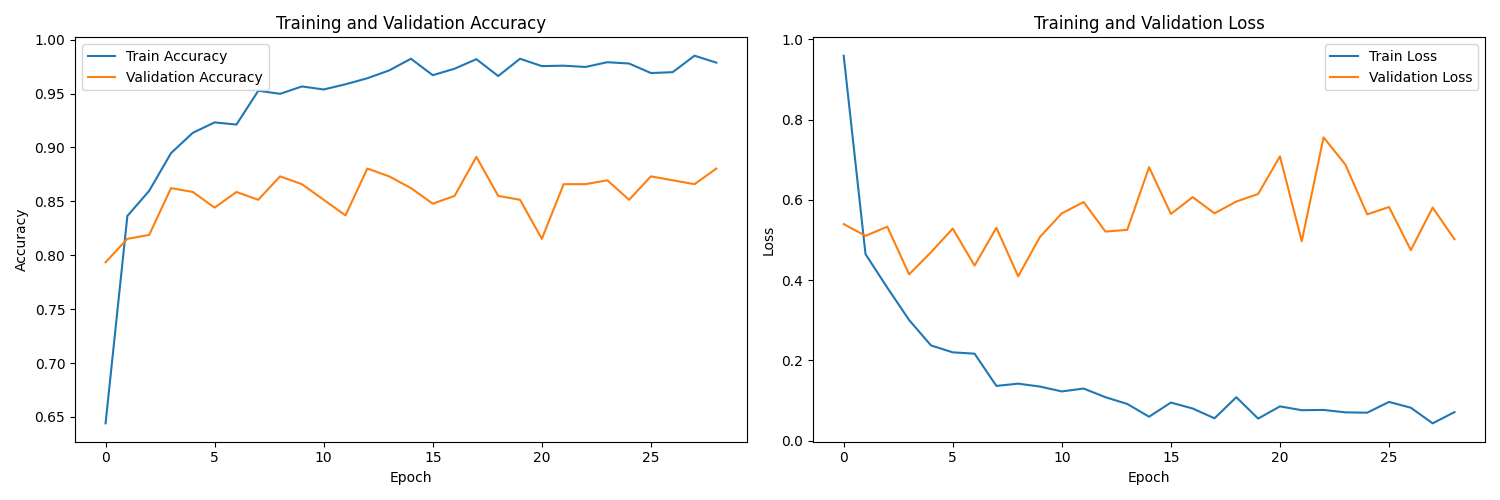}
        \caption{Phase 1 (fine-tuning)}
    \end{subfigure}
    
    \begin{subfigure}{\linewidth}
        \centering
        \includegraphics[width=\linewidth]{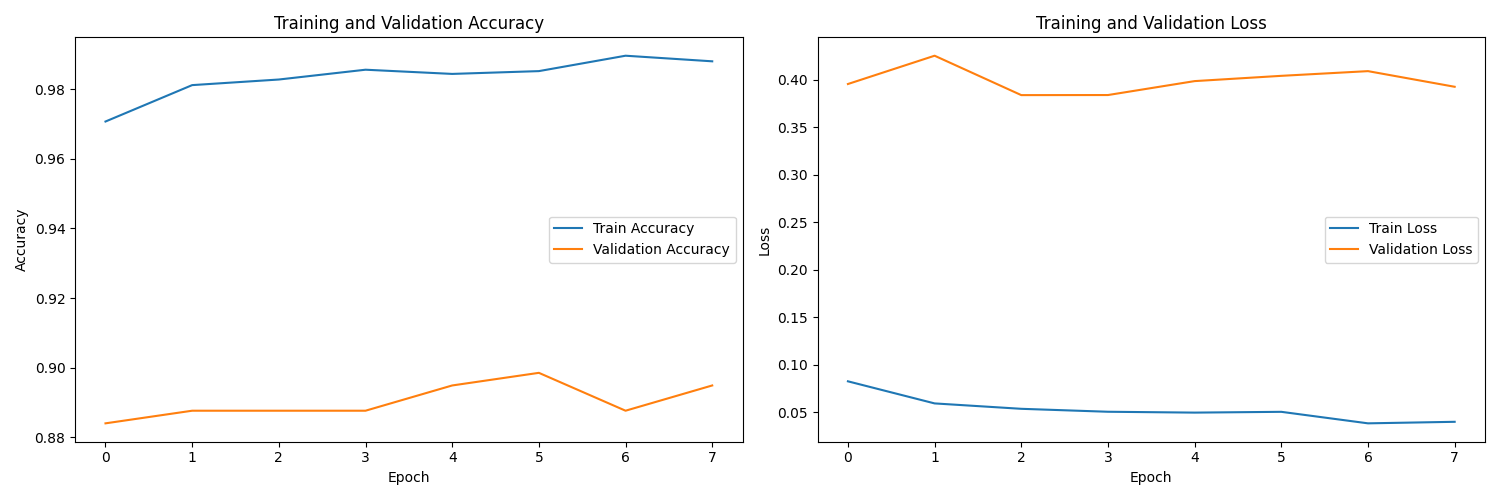}
        \caption{Phase 2 (refinement)}
    \end{subfigure}
    
    \caption{Training and validation accuracy and loss curves of the proposed MoE framework across Phase 1 (fine-tuning) and Phase 2 (refinement) on the potato leaf disease dataset.}
    \label{fig:learning_curves}
\end{figure}

Second, although the two-stage refinement strategy improves optimization stability, the learning curves in Figure~\ref{fig:learning_curves} indicate a remaining generalization gap between training and validation performance, particularly during Phase 1, where training accuracy approaches saturation while validation metrics stabilize at a lower level. Although the refinement stage reduces instability, this gap is not fully eliminated, suggesting that the high representational capacity of the framework may still introduce some overfitting risk under real-world visual variability. This behavior likely reflects sensitivity to irrelevant environmental factors such as background clutter, shadows, and illumination changes, despite the use of data augmentation and regularization.

Third, the current expert selection strategy relies on manually chosen backbone combinations based on validation performance and architectural complementarity. While effective in this study, this design does not guarantee globally optimal expert composition and may require re-evaluation for different datasets or application domains.

\section{Conclusion}\label{sec:conclusion}

This study proposes an adaptive soft Mixture-of-Experts (MoE) framework for robust plant leaf disease classification by integrating three heterogeneous architectures--EfficientNet-B0, DenseNet-121, and Swin-Tiny--as complementary experts. On the primary potato leaf disease benchmark, the proposed framework achieved 92.28\% accuracy and 92.62\% F1-score, outperforming the compared methods, including the strongest standalone baseline, with improvements of 2.89\% in accuracy and 5.03\% in F1-score over the best single-model competitor.
These gains suggest that adaptive routing helps improve recognition of difficult and minority classes under imbalanced conditions. Gating behavior and Grad-CAM analyses further confirmed meaningful expert collaboration and disease-relevant attention patterns. Additionally, cross-dataset evaluations on durian and sesame datasets yielded F1-scores of 94.03\% and 97.04\%, respectively, indicating strong generalization capability. The proposed framework offers a promising strategy for robust plant disease diagnosis under real-world field conditions. However, the computational overhead of multiple experts remains a limitation to large-scale deployment. Future work will focus on developing more efficient routing and lightweight expert architectures, as well as evaluating the framework on larger and more diverse field-acquired datasets to improve scalability and real-world applicability further.




\section*{Funding}

This research did not receive any specific grant from funding agencies in the public, commercial, or not-for-profit sectors.

\section*{Declaration of competing interest}

The authors declare that they have no known competing financial interests or personal relationships that could have appeared to influence the work reported in this paper.



\section*{Declaration of generative AI and AI-assisted technologies in the writing process}

During the preparation of this manuscript, the authors used Grammarly and generative AI tools to improve grammar and refine wording for clarity. After using these tools, the authors reviewed and edited the content as needed and take full responsibility for the final version of the manuscript.

\bibliographystyle{elsarticle-num}
\bibliography{refs}

\end{document}